\documentclass[letterpaper]{article} 
\usepackage{aaai2027} 
\usepackage[hyphens]{url}  
\usepackage{graphicx} 
\usepackage{natbib}  
\usepackage{caption} 
\usepackage{algorithm}
\usepackage{algorithmic}
\usepackage{newfloat}
\usepackage{listings}
\DeclareCaptionStyle{ruled}{labelfont=normalfont,labelsep=colon,strut=off} 
\floatstyle{ruled}
\newfloat{listing}{tb}{lst}{}
\floatname{listing}{Listing}

\usepackage{booktabs}

\usepackage[table]{xcolor}
\definecolor{LightBlueA}{RGB}{245,249,252}
\definecolor{LightBlueB}{RGB}{231,241,248}
\definecolor{LightBlueC}{RGB}{214,231,244}

\definecolor{red}{RGB}{250,235,235}
\definecolor{blue}{RGB}{235,240,250} 
\definecolor{purple}{RGB}{228,220,239}
\usepackage{amsmath}
\usepackage{amssymb}
\usepackage{multirow}

\usepackage{cuted}
\title{Post-Training Pruning for Diffusion Transformers}
\author{
    Chengzhi Hu\textsuperscript{\rm 1,2},
    Xuewen Liu\textsuperscript{\rm 1,2},
    Jing Zhang\textsuperscript{\rm 1,2},
    Mengjuan Chen\textsuperscript{\rm 1},
    Zhikai Li\textsuperscript{\rm 1,*},
    Qingyi Gu\textsuperscript{\rm 1,*}
}

\affiliations{
    \textsuperscript{\rm 1}Institute of Automation, Chinese Academy of Sciences\\
    \textsuperscript{\rm 2}School of Artificial Intelligence, University of Chinese Academy of Sciences\\
    \{huchengzhi2024, liuxuewen2023, zhangjing2024, chenmengjuan2016, zhikai.li,  qingyi.gu\}@ia.ac.cn
}

\usepackage[absolute,overlay]{textpos}
\begin{document}

\maketitle
\begingroup
\renewcommand{\thefootnote}{*}
\footnotetext{Corresponding author: \{zhikai.li, qingyi.gu\}@ia.ac.cn.}
\endgroup
\begin{strip}
\vspace*{-2.8em}  
\centering
\includegraphics[width=0.95\textwidth]{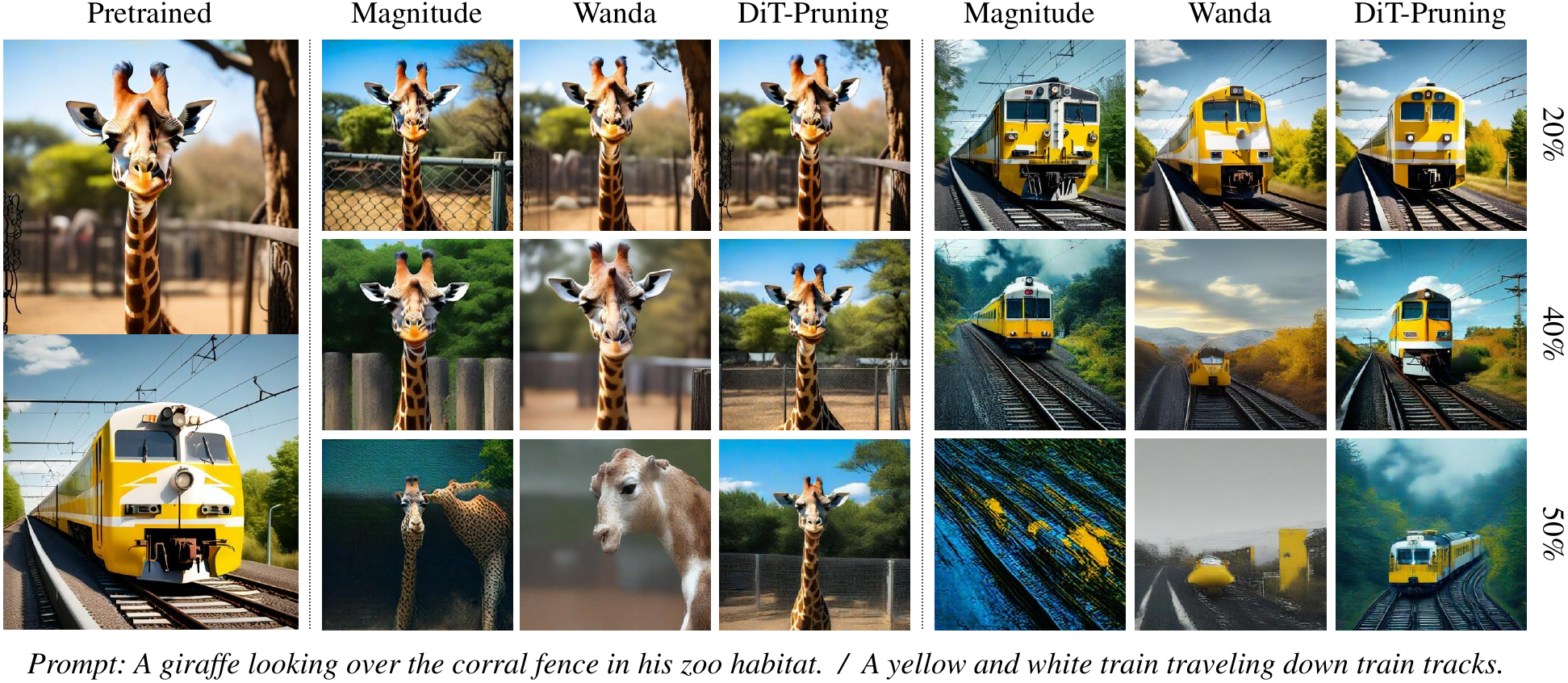}

\vspace{2.5em}

\begin{minipage}{0.95\textwidth}
\small

\vspace{1em}

\textbf{Image quality:}
Comparison of different pruning methods for DiTs. We evaluate magnitude, Wanda, and ours (DiT-Pruning)
on PixArt-$\Sigma$ model. Our method preserves modeling capability for both foreground and background of images, even under high sparsity.
\end{minipage}

\vspace{4.5em} 
\end{strip}

\begin{abstract}
Diffusion Transformers (DiTs) have demonstrated impressive performance in image generation but suffer from substantial computational overhead and high resource consumption. Post-training pruning offers a promising solution; however, due to DiTs' unique architectural design and parameter distribution, traditional pruning methods are inapplicable, leading to significant performance degradation, especially at high sparsity. Specifically, prior methods developed for LLMs, which derive metrics via series of approximations, amplify the relative contribution of weights in the saliency metric. In addition, weights in DiTs exhibit significantly larger magnitudes compared to those in LLMs. Besides, existing pruning granularity overlook variations in model structures. In this paper, we propose DiT-Pruning, which improves pruning performance by introducing customized saliency criteria and pruning granularity. We design a novel metric that balances the contributions of weights and activations from an energy-based modeling perspective, enabling more effective identification of important elements. Furthermore, we observe that DiTs exhibit different distinct clustering patterns in the two-dimensional weight space. Accordingly, we adopt a clustering-aware pruning granularity, which enables effective sparse allocation. Extensive evaluations on various DiTs show that our method consistently preserves image quality, especially under high sparsity. For FLUX.1-dev at 512$\times$512 resolution on MJHQ, DiT-Pruning achieves only a \emph{0.001 loss} in CLIP score at 50\% sparsity, dramatically outperforming recent pruning methods.
\end{abstract}

\section{Introduction}  

Diffusion Transformers~\citep{peebles2023scalable} (DiTs) exhibit remarkable performance in image generation, fully leveraging their advantages in scalability and expressive modeling capabilities. Recent models, such as PixArt ~\citep{chen2023pixart} and FLUX~\citep{labs2025flux}, have demonstrated the ability to handle complex image distributions and generate high-quality samples. However, their numerous number of parameters and prolonged inference process result in substantial computational overhead and high resource consumption, posing significant challenges for deployment in resource-constrained environments.
\begin{figure}[t] 
  \centering
  \includegraphics[width=\columnwidth]{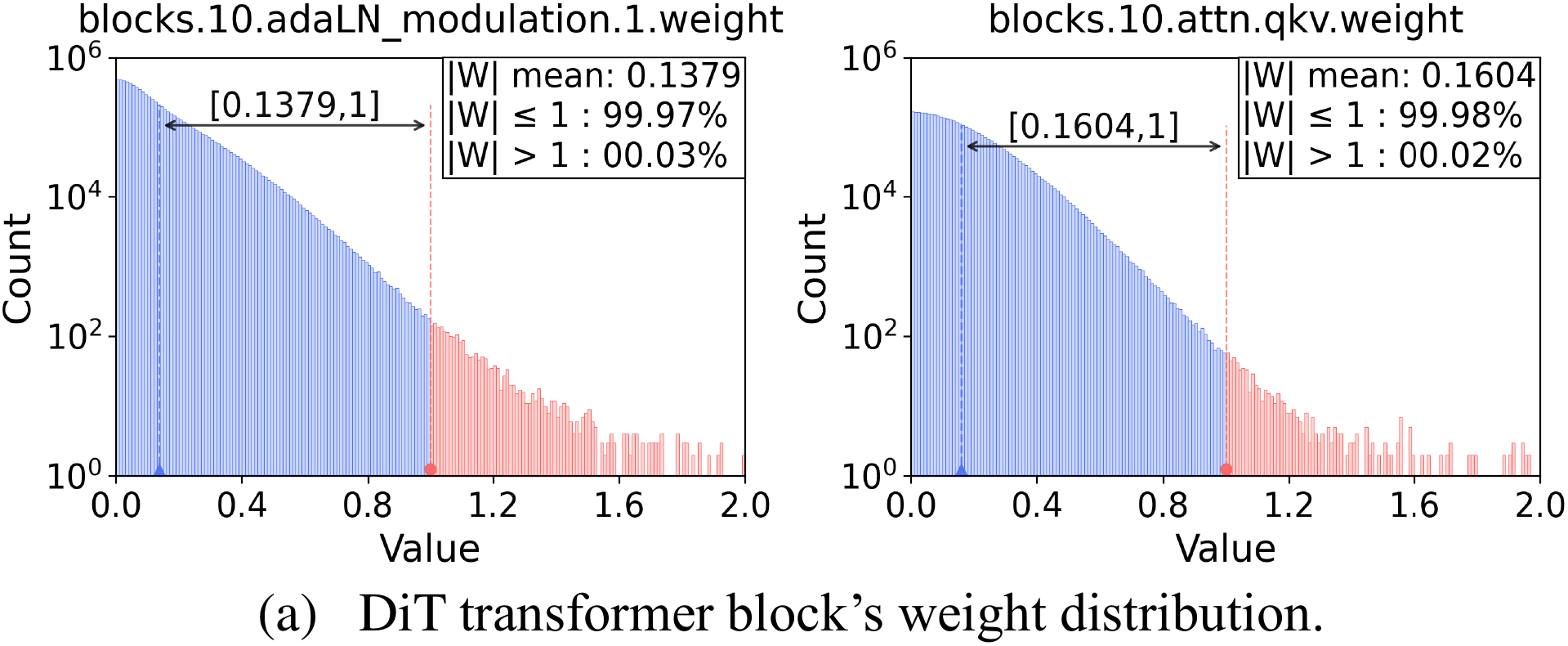}
  \vskip 0.5em 
  \includegraphics[width=\columnwidth]{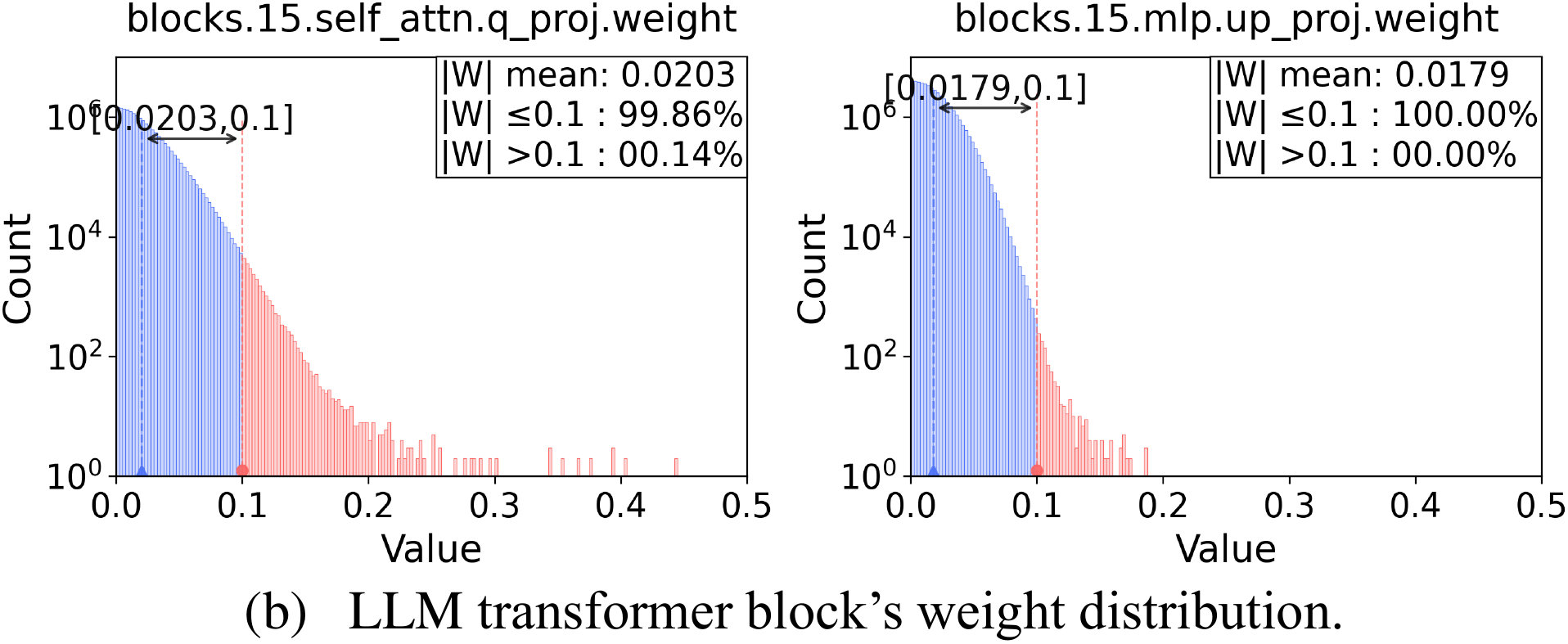}
  \caption{Weight distributions of DiT and LLM transformer layers. Weights in DiT are predominantly distributed within the range of 0 to 1 and show generally larger magnitudes, while LLM weights are concentrated in the range 0 to 0.1.
  }
  \label{weights}
\end{figure}

Neural network pruning~\citep{lecun1989optimal, hassibi1993optimal, han2015learning, frantar2022optimal} reduces computational and storage overhead by eliminating redundant weights. Post-training pruning avoids additional training and typically adopts unstructured approaches treating individual weights as sparse units. For DiTs, the massive parameter scale and complex architecture make training-based methods~\citep{liu2018rethinking, blalock2020state} prohibitively costly, whereas post-training pruning is training-free and offers strong scalability and promising potential. Its effectiveness depends on two aspects: the \textbf{saliency criteria} used to measure parameter importance, and the \textbf{pruning granularity} that dictates sparsity allocation across different structural components like channels or layers. The proper design of both factors is intrinsically associated with the architectural design and the parameter distribution.


Early CNN pruning~\citep{molchanov2016pruning} employs Taylor-based saliency metrics. Similar criteria~\citep{michel2019sixteen} have been adopted to prune attention heads in ViTs. Recently, several methods have been proposed for LLM pruning. For instance, SparseGPT~\citep{frantar2023sparsegpt} based on  Optimal Brain Surgeon(OBS) theory~\citep{hassibi1993optimal} prunes weights that minimally impact output perturbations, while Wanda~\citep{sun2023simple} introduces second-order approximations for computational efficiency. However, applicable pruning methods for Diffusion models remain largely unexplored. Some efforts have targeted U-Net-based diffusion models, for example, Diff-Pruning~\citep{fang2023structural}, which proposes dynamic pruning strategies incorporating timestep information and relies on retraining to restore performance. Nevertheless, due to the unique inference paradigm, architectural design, and parameter distribution of DiTs, existing pruning methods are inapplicable, leading to significant performance degradation, especially at high sparsity rates.

\begin{figure}[t]
  \begin{center}
    \centerline{\includegraphics[width=\columnwidth]{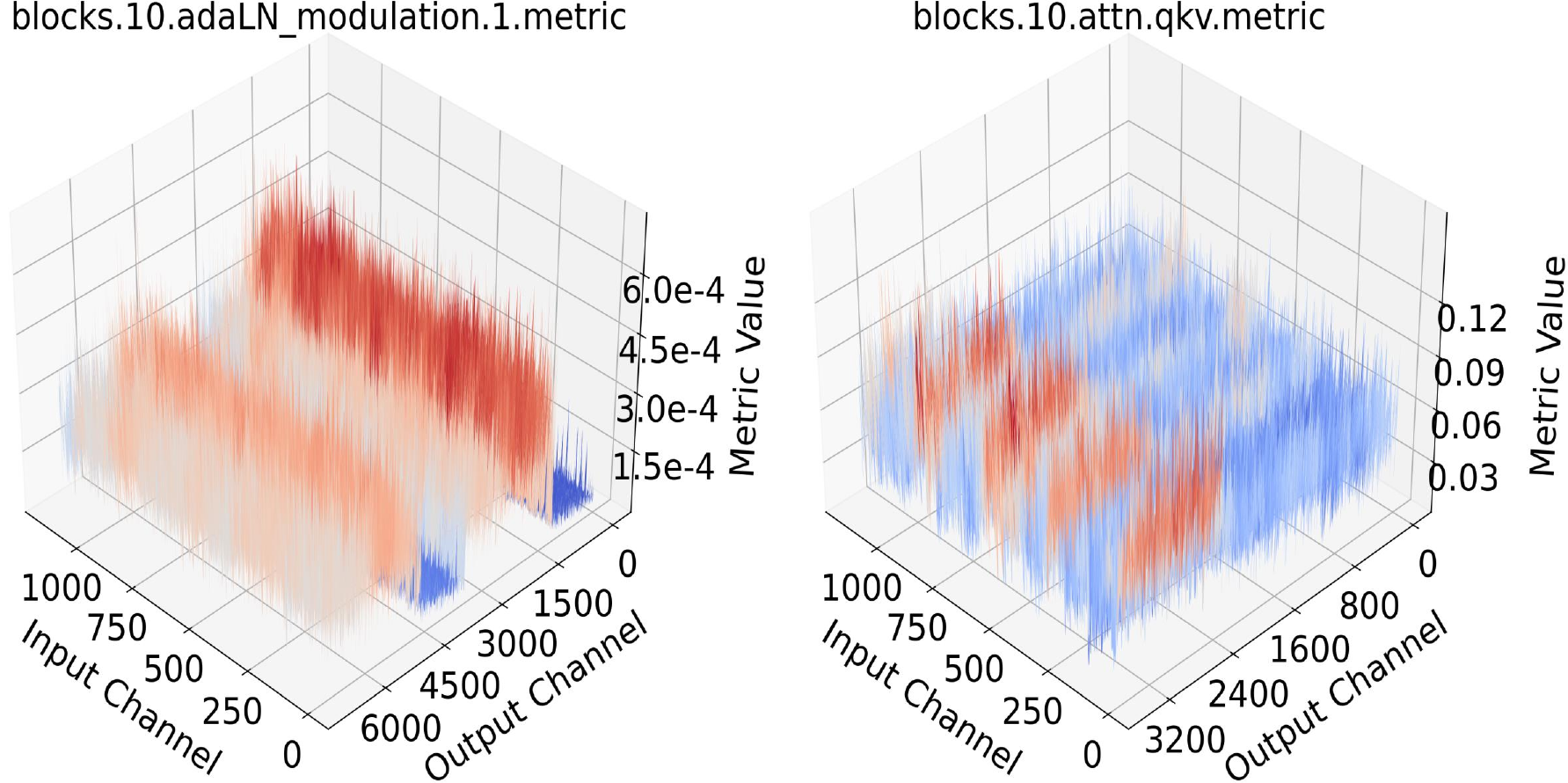}} 
    \caption{
       The distribution characteristics of parameter importance in DiTs, which exhibits distinct clustering patterns in the two-dimensional weight space of DiTs.
    }
    \label{3D}
  \end{center}
\end{figure}

To this end, we thoroughly analyze the characteristics of DiTs from the perspectives of saliency criteria and pruning granularity. Regarding saliency criteria, many existing pruning methods developed for LLMs, represented by Wanda~\citep{sun2023simple}, construct efficient pruning sensitivity estimations based on second-order approximations, achieving remarkable performance. However, these methods have the following limitations: Theoretically, the approximations adopted amplify weight contributions, disrupting the balance with activations in sensitivity estimation; empirically, we observe that DiT and LLM exhibit fundamentally different statistical properties in the relative magnitude of weights and activations. As shown in Figure~\ref{weights}, weights in DiTs reach significantly larger magnitudes, leading to a mismatch between weight and activation scales. This discrepancy further exacerbates errors introduced by the approximations. Furthermore, as shown in Figure~\ref{3D}, we observe that the metric demonstrates distinct clustering patterns in the two-dimensional weight space, indicating that DiTs are incompatible with uniform pruning granularity.

In this work, we propose DiT-Pruning, an efficient post-training pruning method for DiTs, which improves pruning performance by introducing customized saliency criteria and pruning granularity. Specifically, we apply a squared transformation on weights to mitigate the amplification effect dominated by weight magnitudes in sensitivity estimation, which alleviates the scale disparity, facilitating accurate identification of the important elements. To address the clustering patterns observed in saliency criteria, we exploit the distributional structure to dynamically adjust pruning granularity. Per-layer and per-channel strategies each demonstrate advantages under different clustering patterns, enabling more effective and robust sparsity allocation. Our main contributions are summarized as follows:

\begin{itemize}
 \item We propose DiT-Pruning, a post-training pruning method for DiTs, which effectively improves pruning performance at high sparsity utilizing parameter saliency estimation and pruning granularity.
 \item We modulate the saliency metric from an energy-based modeling perspective, providing a stronger theoretical foundation and better alignment with DiTs' properties. Moreover, we propose a clustering-aware pruning granularity to facilitate alignment with distribution patterns.
 \item We conduct extensive evaluations on DiTs, demonstrating that our method consistently preserves image quality under high sparsity. On FLUX.1-dev at 512$\times$512 resolution on MJHQ, we achieve only a \emph{0.001 loss} in CLIP score at 50\% sparsity, outperforming exsisting methods.
\end{itemize}
\section{Related Works}

\begin{figure*}[htb]
  \begin{center}
    \centerline{\includegraphics[width=0.95\textwidth]{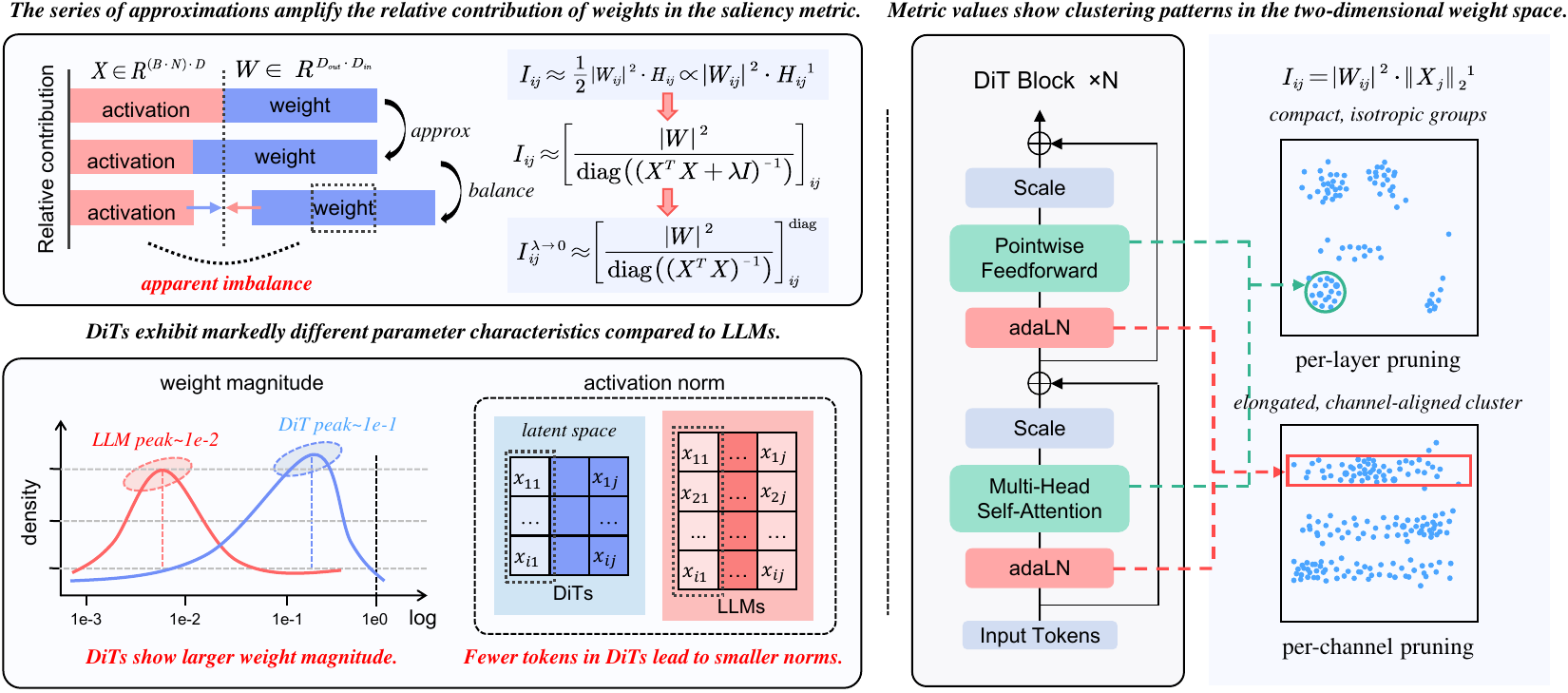}} 
    \caption{
      Overview. \textbf{(Left)} Existing saliency metrics amplify the relative contribution of weights due to a series of approximations. We introduce a squared transformation on the weights (STW) to balance them. \textbf{(Right)} The importance scores exhibit clear clustering patterns in the two-dimensional weight space, enabling our clustering-aware granularity (CAG) for sparse allocation.
    }
    \label{Overview}
  \end{center}
\end{figure*}

\textbf{Diffusion Transformers.} Diffusion Transformers~\citep{peebles2023scalable} (DiTs) have attracted increasing attention and gained great popularity by replacing convolutional U-NeTs~\citep{ronneberger2015u} with transformer-based architectures~\citep{vaswani2017attention} for image generation. DiTs demonstrate state-of-the-art performance and are gradually emerging as a mainstream backbone for generative applications. Models like Pixart~\citep{chen2023pixart}, Flux~\citep{labs2025flux} and SANA~\citep{xie2025sana} further extend DiTs to text-to-image tasks, highlighting their strong capability and scalability to model high-resolution images under complex semantic conditions. However, these advances come at the cost of rapidly growing parameter sizes and computational complexity, leading to substantial resource consumption.

\noindent\textbf{Post-Training Pruning.} Post-training pruning is a widely adopted model compression technique~\citep{han2015deep,kwon2022fast} that aims to directly operate on pre-trained models without introducing additional training or fine-tuning costs. By identifying and removing redundant or less important weight parameters, post-training pruning produces a compact model representation, thereby reducing model sizes and computational overhead in practical deployment, while maintaining competitive performance.

Magnitude pruning~\citep{han2015learning, park2020lookahead} is a classical model compression method and a strong baseline~\citep{blalock2020state} in network pruning. It prunes parameters with small absolute values by ranking weights based on magnitude. However, in complex network architectures, parameter importance is influenced by multiple factors, further undermining the reliability of magnitude-based criteria.

Prior works like CNN pruning~\citep{molchanov2016pruning} combine gradients when designing saliency metrics. Similar criteria~\citep{michel2019sixteen} have been adopted to prune attention heads in ViTs. Recent advanced methods have been proposed for LLMs, such as SparseGPT~\citep{frantar2023sparsegpt} and Wanda~\citep{sun2023simple} , which focus on efficient unstructured pruning. Furthermore, Llm-pruner~\citep{ma2023llm}, SlimGPT~\citep{ling2024slimgpt} and  SoBP~\citep{wei2024structured} explore structured pruning. In contrast, pruning methods for diffusion models remain underexplored. Diff-Pruning~\citep{fang2023structural} target U-Net-based~\citep{ronneberger2015u} diffusion models, which introduces a timestep-aware dynamic pruning but requires retraining. Due to DiTs' unique architectural design and parameter distribution, these methods are inapplicable, and resulting in significant performance degradation. This highlights the need for efficient post-training pruning methods specifically designed for DiTs.

\section{Method}
In this section, we thoroughly analyze the characteristics of DiTs from the perspectives of saliency criteria and pruning granularity and propose a novel post-training pruning method which effectively improves pruning performance. The overview of our method is shown in Figure~\ref{Overview}.

\subsection{Analysis of the OBD-based Criteria}
Classical post-training pruning methods primarily stems from Optimal Brain Damage (OBD)~\citep{lecun1989optimal} and Optimal Brain Surgeon (OBS)~\citep{hassibi1993optimal}, which employ second-order sensitivity analysis to estimate the loss variation caused by pruning individual weights, thereby achieving unstructured sparsity.
\begin{equation}
 I_i \approx \frac{1}{2} h_{ii}  w_i ^2
 \label{OBD}
\end{equation}

OBD are shown in Eq.~\eqref{OBD}, where $I_i$ denotes the importance of the $i_{th}$ weight parameter, defined as the approximate increase in the loss function when this weight is set to zero. $w_i $ denotes the magnitude of pruned weights, $h_{ii}$ denotes the $i_{th}$ diagonal element of the Hessian matrix($\mathbf{H}$) of the loss function. OBS are shown in Eq.~\eqref{OBS}, which further extends OBD. The formulations explicitly show that pruning sensitivity is jointly determined by both the weight magnitude and local curvature information, providing a principled second-order criterion for network sparsification.
\begin{equation}
I_i \approx \frac{1}{2} \frac { w_i ^2} {(\mathbf{H}^{-1})_{ii}}
\label{OBS}
\end{equation}

Based on these principles, a variety of pruning methods have been proposed. However, most of them are tailored to LLM architectures and become ineffective when applied to DiTs. In the following, we provide a detailed analysis from both theoretical and empirical perspectives to illustrate.

Subsequent works, such as SparseGPT~\citep{frantar2023sparsegpt} and Wanda~\citep{sun2023simple}, inherit from OBD and OBS framework and extend to the post-training pruning for LLMs by approximating the Hessian with activation-based second-order statistics under strong independence assumptions. This formulation enables the pruning metric to explicitly account for the combined influence of weight terms and activation terms on pruning sensitivity estimation. To achieve computational efficiency, these methods further adopt approximations, which reduce the accuracy of sensitivity estimation and alter the relative contributions of weight and activation terms.
\begin{equation}
I_{ij} \approx \left[ \frac{|\mathbf{W}|^2}{\mathrm{diag}\big((\mathbf{X}^\top \mathbf{X} + \lambda \mathbf{I})^{-1}\big)} \right]_{ij}
 \label{SparseGPT}
\end{equation}

Specifically, methods like SparseGPT formulates the pruning sensitivity as shown in Eq.~\eqref{SparseGPT} by employing a diagonal Hessian approximation with reconstruction compensation mechanism. In this equation, $\mathbf{W}\in\mathbb{R}^{d_{\text{out}}\times d_{\text{in}}}$ denotes the weight matrix, $\mathbf{X}\in\mathbb{R}^{N\times d_{\text{in}}}$ denotes the input activation matrix, where each row corresponds to a token embedding, and $\mathbf{X}^\top\mathbf{X}$ serves as an empirical Hessian approximation. The diagonal operator $\mathrm{diag}(\cdot)$ retains only the diagonal elements, and the damping term $\lambda\mathbf{I}$ is introduced for numerical stability and partially mitigates the bias caused by the diagonal approximation, but it remains imperfect.
\begin{equation}
I_{ij}^{\lambda \to 0} 
\approx \left[ \frac{|\mathbf{W}|^2}{\mathrm{diag}((\mathbf{X}^\top \mathbf{X})^{-1})} \right]_{ij}^{\text{diag approx}} 
= \left( |\mathbf{W}_{ij}| \cdot \|\mathbf{X}_j\|_2 \right)^2  
 \label{wanda}
\end{equation}

Whereas Wanda in Eq.~\eqref{wanda} further simplifies the estimation by discarding the reconstruction-aware optimization and taking the limit $\lambda \rightarrow 0$, leading to a diagonally approximated, first-order surrogate. Here, $|\mathbf{W}_{ij}|$ denotes the weight connecting the $j$-th input channel to the $i$-th output channel, and $||\mathbf{X}_j||_2$ represents the $\ell_2$ norm of the $j$-th input activation across tokens. The diagonal approximation of $(\mathbf{X}^\top \mathbf{X})^{-1}$  collapses the curvature information into a channel-wise scaling factor, yielding a multiplicative form composed of weight magnitude and activation norm.

Consequently, these approximations operate primarily on the activation terms, reducing their contribution to a bounded scaling factor. Meanwhile, the weight term is no longer modulated by the inverse Hessian and instead directly dominates sensitivity estimation, thus amplifying its contribution.

\subsection{Pruning Metric}
\begin{figure*}[htb]
  \begin{center}
    \centerline{\includegraphics[width=0.95\textwidth]{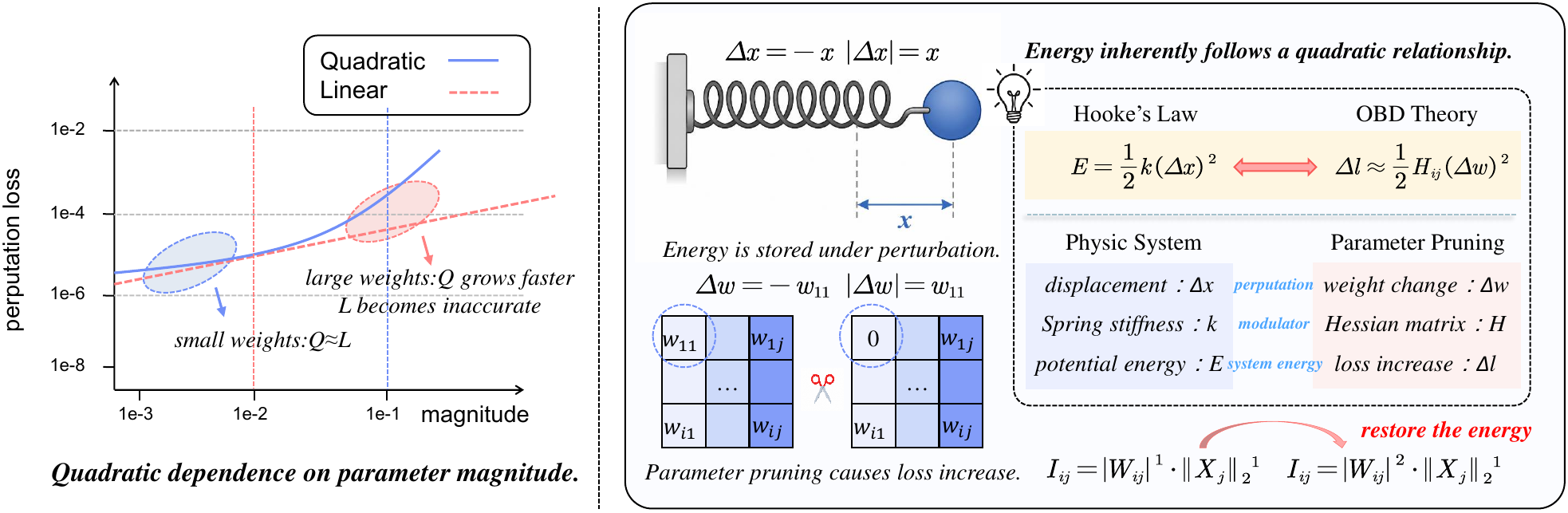}} 
    \caption{
    \textbf{(Left)} Pruning loss increase grows quadratically with parameter magnitude. \textbf{(Right)} Analogous to elastic potential energy, loss increase can be interpreted as perturbation energy, we square the weights to restore the quadratic energy structure.
    }
    \label{energy}
  \end{center}
\end{figure*}

We revisit the design of importance metrics from an energy-based perspective, as illustrated in Figure~\ref{energy}(Right). We start from the observation that parameter pruning can be interpreted as introducing a perturbation to the model, where the resulting loss increase can be formalized via a Taylor expansion. In classical mechanics, elastic potential energy grows quadratically with displacement, where the stored energy is determined by second-order interactions of the system,  which motivates modeling the loss increase as a quadratic energy form governed by the Hessian. In this view, weight importance is treated as an energy allocation problem, where each parameter contributes proportionally to its squared magnitude, weighted by local curvature. This reveals that existing importance metrics, which rely on linear estimations of weight perturbations, inherently ignore second-order effects.

Actually, as shown in Figure~\ref{energy}(Left), such linearization fails to capture the intrinsic quadratic energy structure of the loss landscape, leading to systematic underestimation of relative large-magnitude weights. In particular, linear estimations remain locally accurate in the small-weight regime but increasingly deviate as weight magnitude grows. Moreover, this discrepancy becomes more pronounced as weight magnitude increases, reflecting the important role of second-order effects in large parameter regimes.

This issue is particularly severe in DiTs due to their distinct scale characteristics compared to LLMs. As shown in Figure~\ref{Overview}(Left), DiT weights exhibit a wider range and overall larger magnitudes, which leads to the breakdown of linear estimations. Besides, $\mathbf{X}\in\mathbb{R}^{N\times d_{\text{in}}}$ denotes the input activation matrix, $N$ denotes the number of tokens, since DiTs operate in the latent space with significantly fewer tokens (e.g., $N=256$ or $512$), compared to LLMs with long token sequences (e.g., $N=2048$, $4096$ or $8192$), the resulting $\|\mathbf{X}_j\|_2$ values are substantially smaller. Consequently, the scale imbalance further amplifies the relative dominance of weight terms, making the approximation imbalance more apparent and structurally biased in DiTs. 

To retain the computational efficiency of existing approximations, we adopt $|\mathbf{W}_{ij}|$ and $|\mathbf{X}_j|_2$ as the fundamental weight and activation components. Following classical second-order pruning criteria, we preserve the separable multiplicative structure and introduce exponents $(m,n)$ to parameterize contributions, as shown in Eq.~\eqref{eq:parametric_form}. Since the scaling factors $a$ and $b$ do not affect within-layer ranking, they are omitted, resulting in the simplified form in Eq.~\eqref{eq:parametric_form_1}.
\begin{equation}
I_{ij}
=
\left( a \cdot |\mathbf{W}_{ij}| \right)^{m}
\cdot
\left( b \cdot \|\mathbf{X}_j\|_2 \right)^{n}
\label{eq:parametric_form}
\end{equation}
 \begin{equation}
I_{ij}
=
\left(  |\mathbf{W}_{ij}| \right)^{m}
\cdot
\left(   \|\mathbf{X}_j\|_2 \right)^{n}
\label{eq:parametric_form_1}
\end{equation}
 \begin{equation}
I_{ij}
=
\left(  |\mathbf{W}_{ij}| \right)^{2}
\cdot
\left(   \|\mathbf{X}_j\|_2 \right)^{1}
\label{eq:parametric_form_2}
\end{equation}

Motivated by this observation, we instantiate our parametric formulation by applying a squared transformation to the weight term \textbf{(STW)}, corresponding to $(m=2)$ and $(n=1)$ in Eq.~\eqref{eq:parametric_form_1}, as defined in Eq.~\eqref{eq:parametric_form_2}. This design restores the quadratic energy structure of perturbation-induced loss variations and rebalances the relative contributions of weights and activations, while preserving the original power relationship, yielding a more faithful and reliable sensitivity estimation. Meanwhile, \textbf{STW} compresses the dominant DiT weight scale from $10^{-1}$ to $10^{-2}$, bringing it closer to the regime where Wanda was originally developed for LLMs (Appendix A shows detailed statistics).

\subsection{Pruning Granularity}
\label{Pruning granularity}
 In general, pruning removes parameters with low importance scores within predefined structural groups. As shown in Eq.~\eqref{mask}, $\mathbf{W}$ and $\widetilde{\mathbf{W}}$ denote the original and pruned weight matrices, respectively, while $\mathbf{M}_{\mathcal{G}} \in {0,1}^{|\mathbf{W}|}$ is a group-wise binary mask determined by the predefined pruning groups $\mathcal{G}$. These groups define the comparison scope for ranking importance scores, making the choice of pruning granularity critical to the final sparsification performance.
\begin{equation}
\widetilde{\mathbf{W}} 
= 
\mathbf{W} \odot \mathbf{M}_{\mathcal{G}}
\label{mask}
\end{equation}

Existing post-training pruning methods typically adopt a uniform granularity, overlooking variations in weight distributions and model structures. In contrast, we observe that our metric exhibits distinct clustering patterns in the two-dimensional weight space of DiTs, as illustrated in Figure~\ref{Overview}(Right). Specifically, some weights form elongated, channel-aligned clusters, while others aggregate into compact, isotropic groups. These observations reveal substantial heterogeneity in importance distributions across layers and channels, suggesting that a globally uniform pruning granularity is inherently suboptimal.
\begin{equation}
\bar{I}_i
=
\frac{1}{C_{\text{in}}}
\sum_{j=1}^{C_{\text{in}}}
I_{ij}
\label{eq:per_output_mean}
\end{equation}
\begin{equation}
\mathcal{G}_{\text{out}}
=
\frac{
\mathrm{Var}\!\left(
\left\{
\bar I_i
\right\}_{i=1}^{C_{\text{out}}}
\right)
}{
\left(
\mathrm{Mean}\!\left(
\left\{
\bar I_i
\right\}_{i=1}^{C_{\text{out}}}
\right)
\right)^2
}
\label{eq:pnv}
\end{equation}

 Motivated by the insight, we propose a clustering-aware pruning granularity (CAG) that dynamically adjusts comparison groups according to the underlying importance distribution. Specifically, we use the averaged channel importance $\bar{I}_i$ and the normalized heterogeneity metric $\mathcal{G}_{\text{out}}$ in Eq.~\eqref{eq:per_output_mean} and Eq.~\eqref{eq:pnv} to characterize layer-wise importance concentration. Layers with small $\mathcal{G}_{\text{out}}$, such as MLP and self-attention layers, exhibit compact importance patterns and are pruned at the per-layer level. In contrast, layers with large $\mathcal{G}_{\text{out}}$, notably adaLN layers, show strong channel-wise heterogeneity and are pruned at the per-channel level. This strategy enables more effective sparsity allocation and better preserves model performance under high sparsity.

\section{Experiments} 
\subsection{Experimental Settings}

\textbf{Models.} We evaluate primarily focus on three types of models: the classical pre-trained DiT-XL/2 model, the state-of-the-art PixArt-$\Sigma$ model~\citep{chen2023pixart}, and the FLUX.1-dev model~\citep{labs2025flux}. All evaluated diffusion models adopt transformer-based backbones~\citep{vaswani2017attention}.

\noindent \textbf{Datasets.} Following previous works~\citep{li2023q,zhao2024mixdq}, we randomly sample prompts from COCO~\citep{lin2014microsoft} for calibration. We evaluate DiT-XL/2 on ImageNet~\citep{russakovsky2015imagenet}, generating 10,000 images at 256$\times$256 resolution using the official DDPM sampler, consistent with prior studies~\citep{nichol2021improved,shang2023post}. Robustness is further assessed with 250, 100, and 50 diffusion steps~\citep{wu2024ptq4dit}. For PixArt-$\Sigma$ and FLUX.1-dev, we evaluate on COCO and MJHQ, generating images at 256$\times$256 and 512$\times$512 resolutions with default samplers. We use the first 1,024 COCO annotation prompts and randomly sample 1,024 prompts from MJHQ-5K, a subset of the MJHQ-30K dataset~\citep{li2024playground}.
\begin{table}[t]
\centering
\large 
\resizebox{\columnwidth}{!}{
\setlength{\tabcolsep}{4pt} 
\renewcommand{\arraystretch}{1.2} 
\begin{tabular}{c c c c c c c c c}
\toprule
\multirow{2}{*}{Timesteps} & \multirow{2}{*}{Sparsity} & \multirow{2}{*}{Method} & Params & \multirow{2}{*}{FID $\downarrow$} & \multirow{2}{*}{IS $\uparrow$} & \multirow{2}{*}{sFID $\downarrow$} & \multirow{2}{*}{PR $\uparrow$} & \multirow{2}{*}{SSIM $\uparrow$} \\
 &  &  & (M) &  &  &  &  &  \\
\midrule

\multirow{10}{*}{250}
& 0\% & Dense        & 675.13 & 5.33 & 273.62 & 17.84 & 0.84 & \textbf{--} \\
\cmidrule{2-9} 
& \multirow{3}{*}{20\%} 
& Magnitude & 541.35 & \textbf{5.17} & 272.27 & \textbf{17.35} & 0.84 & 0.788 \\
&           & Wanda     & 541.55 & 5.57 & \textbf{276.77} & 18.38 & 0.84 & 0.785 \\
&           & DiT-Pruning      & 541.43 & 5.46 & 276.26 & 17.83 & \textbf{0.84} & \textbf{0.825} \\
\cmidrule{2-9}
& \multirow{3}{*}{40\%}
& Magnitude & 407.58 & 9.39 & 163.42 & \textbf{19.81} & 0.70 & 0.545 \\
&           & Wanda     & 407.94 & 13.09 & 152.66 & 41.74 & 0.68 & 0.537 \\
&           & DiT-Pruning      & 407.73 & \textbf{6.53} & \textbf{224.27} & 23.39 & \textbf{0.79} & \textbf{0.617} \\
\cmidrule{2-9}
& \multirow{3}{*}{50\%}
& Magnitude & 340.69 & 42.06 & 51.27 & \textbf{34.91} & 0.41 & 0.474 \\
&           & Wanda     & 340.69 & 45.97 & 49.20 & 97.17 & 0.62 & 0.461 \\
&           & \cellcolor{LightBlueC}DiT-Pruning      & \cellcolor{LightBlueC}340.69 & \cellcolor{LightBlueC}\textbf{13.86} & \cellcolor{LightBlueC}\textbf{134.94} & \cellcolor{LightBlueC}36.18 & \cellcolor{LightBlueC}\textbf{0.66} & \cellcolor{LightBlueC}\textbf{0.534} \\

\midrule

\multirow{10}{*}{100}
& 0\% & Dense        & 675.13 & 5.53 & 261.87 & 18.93 & 0.81 & \textbf{--} \\
\cmidrule{2-9}
& \multirow{3}{*}{20\%}
& Magnitude & 541.35 & \textbf{5.41} & 260.92 & \textbf{18.15} & 0.82 & 0.803 \\
&           & Wanda     & 541.55 & 5.72 & 263.46 & 19.06 & 0.82 & 0.796 \\
&           & DiT-Pruning      & 541.43 & 5.56 & \textbf{263.78} & 18.61 & \textbf{0.82} & \textbf{0.835} \\
\cmidrule{2-9}
& \multirow{3}{*}{40\%}
& Magnitude & 407.58 & 12.16 & 150.41 & \textbf{22.02} & 0.66 & 0.559 \\
&           & Wanda     & 407.94 & 15.34 & 142.27 & 42.19 & 0.65 & 0.541 \\
&           & DiT-Pruning      & 407.73 & \textbf{7.19} & \textbf{213.22} & 23.58 & \textbf{0.77} & \textbf{0.627} \\
\cmidrule{2-9}
& \multirow{3}{*}{50\%}
& Magnitude & 340.69 & 50.25 & 46.79 & 40.82 & 0.36 & 0.485 \\
&           & Wanda     & 340.69 & 50.96 & 43.54 & 97.97 & 0.37 & 0.463 \\
&           & \cellcolor{LightBlueC}DiT-Pruning     & \cellcolor{LightBlueC}340.69 & \cellcolor{LightBlueC}\textbf{16.28} & \cellcolor{LightBlueC}\textbf{124.69} & \cellcolor{LightBlueC}\textbf{36.86} & \cellcolor{LightBlueC}\textbf{0.64} & \cellcolor{LightBlueC}\textbf{0.541} \\

\midrule

\multirow{10}{*}{50}
& 0\% & Dense        & 675.13 & 6.68 & 236.27 & 21.62 & 0.78 & \textbf{--} \\
\cmidrule{2-9}
& \multirow{3}{*}{20\%}
& Magnitude & 541.35 & \textbf{6.38} & 237.56 & \textbf{20.43} & 0.79 & 0.823 \\
&           & Wanda     & 541.55 & 6.57 & 238.33 & 20.78 & 0.79 & 0.813 \\
&           & DiT-Pruning      & 541.43 & 6.41 & \textbf{242.23} & 20.68 & \textbf{0.79} & \textbf{0.853} \\
\cmidrule{2-9}
& \multirow{3}{*}{40\%}
& Magnitude & 407.58 & 17.29 & 131.89 & 26.50 & 0.62 & 0.584 \\
&           & Wanda     & 407.94 & 19.20 & 122.57 & 43.60 & 0.62 & 0.555 \\
&           & DiT-Pruning      & 407.73 & \textbf{8.78} & \textbf{194.21} & \textbf{24.50} & \textbf{0.75} & \textbf{0.646} \\
\cmidrule{2-9}
& \multirow{3}{*}{50\%}
& Magnitude & 340.69 & 62.40 & 37.98 & 50.50 & 0.32 & 0.508 \\
&           & Wanda     & 340.69 & 57.32 & 38.84 & 101.30 & 0.34 & 0.473 \\
&           & \cellcolor{LightBlueC}DiT-Pruning      & \cellcolor{LightBlueC}340.69 & \cellcolor{LightBlueC}\textbf{20.29} & \cellcolor{LightBlueC}\textbf{111.95} & \cellcolor{LightBlueC}\textbf{38.95} & \cellcolor{LightBlueC}\textbf{0.60} & \cellcolor{LightBlueC}\textbf{0.559} \\

\bottomrule
\end{tabular}
}
\caption{Quantitative comparison on DiT-XL/2 at 256$\times$256 resolution on ImageNet under different timesteps and sparsity levels. The best result per metric is highlighted in bold.}
\label{tab:dit_imagenet}
\end{table}

\noindent \textbf{Baselines.} We compare our method with two representative approaches. Magnitude pruning~\citep{han2015learning, park2020lookahead} is a classical pruning strategy and serves as a strong baseline in our experiments. Wanda~\citep{sun2023simple} is a representative second-order pruning method originally designed for LLMs. To construct the calibration data, we collect 128 input samples from the model at randomly sampled diffusion timesteps~\citep{chen2025q}.

\noindent \textbf{Metrics.} To comprehensively assess generated image quality, we employ the following metrics: We choose Fréchet Inception Distance (FID)~\citep{heusel2017gans} for fidelity evaluation, spatial FID (sFID)~\citep{nash2021generating, NIPS2016_8a3363ab} for spatial consistency, Inception Score (IS)~\citep{barratt2018note} for visual quality and diversity, Clipscore (CLIP)~\citep{hessel2021clipscore} for text-image alignment, ImageReward (IR)~\citep{xu2023imagereward} for human preference, Precision (PR)~\citep{kynkaanniemi2019improved} for sample realism, SSIM~\citep{wang2004image} for structural coherence. All experiments are conducted under identical experiment settings on NVIDIA RTX A6000 GPUs.

\subsection{Pruning Performance}  
We present a comprehensive assessment of our method against prevalent baseline methods in various settings.

\textbf{DiT-XL/2 on ImageNet.} Table~\ref{tab:dit_imagenet} reports the performance of pruned DiT-XL/2 model. At 20\% sparsity, a mild pruning, all methods yield similar outcomes. As sparsity increase to 40\%, our method consistently outperforms the baselines and remains close to the dense model, suggesting the existence of highly effective sparse sub-networks within DiTs. The advantage becomes more pronounced at 50\% sparsity, where baseline methods suffer substantial degradation while our method maintains strong generative performance. For example, under 250 sampling steps and 50\% sparsity, magnitude pruning and Wanda increase FID by 36.73 and 40.64, respectively, whereas our method incurs only an 8.53 increase, demonstrating the effectiveness under aggressive sparse.
\begin{table}[t]
\centering
\large
\resizebox{\columnwidth}{!}{
\setlength{\tabcolsep}{4pt}
\renewcommand{\arraystretch}{1.4}
\begin{tabular}{c c c c c c c c c c c}
\toprule
\multirow{2}{*}{Resolution} & \multirow{2}{*}{Sparsity} & \multirow{2}{*}{Method} & Params & \multirow{2}{*}{FID $\downarrow$} & \multirow{2}{*}{IS $\uparrow$} & \multirow{2}{*}{CLIP $\uparrow$}
& \multirow{2}{*}{IR $\uparrow$} & \multirow{2}{*}{sFID $\downarrow$} & \multirow{2}{*}{PRE $\uparrow$} & \multirow{2}{*}{SSIM $\uparrow$} \\ 
 &  &  &  (M)  &  &  &  &  &  &  &\\
\midrule

\multirow{10}{*}{256$\times$256}
& 0\%  & Dense    & 612.08 & 70.31 & 33.33 & 0.265 & 0.807 & 249.59 & 0.49 & \textbf{--} \\
\cmidrule{2-11}

& \multirow{3}{*}{20\%}
& Magnitude & 493.42 & 69.44 & \textbf{32.92} & \textbf{0.266} & \textbf{0.821} & 249.16 & 0.49 & 0.543 \\
&         & Wanda     & 493.64 & \textbf{68.60} & 30.62 & 0.264 & 0.727 & \textbf{248.03} & 0.50 & 0.631 \\
&         & DiT-Pruning      & 493.46 & 69.01 & 31.84 & 0.265 & 0.793 & 249.52 & \textbf{0.51} & \textbf{0.688} \\
\cmidrule{2-11}

& \multirow{3}{*}{40\%}
& Magnitude & 374.48 & 69.87 & \textbf{31.36} & 0.258 & 0.498 & 239.25 & 0.44 & 0.407 \\
&         & Wanda     & 374.87 & 69.32 & 28.86 & 0.256 & 0.244 & \textbf{235.79} & 0.47 & 0.456 \\
&         & DiT-Pruning      & 374.55 & \textbf{67.50} & 31.06 & \textbf{0.260} & \textbf{0.635} & 244.21 & \textbf{0.51} & \textbf{0.484} \\
\cmidrule{2-11}

& \multirow{3}{*}{50\%}
& Magnitude & 315.02 & 92.70 & 21.03 & 0.231 & -0.728 & 240.16 & 0.23 & 0.312 \\
&         & Wanda     & 315.10 & 89.66 & 20.47 & 0.229 & -0.923 & \textbf{234.46} & 0.27 & 0.381 \\
&         & \cellcolor{LightBlueC}DiT-Pruning      & \cellcolor{LightBlueC}315.10 & \cellcolor{LightBlueC}\textbf{69.79} & \cellcolor{LightBlueC}\textbf{29.76} & \cellcolor{LightBlueC}\textbf{0.251} & \cellcolor{LightBlueC}\textbf{0.183} & \cellcolor{LightBlueC}237.60 & \cellcolor{LightBlueC}\textbf{0.42} & \cellcolor{LightBlueC}\textbf{0.385} \\

\midrule

\multirow{10}{*}{512$\times$512}
& 0\%  & Dense    & 612.08 & 62.36 & 36.12 & 0.256 & 0.929 & 274.47 & 0.59 & \textbf{--} \\
\cmidrule{2-11}

& \multirow{3}{*}{20\%}
& Magnitude & 494.29 & 61.25 & \textbf{37.76} & 0.257 & \textbf{0.968} & \textbf{268.88} & 0.59 & 0.484 \\
&         & Wanda     & 494.51 & \textbf{61.22} & 36.92 & 0.255 & 0.879 & 269.71 & 0.60 & 0.674 \\
&         & DiT-Pruning      & 494.34 & 61.71 & 35.78 & \textbf{0.257} & 0.942 & 272.10 & \textbf{0.60} & \textbf{0.693} \\
\cmidrule{2-11}

& \multirow{3}{*}{40\%}
& Magnitude & 375.35 & 61.12 & 34.87 & 0.252 & 0.587 & 257.66 & 0.53 & 0.378 \\
&         & Wanda     & 375.75 & 61.26 & 33.33 & 0.247 & 0.360 & \textbf{256.87} & 0.57 & 0.475 \\
&         & DiT-Pruning      & 375.43 & \textbf{56.74} & \textbf{35.98} & \textbf{0.253} & \textbf{0.803} & 264.47 & \textbf{0.62} & \textbf{0.514} \\
\cmidrule{2-11}

& \multirow{3}{*}{50\%}
& Magnitude & 315.90 & 108.86 & 19.66 & 0.214 & -1.142 & 272.80 & 0.18 & 0.261 \\
&         & Wanda     & 315.97 & 93.97 & 19.81 & 0.219 & -1.143 & 263.23 & 0.29 & 0.373 \\
&         & \cellcolor{LightBlueC}DiT-Pruning      & \cellcolor{LightBlueC}315.97 & \cellcolor{LightBlueC}\textbf{61.78} & \cellcolor{LightBlueC}\textbf{33.27} & \cellcolor{LightBlueC}\textbf{0.244} & \cellcolor{LightBlueC}\textbf{0.285} & \cellcolor{LightBlueC}\textbf{254.82} & \cellcolor{LightBlueC}\textbf{0.51} & \cellcolor{LightBlueC}\textbf{0.439} \\

\bottomrule
\end{tabular}
}
\caption{Image quality and text-image alignment results of PixArt-$\Sigma$ on COCO across different resolutions and sparsity.}
\label{tab:pixart_coco}
\end{table}

\textbf{PixArt-$\Sigma$ on COCO.} Table~\ref{tab:pixart_coco} presents the results of pruning PixArt-$\Sigma$ on COCO. Our method consistently achieves competitive or superior performance in both image quality and text-image alignment. At 40\% sparsity, it attains the best FID and CLIP scores, indicating strong preservation of visual quality and semantic fidelity. The advantage becomes more pronounced at 50\% sparsity, suggesting that the proposed method maintains performance better as sparsity increases.

\textbf{FLUX.1-dev on COCO and MJHQ.} Table~\ref{tab:flux_256} and~\ref{tab:flux_512} report the results on FLUX.1-dev across COCO and MJHQ.  Similar to the observations on PixArt-$\Sigma$, all methods perform comparably at 20\% and 40\% sparsity, with only minor differences in image quality and text-image alignment metrics. As sparsity increases to 50\%, the performance gap becomes substantially larger, where our method consistently achieves superior results and demonstrates stronger robustness.

\subsection{Ablation Study}

To validate the effectiveness of our method, we conduct ablation studies on DiT-XL/2. We further evaluate FLUX.1-dev under structured sparsity on MJHQ at different resolutions to verify the effectiveness of \textbf{STW}. Starting from Wanda as the baseline, we incrementally incorporate  \textbf{STW} and \textbf{CAG}, and evaluate their individual contributions. As shown in Tables~\ref{tab:ablation1} and~\ref{tab:ablation2}, each component improves performance and their combination achieves the best. Appendix E shows more results.
\begin{table}[h]
\centering
\large
\setlength{\tabcolsep}{4pt}
\renewcommand{\arraystretch}{1.3}
\resizebox{\columnwidth}{!}{
\begin{tabular}{c c c c c c c c c c c}
\toprule
\multirow{2}{*}{Dataset} & \multirow{2}{*}{Sparsity} & \multirow{2}{*}{Method} &Params
& \multirow{2}{*}{FID $\downarrow$} & \multirow{2}{*}{IS $\uparrow$} & \multirow{2}{*}{CLIP $\uparrow$}
& \multirow{2}{*}{IR $\uparrow$} & \multirow{2}{*}{sFID $\downarrow$} & \multirow{2}{*}{PRE $\uparrow$} & \multirow{2}{*}{SSIM $\uparrow$} \\
 &  &  &  (B)  &  &  &  &  &  &  &\\
\midrule

\multirow{10}{*}{MJHQ}
& 0\% & Dense & 11.9
& 54.94 & 26.11 & 0.260 & 0.686 & 259.18 & 0.75 & \textbf{--} \\
\cmidrule{2-11}

& \multirow{3}{*}{20\%}
& Magnitude & 9.53 & 57.05 & 24.32 & 0.257 & 0.607 & 262.01 & 0.73 & 0.692 \\
&           & Wanda     & 9.54 & \textbf{54.48} & 25.61 & 0.259 & 0.681 & \textbf{259.97} & 0.74 & 0.786 \\
&           & DiT-Pruning      & 9.53 & 54.54 & \textbf{26.37} & \textbf{0.259} & \textbf{0.689} & 260.67 & \textbf{0.74} & \textbf{0.836} \\
\cmidrule{2-11}

& \multirow{3}{*}{40\%}
& Magnitude & 7.16 & 64.56 & 20.00 & 0.246 & 0.251 & 267.40 & 0.63 & 0.499 \\
&           & Wanda     & 7.17 & 57.28 & 23.83 & 0.257 & 0.582 & 256.94 & 0.69 & 0.584 \\
&           & DiT-Pruning      &7.17 & \textbf{54.83} & \textbf{23.64} & \textbf{0.259} & \textbf{0.647} & \textbf{255.83} & \textbf{0.71} & \textbf{0.663} \\
\cmidrule{2-11}

& \multirow{3}{*}{50\%}
& Magnitude & 5.97 & 74.38 & 16.30 & 0.235 & -0.384 & 273.47 & 0.49 & 0.422 \\
&           & Wanda     &5.98 & 62.43 & 19.00 & 0.256 & 0.281 & \textbf{255.13} & 0.58 & 0.477 \\
&           & \cellcolor{LightBlueC}DiT-Pruning     &\cellcolor{LightBlueC}5.98  & \cellcolor{LightBlueC}\textbf{56.78} & \cellcolor{LightBlueC}\textbf{23.51} & \cellcolor{LightBlueC}\textbf{0.257} & \cellcolor{LightBlueC}\textbf{0.517} & \cellcolor{LightBlueC}256.29 & \cellcolor{LightBlueC}\textbf{0.69} & \cellcolor{LightBlueC}\textbf{0.567} \\

\midrule

\multirow{10}{*}{COCO}
& 0\% & Dense & 11.9
& 66.01 & 38.78 & 0.154 & -0.964 & 295.08 & 0.73 & \textbf{--} \\
\cmidrule{2-11}

& \multirow{3}{*}{20\%}
& Magnitude & 9.53 & 66.54 & 37.26 & 0.154 & -1.037 & 296.29 & 0.68 & 0.675 \\
&           & Wanda     &9.54 & 65.62 & 37.57 & 0.154 & -0.991 & \textbf{293.87} & 0.72 & 0.784 \\
&           & DiT-Pruning      &9.53 & \textbf{65.31} & \textbf{37.69} & \textbf{0.154} & \textbf{-0.965} & 294.94 & \textbf{0.72} & \textbf{0.828} \\
\cmidrule{2-11}

& \multirow{3}{*}{40\%}
& Magnitude & 7.16 & 71.11 & 30.49 & 0.155 & -1.227 & 308.19 & 0.57 & 0.493 \\
&           & Wanda     & 7.17 & 68.27 & 36.15 & 0.155 & -1.049 & 294.71 & 0.61 & 0.561 \\
&           & DiT-Pruning      & 7.17 & \textbf{64.96} & \textbf{37.71} & \textbf{0.156} & \textbf{-0.992} & \textbf{292.77} & \textbf{0.69} & \textbf{0.645} \\
\cmidrule{2-11}

& \multirow{3}{*}{50\%}
& Magnitude & 5.97& 78.46 & 24.14 & 0.159 & -1.476 & 317.77 & 0.42 & 0.434 \\
&           & Wanda     & 5.98 & 74.85 & 27.87 & 0.155 & -1.179 & \textbf{287.41} & 0.44 & 0.452 \\
&           & \cellcolor{LightBlueC}DiT-Pruning      & \cellcolor{LightBlueC}5.98 & \cellcolor{LightBlueC}\textbf{67.56} & \cellcolor{LightBlueC}\textbf{35.91} & \cellcolor{LightBlueC}\textbf{0.161} & \cellcolor{LightBlueC}\textbf{-1.089} & \cellcolor{LightBlueC}295.79 & \cellcolor{LightBlueC}\textbf{0.60} & \cellcolor{LightBlueC}\textbf{0.537} \\

\bottomrule
\end{tabular}
}
\caption{Evaluation of FLUX.1-dev on COCO and MJHQ at 256$\times$256 resolution under increasing sparsity.}
\label{tab:flux_256}
\medskip
\large
\setlength{\tabcolsep}{4pt}
\renewcommand{\arraystretch}{1.3}

\resizebox{\columnwidth}{!}{
\begin{tabular}{c c c c c c c c c c c}
\toprule
\multirow{2}{*}{Dataset} & \multirow{2}{*}{Sparsity} & \multirow{2}{*}{Method} & Params
& \multirow{2}{*}{FID $\downarrow$} & \multirow{2}{*}{IS $\uparrow$} & \multirow{2}{*}{CLIP $\uparrow$}
& \multirow{2}{*}{IR $\uparrow$} & \multirow{2}{*}{sFID $\downarrow$} & \multirow{2}{*}{PRE $\uparrow$} & \multirow{2}{*}{SSIM $\uparrow$} \\
 &  &  &  (B)  &  &  &  &  &  &  &\\
\midrule

\multirow{10}{*}{MJHQ}
& 0\% & Dense & 11.9
& 48.64 & 29.56 & 0.272 & 0.899 & 266.47 & 0.78 & \textbf{--} \\
\cmidrule{2-11}

& \multirow{3}{*}{20\%}
& Magnitude & 9.53 & 50.76 & 26.74 & 0.271 & 0.867 & 270.15 & 0.76 & 0.731 \\
&           & Wanda     & 9.54 & \textbf{48.82} & 26.98 & 0.272 & \textbf{0.921} & \textbf{265.42} & 0.77 & 0.801 \\
&           & DiT-Pruning      & 9.53 & 48.93 & \textbf{28.75} & \textbf{0.273} & 0.913 & 265.72 & \textbf{0.78} & \textbf{0.847} \\
\cmidrule{2-11}

& \multirow{3}{*}{40\%}
& Magnitude & 7.16 & 57.36 & 22.32 & 0.262 & 0.635 & 269.75 & 0.71 & 0.561 \\
&           & Wanda     & 7.17 & 51.98 & 23.66 & 0.271 & 0.895 & 267.40 & 0.75 & 0.624 \\
&           & DiT-Pruning      & 7.17 & \textbf{49.56} & \textbf{27.07} & \textbf{0.272} & \textbf{0.899} & \textbf{265.03} & \textbf{0.77} & \textbf{0.692} \\
\cmidrule{2-11}

& \multirow{3}{*}{50\%}
& Magnitude & 5.97 & 66.03 & 19.77 & 0.249 & 0.133 & 269.02 & 0.57 & 0.493 \\
&           & Wanda     & 5.98 & 61.44 & 18.73 & 0.265 & 0.684 & 270.97 & 0.62 & 0.504 \\
&           & \cellcolor{LightBlueC}DiT-Pruning      & \cellcolor{LightBlueC}5.98 & \cellcolor{LightBlueC}\textbf{52.42} & \cellcolor{LightBlueC}\textbf{24.10} & \cellcolor{LightBlueC}\textbf{0.271} & \cellcolor{LightBlueC}\textbf{0.828} & \cellcolor{LightBlueC}\textbf{264.81} & \cellcolor{LightBlueC}\textbf{0.76} & \cellcolor{LightBlueC}\textbf{0.603} \\

\midrule

\multirow{10}{*}{COCO}
& 0\% & Dense & 11.9
& 63.41 & 39.13 & 0.154 & -0.801 & 298.68 & 0.75 & \textbf{--} \\
\cmidrule{2-11}

& \multirow{3}{*}{20\%}
& Magnitude & 9.53 & 63.82 & 36.45 & 0.155 & \textbf{-0.811} & 300.94 & \textbf{0.74} & 0.717 \\
&           & Wanda     & 9.54 & 63.58 & 37.37 & 0.155 & -0.812 & 299.05 & 0.72 & 0.793 \\
&           & DiT-Pruning      & 9.54 & \textbf{63.12} & \textbf{37.73} & \textbf{0.155} & -0.813 & \textbf{298.59} & 0.73 & \textbf{0.839} \\
\cmidrule{2-11}

& \multirow{3}{*}{40\%}
& Magnitude & 7.16 & 66.58 & 29.94 & 0.157 & -0.916 & 306.87 & 0.71 & 0.546 \\
&           & Wanda     & 7.17 & 65.73 & 34.38 & 0.156 & -0.836 & 305.19 & 0.66 & 0.616 \\
&           & DiT-Pruning      & 7.17 & \textbf{64.14} & \textbf{36.53} & \textbf{0.158} & \textbf{-0.824} & \textbf{299.66} & \textbf{0.71} & \textbf{0.684} \\
\cmidrule{2-11}

& \multirow{3}{*}{50\%}
& Magnitude & 5.97 & 69.18 & 26.77 & 0.156 & -1.139 & 308.31 & 0.58 & 0.495 \\
&           & Wanda     & 5.98 & 74.19 & 26.63 & \textbf{0.158} & -0.915 & \textbf{310.66} & 0.49 & 0.495 \\
&           & \cellcolor{LightBlueC}DiT-Pruning      & \cellcolor{LightBlueC}5.98 & \cellcolor{LightBlueC}\textbf{66.15} & \cellcolor{LightBlueC}\textbf{33.29} & \cellcolor{LightBlueC}\textbf{0.158} & \cellcolor{LightBlueC}\textbf{-0.863} & \cellcolor{LightBlueC}304.46 & \cellcolor{LightBlueC}\textbf{0.64} & \cellcolor{LightBlueC}\textbf{0.588} \\

\bottomrule
\end{tabular}
}
\caption{Evaluation of FLUX.1-dev on COCO and MJHQ at 512$\times$512 resolution under increasing sparsity.}
\label{tab:flux_512}
\end{table}

\textbf{Parameter Importance Evaluated by STW.} Under 50\% sparsity with 50  steps, STW reduces FID and sFID by 10.21 and 8.73, respectively, over the baseline under the same pruning granularity. Similar gains are observed under 2:4 and 4:8 structured sparsity on FLUX.1-dev. For example, under 2:4 sparsity at 256$\times$256 resolution, STW lowers FID from 79.35 to 69.21 while improving CLIP and IR, demonstrating its effectiveness across both unstructured and structured pruning.

\textbf{Sparse Allocation by CAG.} Furthermore, we evaluate \textbf{CAG} under the same importance metric. By dynamically adapting pruning granularity to the underlying importance distribution, \textbf{CAG} consistently improves performance. For example, as shown in Table~\ref{tab:ablation1}, at 40\% sparsity with 100 denoising steps, FID decreases from 11.79 to 7.19, while at 50\% sparsity with 50 denoising steps, it further drops from 47.11 to 20.29. These results suggest that adaptive granularity allocation enables more effective sparse allocation.
\begin{table}[t]
\centering
\scriptsize
\resizebox{\columnwidth}{!}{
\setlength{\tabcolsep}{3.5pt}
\renewcommand{\arraystretch}{0.9}
\begin{tabular}{c c !{\vrule width 0.5pt} c c c c c c}
\toprule
Timesteps & Sparsity & Ablation & FID $\downarrow$ & IS $\uparrow$ & sFID $\downarrow$ & PRE $\uparrow$ & SSIM $\uparrow$ \\
\midrule

\multirow{7}{*}{100}
& 0\%  & Dense                & 5.53  & 261.87 & 18.93 & 0.81 & / \\
\cmidrule{2-8}

& \multirow{3}{*}{40\%}
& Wanda & 15.34 & 142.27 & 42.19 & 0.65 & 0.541 \\
& & + STW   &  11.79 &  161.01 &  42.53 &  0.74 &  0.561 \\
& & \textbf{+ STW + CAG}    & \textbf{7.19} & \textbf{213.22} & \textbf{23.58} & \textbf{0.77} & \textbf{0.627} \\
\cmidrule{2-8}

& \multirow{3}{*}{50\%}
& \cellcolor{LightBlueB}Wanda & \cellcolor{LightBlueB}50.96 & \cellcolor{LightBlueB}43.54  & \cellcolor{LightBlueB}97.97 & \cellcolor{LightBlueB}0.37 & \cellcolor{LightBlueB}0.463 \\
& & + \cellcolor{LightBlueA}STW   & \cellcolor{LightBlueA}43.47 & \cellcolor{LightBlueA}54.34  & \cellcolor{LightBlueA}89.22 & \cellcolor{LightBlueA}0.41 & \cellcolor{LightBlueA}0.477 \\
& & \textbf{\cellcolor{LightBlueC} + STW + CAG}    & \cellcolor{LightBlueC}\textbf{16.28} & \cellcolor{LightBlueC}\textbf{124.69} & \cellcolor{LightBlueC}\textbf{36.86} & \cellcolor{LightBlueC}\textbf{0.64} & \cellcolor{LightBlueC}\textbf{0.541} \\

\midrule

\multirow{7}{*}{50}
& 0\%  & Dense                & 6.68  & 236.27 & 21.62 & 0.78 & / \\
\cmidrule{2-8}

& \multirow{3}{*}{40\%}
& Wanda & 19.20 & 122.57 & 43.60 & 0.62 & 0.555 \\
& & + STW   & 12.86 &  143.31 &  43.61 &  0.73 &  0.576 \\
& & \textbf{+ STW + CAG}    & \textbf{8.78} & \textbf{194.21} & \textbf{24.50} & \textbf{0.75} & \textbf{0.646} \\
\cmidrule{2-8}
& \multirow{3}{*}{50\%}
& \cellcolor{LightBlueB} Wanda & \cellcolor{LightBlueB}57.32 & \cellcolor{LightBlueB}38.84  & \cellcolor{LightBlueB}101.30 & \cellcolor{LightBlueB}0.34 & \cellcolor{LightBlueB}0.473 \\
& & \cellcolor{LightBlueA} + STW   & \cellcolor{LightBlueA}47.11 & \cellcolor{LightBlueA}47.01  & \cellcolor{LightBlueA}92.57 & \cellcolor{LightBlueA}0.39 & \cellcolor{LightBlueA}0.488 \\
& & \textbf{\cellcolor{LightBlueC} + STW + CAG}     & \cellcolor{LightBlueC}\textbf{20.29} & \cellcolor{LightBlueC}\textbf{111.95} & \cellcolor{LightBlueC}\textbf{38.95} & \cellcolor{LightBlueC}\textbf{0.60} & \cellcolor{LightBlueC}\textbf{0.559} \\

\bottomrule
\end{tabular}
}
\caption{Ablation study of each component on DiT-XL/2.}
\label{tab:ablation1}
\medskip
\large
\setlength{\tabcolsep}{4pt}
\renewcommand{\arraystretch}{1.5}
\resizebox{\columnwidth}{!}{
\begin{tabular}{c c !{\vrule width 0.5pt} c c c c c c c c}
\toprule
Resolution & Sparsity & Ablation & FID $\downarrow$ & IS $\uparrow$ & CLIP $\uparrow$ & IR $\uparrow$ & SFID $\downarrow$ & PRE $\uparrow$ & SSIM $\uparrow$ \\
\midrule
\multirow{5}{*}{256$\times$256}
& 0\% & Dense & 54.94 & 26.11 & 0.260 & 0.686 & 259.18 & 0.75 & / \\
\cline{2-10}
& \multirow{2}{*}{4:8} & Wanda & 71.56 & 16.38 & 0.243 & -0.206 & 260.76 & 0.49 & 0.441 \\
& & \textbf{\cellcolor{LightBlueC}+ STW}
& \cellcolor{LightBlueC}\textbf{63.95}
& \cellcolor{LightBlueC}\textbf{20.51}
& \cellcolor{LightBlueC}\textbf{0.246}
& \cellcolor{LightBlueC}\textbf{0.189}
& \cellcolor{LightBlueC}\textbf{257.96}
& \cellcolor{LightBlueC}\textbf{0.61}
& \cellcolor{LightBlueC}\textbf{0.489} \\
\cline{2-10}
& \multirow{2}{*}{2:4} & Wanda & 79.35 & 14.63 & 0.235 & -0.613 & 263.35 & 0.42 & 0.412 \\
& & \textbf{\cellcolor{LightBlueC}+ STW}
& \cellcolor{LightBlueC}\textbf{69.21}
& \cellcolor{LightBlueC}\textbf{19.98}
& \cellcolor{LightBlueC}\textbf{0.245}
& \cellcolor{LightBlueC}\textbf{-0.078}
& \cellcolor{LightBlueC}\textbf{254.42}
& \cellcolor{LightBlueC}\textbf{0.55}
& \cellcolor{LightBlueC}\textbf{0.444} \\
\midrule
\multirow{5}{*}{512$\times$512}
& 0\% & Dense & 48.64 & 29.56 & 0.272 & 0.899 & 266.47 & 0.78 & / \\
\cline{2-10}
& \multirow{2}{*}{4:8} &Wanda & 61.31 & 20.09 & 0.259 & 0.414 & 256.52 & 0.64 & 0.522 \\
& & \textbf{\cellcolor{LightBlueC}+ STW}
& \cellcolor{LightBlueC}\textbf{57.58}
& \cellcolor{LightBlueC}\textbf{23.27}
& \cellcolor{LightBlueC}\textbf{0.261}
& \cellcolor{LightBlueC}\textbf{0.575}
& \cellcolor{LightBlueC}\textbf{254.47}
& \cellcolor{LightBlueC}\textbf{0.71}
& \cellcolor{LightBlueC}\textbf{0.548} \\
\cline{2-10}
& \multirow{2}{*}{2:4} & Wanda & 63.07 & 18.36 & 0.251 & 0.054 & 242.66 & 0.56 & 0.466 \\
& & \textbf{\cellcolor{LightBlueC}+ STW}
& \cellcolor{LightBlueC}\textbf{59.89}
& \cellcolor{LightBlueC}\textbf{23.32}
& \cellcolor{LightBlueC}\textbf{0.255}
& \cellcolor{LightBlueC}\textbf{0.340}
& \cellcolor{LightBlueC}\textbf{241.22}
& \cellcolor{LightBlueC}\textbf{0.65}
& \cellcolor{LightBlueC}\textbf{0.497} \\
\bottomrule
\end{tabular}
}
\caption{Ablation results of structured pruning on FLUX.1-dev at 256$\times$256 and 512$\times$512 resolutions on MJHQ.}
\label{tab:ablation2}
\end{table}

\section{Conclusion}
In this paper, we propose \textbf{DiT-Pruning}, a post-training pruning method for DiTs. Existing pruning methods suffer substantial performance degradation on DiTs due to their unique architecture, parameter distribution, and approximations that amplify weight contributions. To address this issue, we introduce a customized saliency criterion and pruning granularity to improve pruning performance. Specifically, we balance weight and activation contributions by squaring the weights, enabling more accurate identification of important elements. Furthermore, we observe that the proposed metric exhibits distinct clustering patterns in the two-dimensional weight space. Accordingly, we adopt a clustering-aware pruning granularity for effective sparse allocation. Extensive experiments show that our method significantly outperforms existing approaches, especially under high sparsity.


\appendix

\bibliography{aaai2027}

 \clearpage

 \appendix

\section{Weights and Activations}
\textbf{Weights.} Due to space limitations, we present the weight visualizations of only two representative layers in the main text. Similar distribution patterns are consistently observed across all layers of the DiTs, indicating that the observed phenomenon is a general characteristic of the model rather than an isolated case. For completeness, the weight distributions of all remaining model layers are presented below. Figure~\ref{weights1} shows detailed information, weights are predominantly distributed within the range of 0 to 1, with most values concentrated around the $10^{-1}$ scale , which is closely consistent with the distribution described before.
\begin{figure}[h] 
  \centering
  \includegraphics[width=0.5\columnwidth]{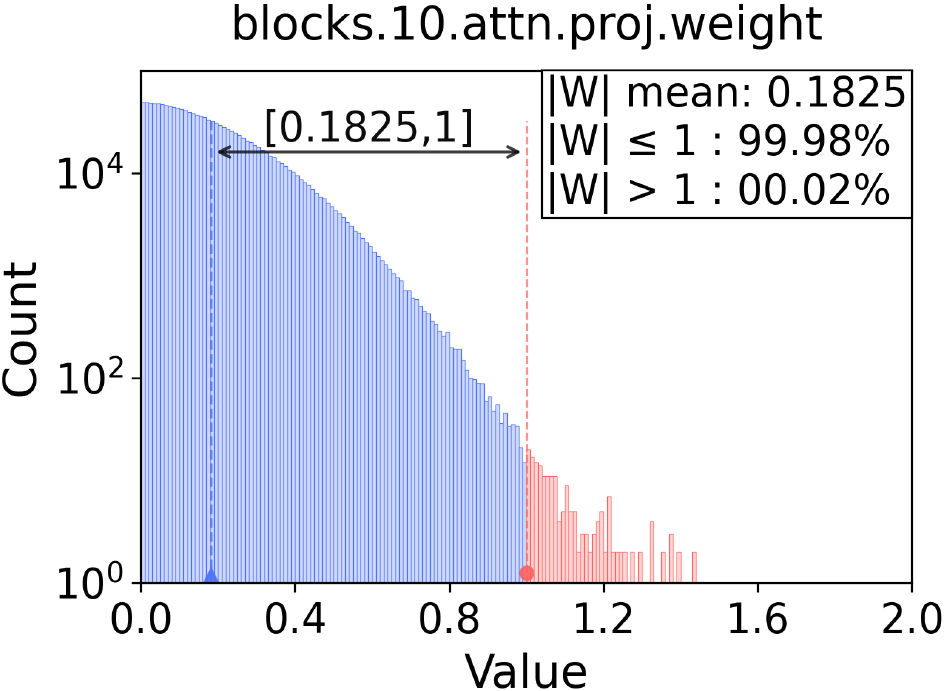}
  \vskip 0.5em 
  \includegraphics[width=\columnwidth]{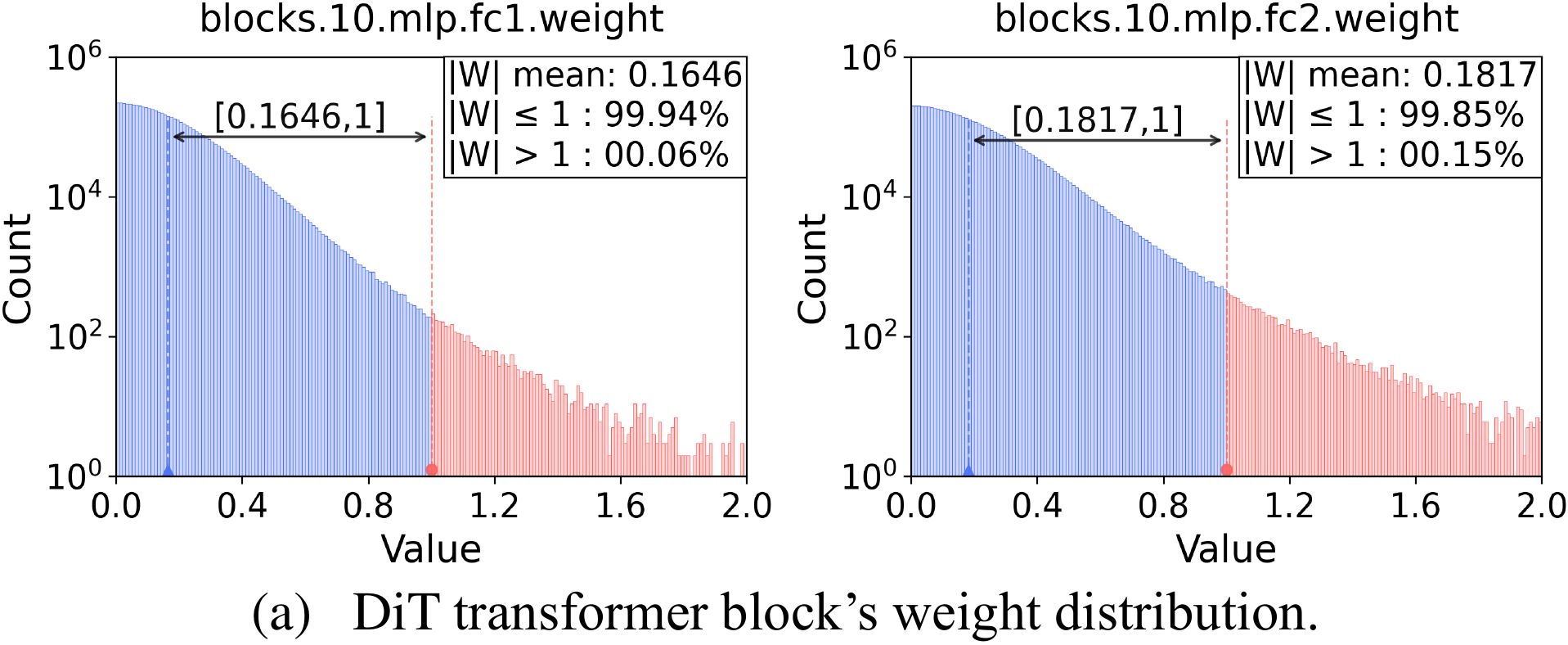}
  \caption{Remaining weight distribution of DiT transformer layers. All layers show similar characteristics.
  }
  \label{weights1}  
\end{figure}
\begin{table}[h]
\centering
\large
\setlength{\tabcolsep}{4pt}
\renewcommand{\arraystretch}{1.2}
\resizebox{\columnwidth}{!}{
\begin{tabular}{c c c c c c c}
\toprule
Layer & Metric & block5 & block10 & block15 & block20 & block25 \\
\midrule

\multirow{4}{*}{atten.qkv}
& $|X|$ mean & 1.1012 & 1.6378 & 1.7370 & 1.8147 & 1.9152 \\
& $|X|$ max & 22.3234 & 10.8295 & 9.2846 & 9.0306 & 9.4552 \\
& $|W|/|X|_{\rm mean}$ & 0.1052 & 0.0979 & 0.0972 & 0.0905 & 0.0832 \\
& $|W|^2/|X|_{\rm mean}$ & 0.0122 & 0.0157 & 0.0164 & 0.0149 & 0.0133 \\
\midrule

\multirow{4}{*}{atten.proj}
& $|X|$ mean & 4.9624 & 8.5670 & 11.7076 & 9.9487 & 13.3197 \\
& $|X|$ max & 16.7251 & 16.3386 & 36.9063 & 27.2713 & 36.1129 \\
& $|W|/|X|_{\rm mean}$ & 0.0283 & 0.0213 & 0.0159 & 0.0193 & 0.0149 \\
& $|W|^2/|X|_{\rm mean}$ & 0.0039 & 0.0039 & 0.0037 & 0.0037 & 0.0029 \\
\midrule

\multirow{4}{*}{mlp.fc1}
& $|X|$ mean & 1.5179 & 1.5725 & 1.1350 & 1.1803 & 1.4208 \\
& $|X|$ max & 17.1799 & 16.0704 & 8.7672 & 22.1027 & 41.0000 \\
& $|W|/|X|_{\rm mean}$ & 0.0936 & 0.1047 & 0.1453 & 0.1373 & 0.1098 \\
& $|W|^2/|X|_{\rm mean}$ & 0.0133 & 0.0172 & 0.0239 & 0.0223 & 0.0171 \\
\midrule

\multirow{4}{*}{mlp.fc2}
& $|X|$ mean & 2.7627 & 3.4125 & 2.9495 & 2.8912 & 3.0111 \\
& $|X|$ max & 16.0696 & 25.4478 & 22.8027 & 27.8377 & 36.2336 \\
& $|W|/|X|_{\rm mean}$ & 0.0589 & 0.0532 & 0.0647 & 0.0609 & 0.0646 \\
& $|W|^2/|X|_{\rm mean}$ & 0.0095 & 0.0097 & 0.0127 & 0.0129 & 0.0126 \\
\midrule

\multirow{4}{*}{adaLN}
& $|X|$ mean & 0.0084 & 0.0084 & 0.0084 & 0.0084 & 0.0084 \\
& $|X|$ max & 0.3409 & 0.3409 & 0.3409 & 0.3409 & 0.3409 \\
& $|W|/|X|_{\rm mean}$ & 11.45 & 16.42 & 16.00 & 13.76 & 13.07 \\
& $|W|^2/|X|_{\rm mean}$ & 1.1017 & 2.2638 & 2.1504 & 1.5908 & 1.4274 \\
\bottomrule
\end{tabular}
}
\caption{Statistics of activation norms of DiT layers.}
\label{tab:activation_stats1}
\end{table}

\textbf{Activations.} Table~\ref{tab:activation_stats1} and Table~\ref{tab:activation_stats2} report the activation statistics and weight-to-activation ratios of representative DiT and LLM layers, respectively. Owing to their much longer token sequences, LLMs exhibit significantly larger activation outliers. Despite this difference, the relative scales of weights and activations remain substantially more balanced in LLMs than in DiTs. \textbf{STW} reduces the dominant DiT weight scale from \(10^{-1}\) to \(10^{-2}\), yielding a quantitative relationship between weights and activations that is closer to the regime where Wanda has been shown to be effective.
\begin{table}[h]
\centering
\large
\setlength{\tabcolsep}{4pt}
\renewcommand{\arraystretch}{1.3}
\resizebox{\columnwidth}{!}{
\begin{tabular}{c c c c c c c}
\toprule
Layer & Metric & block5 & block10 & block15 & block20 & block25 \\
\midrule

\multirow{3}{*}{atten.q}
& $|X|$ mean & 11.4072 & 14.1838 & 15.8599 & 17.7827 & 19.9309 \\
& $|X|$ max & 214.7673 & 229.1075 & 210.9702 & 185.4212 & 189.7775 \\
& $|W|/|X|_{\rm mean}$ & 0.0019 & 0.0015 & 0.0013 & 0.0011 & 0.0009 \\
\midrule

\multirow{3}{*}{atten.k}
& $|X|$ mean & 11.4072 & 14.1838 & 15.8599 & 17.7827 & 19.9309 \\
& $|X|$ max & 214.7673 & 229.1075 & 210.9702 & 185.4212 & 189.7775 \\
& $|W|/|X|_{\rm mean}$ & 0.0019 & 0.0009 & 0.0013 & 0.0011 & 0.0009 \\
\midrule

\multirow{3}{*}{atten.v}
& $|X|$ mean & 11.4072 & 14.1838 & 15.8599 & 17.7827 & 19.9309 \\
& $|X|$ max & 214.7673 & 229.1075 & 210.9702 & 185.4212 & 189.7775 \\
& $|W|/|X|_{\rm mean}$ & 0.0012 & 0.0009 & 0.0013 & 0.0009 & 0.0009 \\
\midrule

\multirow{3}{*}{atten.o}
& $|X|$ mean & 1.7841 & 5.1910 & 6.5438 & 6.4441 & 6.3077 \\
& $|X|$ max & 14.9358 & 37.6035 & 26.8657 & 43.6976 & 36.2537 \\
& $|W|/|X|_{\rm mean}$ & 0.0073 & 0.0027 & 0.0023 & 0.0027 & 0.0029 \\
\midrule

\multirow{3}{*}{mlp.up}
& $|X|$ mean & 6.9415 & 8.6943 & 10.9269 & 14.2518 & 16.3169 \\
& $|X|$ max & 95.0688 & 83.8602 & 53.8245 & 43.4382 & 58.0632 \\
& $|W|/|X|_{\rm mean}$ & 0.0024 & 0.0020 & 0.0016 & 0.0012 & 0.0011 \\
\midrule

\multirow{3}{*}{mlp.gate}
& $|X|$ mean & 6.9415 & 8.6943 & 10.9269 & 14.2518 & 16.3169 \\
& $|X|$ max & 95.0688 & 83.8602 & 53.8245 & 43.4382 & 58.0632 \\
& $|W|/|X|_{\rm mean}$ & 0.0027 & 0.0021 & 0.0016 & 0.0013 & 0.0011 \\
\midrule

\multirow{3}{*}{mlp.down}
& $|X|$ mean & 1.7290 & 2.9034 & 4.5845 & 7.4560 & 8.5851 \\
& $|X|$ max & 16.3258 & 18.8423 & 31.4849 & 65.0971 & 55.4523 \\
& $|W|/|X|_{\rm mean}$ & 0.0096 & 0.0058 & 0.0038 & 0.0024 & 0.0021 \\
\bottomrule
\end{tabular}
}
\caption{Statistics of activation norms of LLM layers.}
\label{tab:activation_stats2}
\end{table}

\section{Compression and Speedup Effects}

\textbf{Compression and Acceleration.} DiT-Pruning introduces no additional parameters and therefore preserves the inherent benefits of sparsification. As shown in Table~\ref{tab:efficiency}, both parameter count and computational cost decrease consistently with increasing sparsity. Specifically, TFLOPs is used to measure the computational cost and reflect the acceleration effect, while Params is used to measure the reduction in model storage. The reports demonstrate that increasing compression sparsity results in significant reduction in both memory footprint and inference cost, highlighting its effectiveness.
\begin{table}[h]
\centering
\Large
\setlength{\tabcolsep}{4pt}
\renewcommand{\arraystretch}{1.3}
\resizebox{\columnwidth}{!}{
\begin{tabular}{c c c c c c c}
\toprule
Model & Method & Sparsity & Params$\downarrow$ & TFLOPs$\downarrow$ & TF/50steps$\downarrow$ & Speedup$\uparrow$ \\
\midrule
\multirow{5}{*}{DiT-XL/2}
& \multirow{5}{*}{DiT-Pruning}
& 0\%  & 675.13M & 0.114 & 5.723 &  / \\
& & 20\% & 541.43M & 0.092 & 4.589 & $\sim$1.2$\times$ \\
& & 40\% & 407.73M & 0.069 & 3.456 & $\sim$1.4$\times$ \\
& & 50\% & 340.69M & 0.058 & 2.888 & $\sim$1.6$\times$ \\
& & 2:4  & 340.68M & 0.058 & 2.888 & $\sim$1.6$\times$ \\
\bottomrule
\end{tabular}
}
\caption{Model compression and inference efficiency of DiT-Pruning on DiT-XL/2 under different sparsity levels.}
\label{tab:efficiency}
\end{table}

\textbf{Acceleration Analysis.} Existing sparse inference frameworks~\citep{lu2023dasp, macko2025macko} have demonstrated substantial acceleration for unstructured sparse computation, while prior work reports approximately (1.6$\times$) runtime speedup for Transformer layers under 2:4 structured sparsity~\citep{sun2023simple}. Since DiT-Pruning follows the same unstructured and 2:4 sparsity patterns without introducing additional modules, the achieved reductions in model size and FLOPs can be directly translated into practical efficiency gains under these sparse execution frameworks.

\section{Experiment Settings}
\textbf{Parameter Settings.} For DiT-XL/2, we follow the standard ImageNet generation setup with a classifier-free guidance (CFG) scale of 1.5, a fixed random seed of 42, and the MSE VAE for latent coding. Evaluations are conducted under different timesteps. For PixArt-$\Sigma$, we use DPM-Solver with 20 denoising steps and a CFG scale of 4.5. For FLUX.1-dev, we adopt 50 inference steps with a guidance scale of 3.5. For fair comparison across resolutions, all reference images are resized to the target resolution when necessary.

\textbf{Evaluations.} We evaluate model performance using generation quality, text-image alignment, and structural consistency metrics. Generation quality is assessed by FID, sFID, Precision, and Inception Score (IS), following the official evaluation pipeline of OpenAI's guided-diffusion repository with the corresponding reference batches. Text-image alignment is measured using CLIP with the official pretrained weights. For FLUX.1-dev, we additionally report Image Reward (IR), following the official evaluation protocol. Structural consistency is evaluated using SSIM. All evaluation settings follow the corresponding official implementations.

\section{Calibration Samples}
\textbf{Number of Calibration Samples.} To evaluate the sensitivity to calibration set size, we progressively increase the number of calibration samples while keeping all other settings fixed. Experiments are conducted on DiT-XL/2 at 256$\times$256 resolution with 50 denoising steps and 5,000 generated samples. As shown in Table~\ref{tab:calibration_size}, the performance remains stable across a wide range of calibration sizes, demonstrating strong robustness to calibration data selection. Moreover, increasing the calibration set beyond 128 samples yields only marginal improvements at all sparsity levels, indicating that 128 samples are sufficient for reliable pruning calibration.

\begin{figure}[h]
  \begin{center}
    \centerline{\includegraphics[width=0.95\columnwidth]{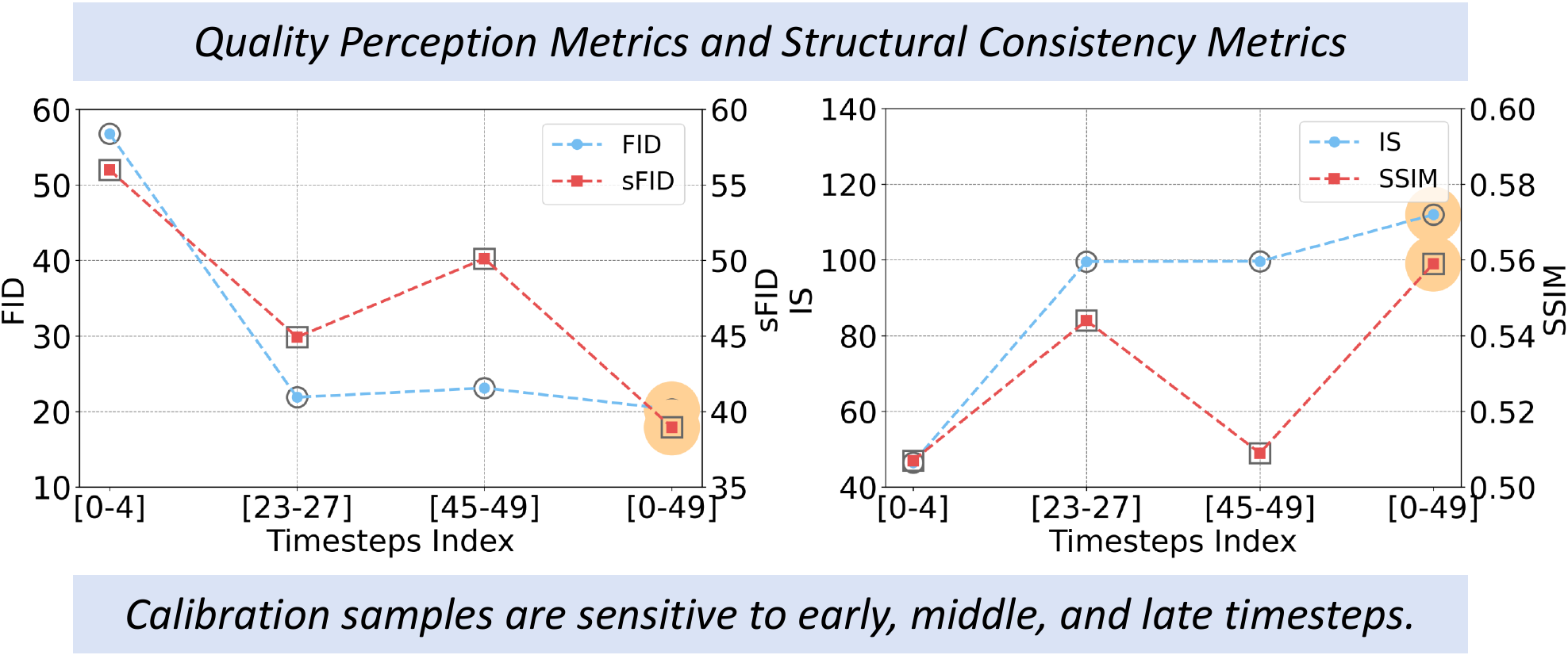}}
    \caption{
       Calibration timestep evaluation.
    }
    \label{timesteps}
  \end{center}
\end{figure}

\textbf{Selection of Calibration Samples.} We further investigate the impact of calibration data selection. Specifically, we fix the calibration set size to 128 and sample data from different diffusion timesteps. As shown in Figure~\ref{timesteps}, pruning performance is highly sensitive to timestep coverage. Calibration data drawn exclusively from early or late stages consistently lead to inferior results, whereas broad timestep coverage yields better performance. This observation suggests that different timesteps exhibit distinct activation distributions and parameter sensitivity patterns in DiTs. Therefore, effective calibration data should provide diversity across diffusion timesteps to ensure representative activation statistics and more reliable pruning decisions.
\begin{table}[t]
\centering
\scriptsize
\setlength{\tabcolsep}{4pt}
\renewcommand{\arraystretch}{1}
\resizebox{\columnwidth}{!}{
\begin{tabular}{c c c !{\vrule width 0.5pt} c c c c c}
\toprule
Model & Num. Calib. & Sparsity & FID $\downarrow$ & sFID $\downarrow$ & IS $\uparrow$ & PRE $\uparrow$ & SSIM $\uparrow$ \\
\midrule

\multirow{12}{*}{DiT-XL/2}
& \multirow{3}{*}{128}
& 20\% & 10.32 & 250.38 & 37.52 & 0.79 & 0.853 \\
& & 40\% & 12.75 & 194.92 & 40.46 & 0.74 & 0.649 \\
& & 50\% & \cellcolor{purple}24.14 & \cellcolor{purple}111.01 & \cellcolor{purple}53.50 & \cellcolor{purple}0.59 & \cellcolor{purple}0.560 \\
\cmidrule(lr){2-8}

& \multirow{3}{*}{256}
& 20\% & 10.31 & 251.53 & 37.59 & 0.78 & 0.853 \\
& & 40\% & 12.76 & 198.04 & 40.36 & 0.74 & 0.649 \\
& & 50\% & \cellcolor{purple}23.99 & \cellcolor{purple}112.32 & \cellcolor{purple}53.86 & \cellcolor{purple}0.59 & \cellcolor{purple}0.562 \\
\cmidrule(lr){2-8}

& \multirow{3}{*}{512}
& 20\% & 10.47 & 250.96 & 37.45 & 0.77 & 0.854 \\
& & 40\% & 12.76 & 197.51 & 40.33 & 0.73 & 0.651 \\
& & 50\% & \cellcolor{purple}23.98 & \cellcolor{purple}111.94 & \cellcolor{purple}53.82 & \cellcolor{purple}0.59 & \cellcolor{purple}0.561 \\
\cmidrule(lr){2-8}

& \multirow{3}{*}{1024}
& 20\% & 10.45 & 251.55 & 37.32 & 0.78 & 0.853 \\
& & 40\% & 12.75 & 197.33 & 40.32 & 0.75 & 0.649 \\
& & 50\% & \cellcolor{purple}24.01 & \cellcolor{purple}112.03 & \cellcolor{purple}53.01 & \cellcolor{purple}0.58 & \cellcolor{purple}0.560 \\

\bottomrule
\end{tabular}
}
\caption{Effect of calibration set size under different sparsity.}
\label{tab:calibration_size}
\end{table}

\section{More Ablation Materials}
Due to space limitations, the main text reports ablation results only on DiT-XL/2 at 40\% and 50\% sparsity, as well as N:M structured pruning results on FLUX. Table~\ref{tab:resolution_sparsity} provides additional ablation results on PixArt-$\Sigma$ under the same settings, offering a comprehensive evaluation of the proposed method. Similar to the observations on DiT-XL/2, \textbf{STW} consistently improves the baseline, while \textbf{CAG} yields further gains when combined with \textbf{STW}. The results indicate that the two components are complementary and generalize well across different DiT-based architectures.

\begin{table}[ht]
\centering
\normalsize
\setlength{\tabcolsep}{4pt}
\renewcommand{\arraystretch}{1.4}
\resizebox{\columnwidth}{!}{
\begin{tabular}{c c !{\vrule width 0.5pt} c c c c c c c c}
\toprule
Resolution & Sparsity & Method 
& FID $\downarrow$ & IS $\uparrow$ & CLIP $\uparrow$ 
& IR $\uparrow$ & sFID $\downarrow$ & PRE $\uparrow$ & SSIM $\uparrow$ \\
\midrule

\multirow{7}{*}{256$\times$256}
& 0\%  & Dense & 70.31 & 33.33 & 0.265 & 0.807 & 249.59 & 0.49 & / \\
\cmidrule(lr){2-10}
& \multirow{3}{*}{40\%}
& Wanda & 69.32 & 28.86 & 0.256 & 0.244 & \textbf{235.79} & 0.47 & 0.456 \\
& & + STW   & 67.88 & 30.12 & 0.260 & 0.593 & 239.47 & 0.48 & 0.468 \\
& & \textbf{+ STW + CAG} & \textbf{67.5} & \textbf{31.06} & \textbf{0.260} & \textbf{0.635} & 244.21 & \textbf{0.51} & \textbf{0.484} \\
\cmidrule(lr){2-10}
& \multirow{3}{*}{50\%}
& \cellcolor{LightBlueB}Wanda & \cellcolor{LightBlueB}89.66 & \cellcolor{LightBlueB}20.47 & \cellcolor{LightBlueB}0.229 & \cellcolor{LightBlueB}-0.923 & \cellcolor{LightBlueB}234.46 & \cellcolor{LightBlueB}0.27 & \cellcolor{LightBlueB}0.381 \\
& & \cellcolor{LightBlueA}+ STW   & \cellcolor{LightBlueA}75.7  & \cellcolor{LightBlueA}26.52 & \cellcolor{LightBlueA}0.247 & \cellcolor{LightBlueA}-0.122 & \cellcolor{LightBlueA}231.97 & \cellcolor{LightBlueA}0.37 & \cellcolor{LightBlueA}0.382 \\
& & \textbf{\cellcolor{LightBlueC}+ STW + CAG}  & \cellcolor{LightBlueC}\textbf{69.79} & \cellcolor{LightBlueC}\textbf{29.76} & \cellcolor{LightBlueC}\textbf{0.251} & \cellcolor{LightBlueC}\textbf{0.183} & \cellcolor{LightBlueC}\textbf{237.6} & \cellcolor{LightBlueC}\textbf{0.42} & \cellcolor{LightBlueC}\textbf{0.385} \\

\midrule

\multirow{7}{*}{512$\times$512}
& 0\%  & Dense & 62.36 & 36.12 & 0.256 & 0.929 & 274.47 & 0.59 & / \\
\cmidrule(lr){2-10}
& \multirow{3}{*}{40\%}
& Wanda & 61.26 & 33.33 & 0.247 & 0.360 & \textbf{256.87} & 0.57 & 0.475 \\
& & + STW   & 59.63 & 35.29 & 0.252 & 0.754 & 259.69 & 0.61 & 0.483 \\
& & \textbf{+ STW + CAG}  & \textbf{56.74} & \textbf{35.98} & \textbf{0.253} & \textbf{0.803} & 264.47 & \textbf{0.62} & \textbf{0.514} \\
\cmidrule(lr){2-10}
& \multirow{3}{*}{50\%}
& \cellcolor{LightBlueB}Wanda & \cellcolor{LightBlueB}93.97 & \cellcolor{LightBlueB}19.81 & \cellcolor{LightBlueB}0.219 & \cellcolor{LightBlueB}-1.143 & \cellcolor{LightBlueB}263.23 & \cellcolor{LightBlueB}0.29 & \cellcolor{LightBlueB}0.373 \\
& & \cellcolor{LightBlueA}+ STW  & \cellcolor{LightBlueA}73.27 & \cellcolor{LightBlueA}28.47 & \cellcolor{LightBlueA}0.242 & \cellcolor{LightBlueA}-0.154 & \cellcolor{LightBlueA}253.74 & \cellcolor{LightBlueA}0.41 & \cellcolor{LightBlueA}0.428 \\
& & \textbf{\cellcolor{LightBlueC}+ STW + CAG}  & \cellcolor{LightBlueC}\textbf{61.78} & \cellcolor{LightBlueC}\textbf{33.27} & \cellcolor{LightBlueC}\textbf{0.244} & \cellcolor{LightBlueC}\textbf{0.285} & \cellcolor{LightBlueC}\textbf{254.82} & \cellcolor{LightBlueC}\textbf{0.51} & \cellcolor{LightBlueC}\textbf{0.439} \\

\bottomrule
\end{tabular}
}
\caption{Ablation study of each component on PixArt-$\Sigma$.}
\label{tab:resolution_sparsity}
\end{table}

\textbf{Effect of CFG Scale.} We evaluate pruning performance under different classifier-free guidance (CFG) scales while keeping the sparsity ratio and denoising steps fixed. Following the official DiT setup, CFG=1.5 is used as the default, while CFG=1 and CFG=2 represent weaker and stronger guidance strengths, respectively. As shown in Table~\ref{tab:cfg_ablation}, varying CFG changes the trade-off between conditional guidance and generation diversity. Nevertheless, our method consistently achieves the best performance across all CFG settings. These results demonstrate the robustness of our method to variations in guidance strength and sampling configuration.
\begin{table}[ht]
\centering
\normalsize
\setlength{\tabcolsep}{4pt}
\renewcommand{\arraystretch}{1.3}
\resizebox{\columnwidth}{!}{
\begin{tabular}{c c !{\vrule width 0.5pt} c c c c c c c}
\toprule
Timesteps & Sparsity & CFG Scale & Method & FID $\downarrow$ & IS $\uparrow$ & sFID $\downarrow$ & PRE $\uparrow$ & SSIM $\uparrow$ \\
\midrule

\multirow{9}{*}{50}
& \multirow{9}{*}{50\%}
& \multirow{3}{*}{2.0}
& Magnitude & 38.14 & 72.04 & 40.08 & 0.44 & 0.516 \\
&&& Wanda & 33.39 & 81.99 & 77.78 & 0.49 & 0.491 \\
&&& \cellcolor{LightBlueC}DiT-Pruning
& \cellcolor{LightBlueC}\textbf{10.82}
& \cellcolor{LightBlueC}\textbf{199.70}
& \cellcolor{LightBlueC}\textbf{34.02}
& \cellcolor{LightBlueC}\textbf{0.73}
& \cellcolor{LightBlueC}\textbf{0.565} \\
\cmidrule(lr){3-9}

&& \multirow{3}{*}{1.5}
& Magnitude & 62.40 & 37.98 & 50.50 & 0.32 & 0.508 \\
&&& Wanda & 57.32 & 38.84 & 101.30 & 0.34 & 0.473 \\
&&& \cellcolor{LightBlueC}DiT-Pruning
& \cellcolor{LightBlueC}\textbf{20.29}
& \cellcolor{LightBlueC}\textbf{111.95}
& \cellcolor{LightBlueC}\textbf{38.95}
& \cellcolor{LightBlueC}\textbf{0.60}
& \cellcolor{LightBlueC}\textbf{0.559} \\
\cmidrule(lr){3-9}

&& \multirow{3}{*}{1.0}
& Magnitude & 97.76 & 18.62 & 67.32 & 0.22 & 0.220 \\
&&& Wanda & 90.50 & 17.23 & 135.69 & 0.21 & 0.316 \\
&&& \cellcolor{LightBlueC}DiT-Pruning
& \cellcolor{LightBlueC}\textbf{46.54}
& \cellcolor{LightBlueC}\textbf{43.67}
& \cellcolor{LightBlueC}\textbf{48.98}
& \cellcolor{LightBlueC}\textbf{0.42}
& \cellcolor{LightBlueC}\textbf{0.519} \\
\bottomrule
\end{tabular}
}
\caption{Effect of CFG scale on DiT-XL/2 pruning.}
\label{tab:cfg_ablation}
\end{table}

\textbf{Effect of Optional $(m,n)$ Settings.} To validate the choice of the proposed squared transformation, we perform an ablation study by varying the exponents $(m,n)$ in Eq.~\eqref{eq:parametric_form_1}, while keeping the \textbf{CAG} strategy fixed. As shown in Table~\ref{tab:mn_settings}, the proposed setting $(m=2,n=1)$ consistently achieves the best overall performance, yielding the lowest FID and highest IS, sFID, PRE, and SSIM. In contrast, both smaller and larger weight exponents lead to inferior results. These observations suggest that $(m=2,n=1)$ provides a more suitable balance between weight and activation contributions, supporting the effectiveness of the proposed \textbf{STW}.
\begin{table}[ht]
\centering
\large
\resizebox{\columnwidth}{!}{
\setlength{\tabcolsep}{4pt}
\renewcommand{\arraystretch}{1.4}
\begin{tabular}{c c !{\vrule width 0.5pt} c c c c c c}
\toprule
Timesteps & Sparsity & Settings 
& FID $\downarrow$ 
& IS $\uparrow$ 
& sFID $\downarrow$ 
& PRE $\uparrow$ 
& SSIM $\uparrow$ \\
\midrule

\multirow{4}{*}{50}
& \multirow{4}{*}{50\%}
& $m=1,\ n=1$     & 22.23 & 100.91 & 41.71 & 0.57 & 0.557 \\
\cmidrule{3-8}
& 
& $m=0.5,\ n=1$   & 41.92 &  53.64 & 57.71 & 0.42 & 0.521 \\
& 
& $m=3,\ n=1$     & 22.76 & 104.36 & 42.27 & 0.57 & 0.547 \\
& 
& \cellcolor{purple}$m=2,\ n=1$     & \cellcolor{purple}\textbf{20.29} & \cellcolor{purple}\textbf{111.95} & \cellcolor{purple}\textbf{38.95} & \cellcolor{purple}\textbf{0.60} & \cellcolor{purple}\textbf{0.559} \\
\bottomrule
\end{tabular}
}
\caption{Effect of $(m,n)$ settings under CAG on  DiT-XL/2.}
\label{tab:mn_settings}
\end{table}

\section{Additional Visualization Results}
\begin{figure}[t] 
  \centering
  \includegraphics[width=\columnwidth]{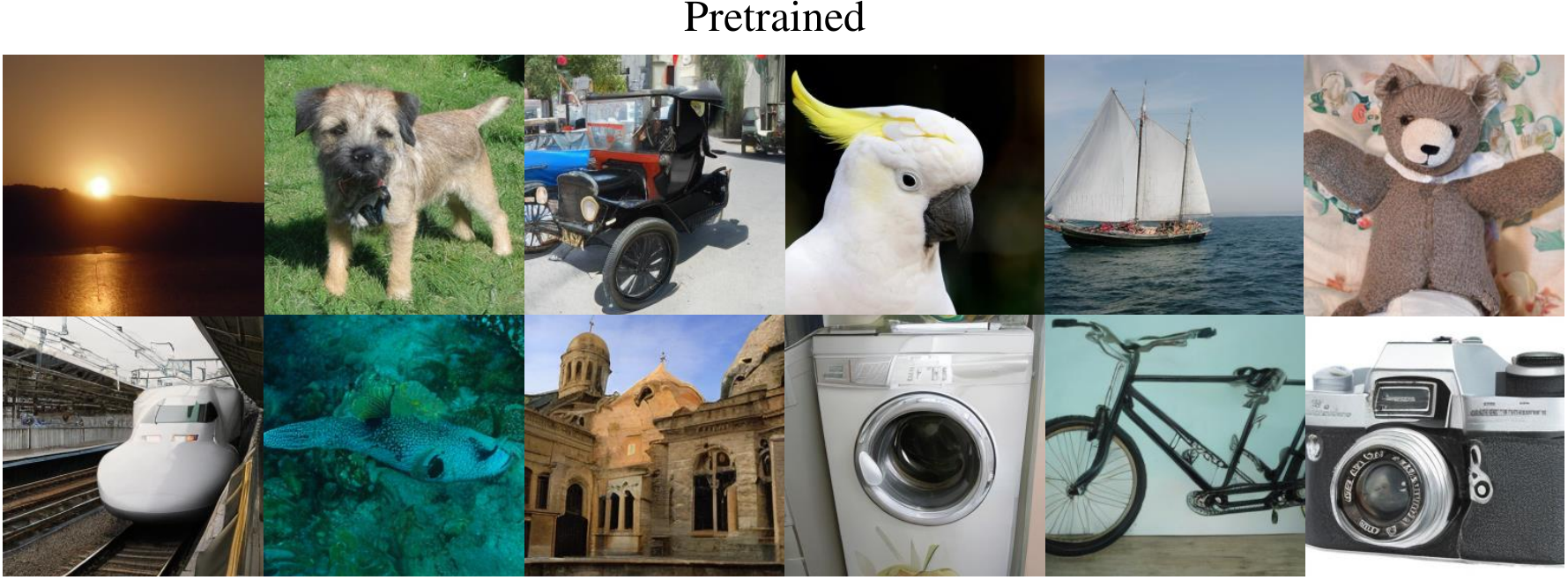}
  \vskip 0.5em 
  \includegraphics[width=\columnwidth]{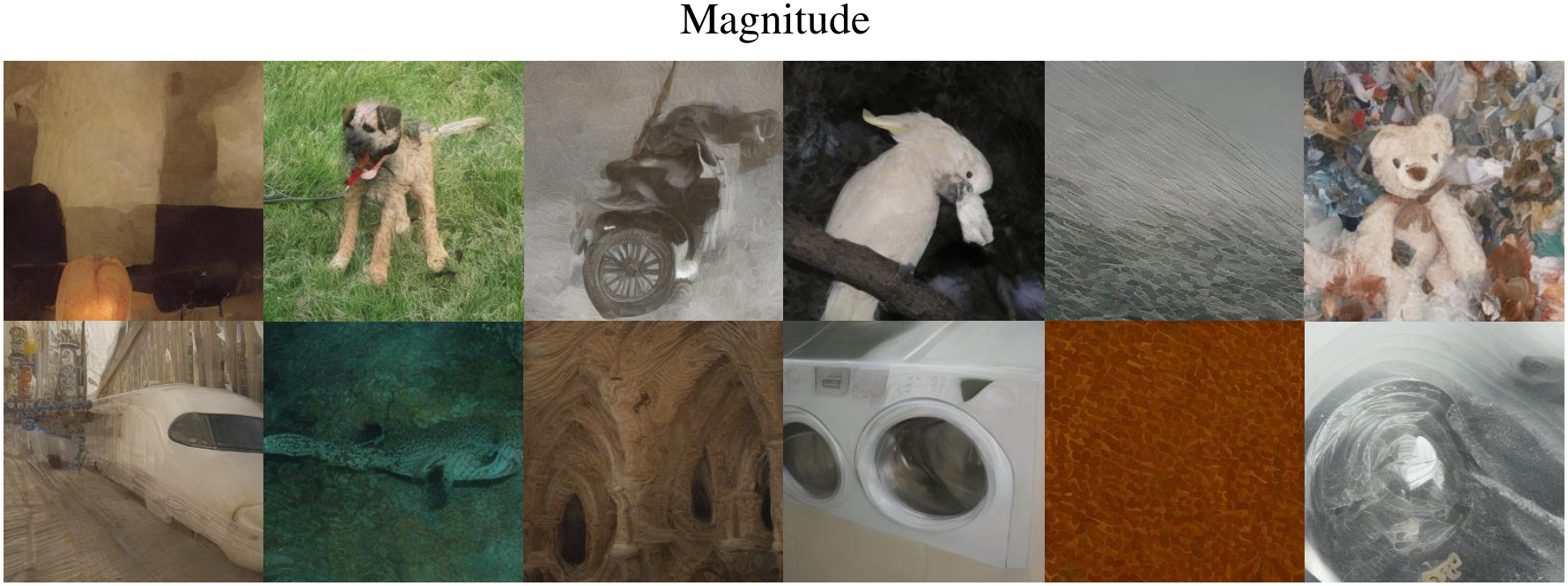}
    \vskip 0.5em 
  \includegraphics[width=\columnwidth]{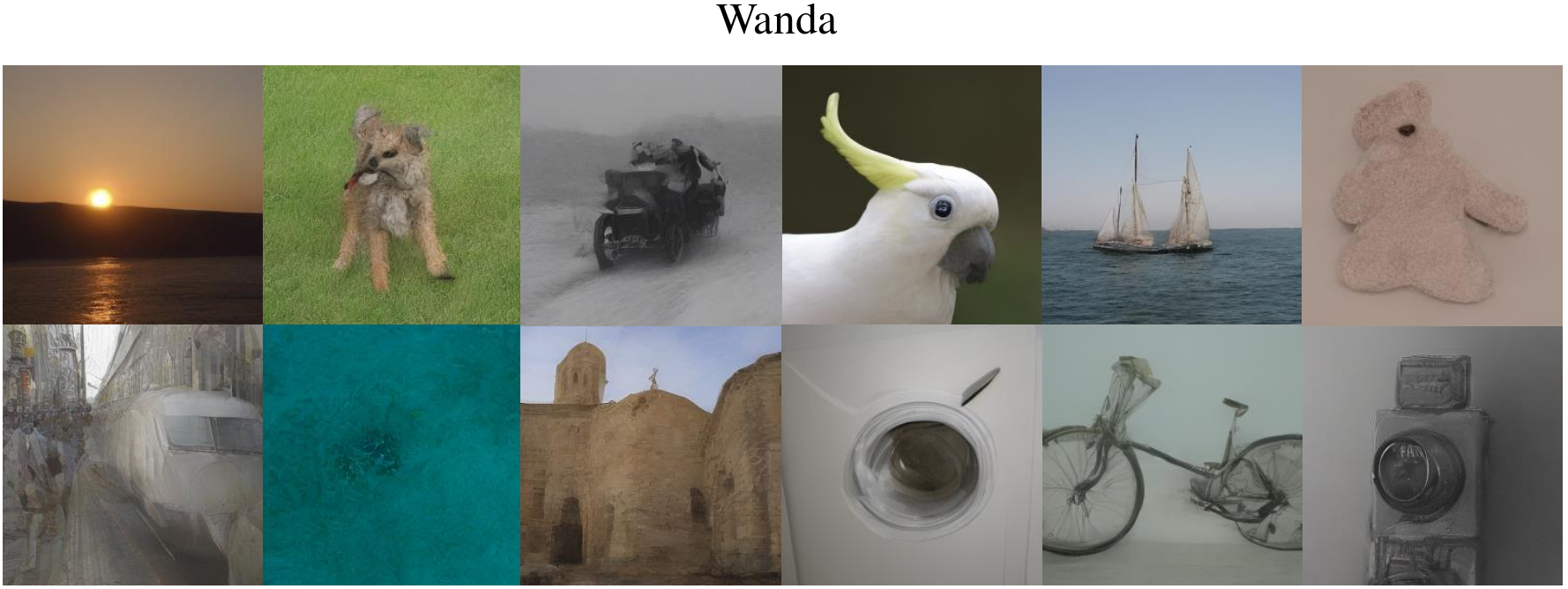}
    \vskip 0.5em 
  \includegraphics[width=\columnwidth]{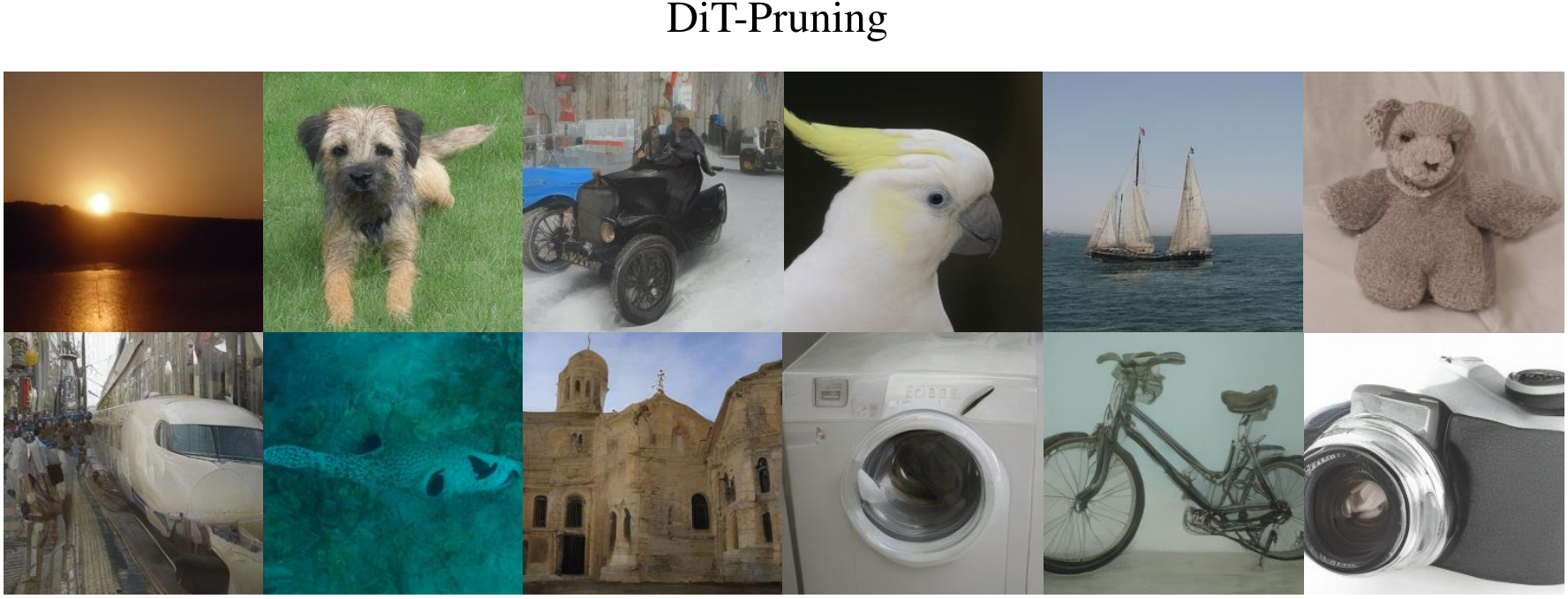}
  \caption{Random samples generated by the pruned DiT-XL/2 model at 50\% sparsity and a resolution of 256$\times$256. Our method preserves image fidelity, fine-grained details and structural integrity, while baseline methods exhibit substantial degradation in visual quality.
  }
  \label{samplesss}
\end{figure}

\begin{figure}[t]
  \centering
  \includegraphics[width=\columnwidth,height=0.125\textheight]{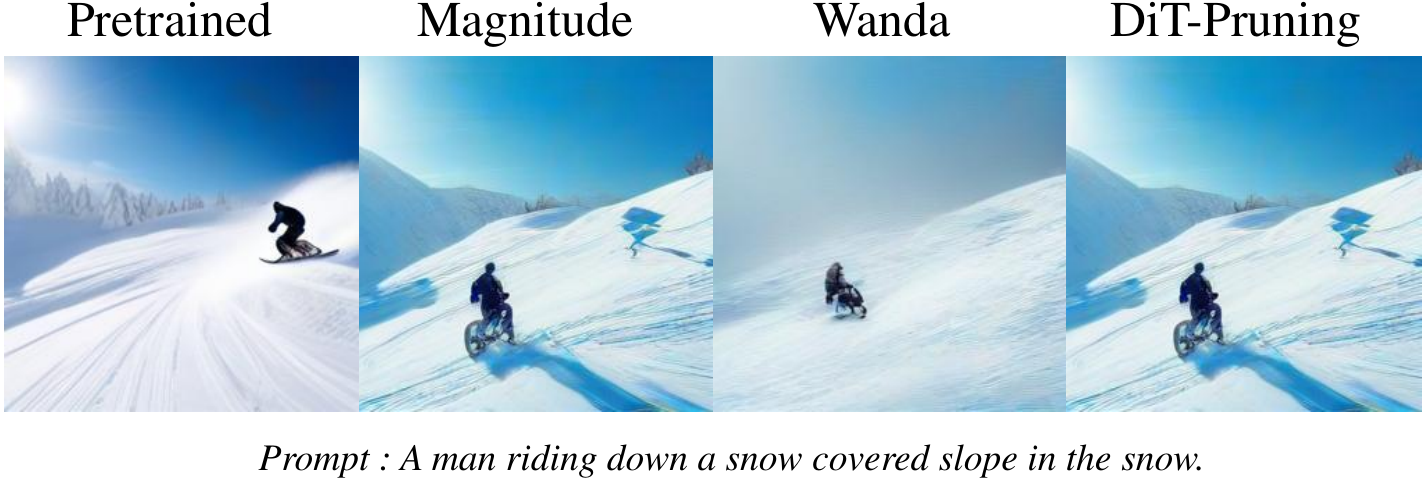}
  \vskip 0.5em 
  \includegraphics[width=\columnwidth,height=0.11\textheight]{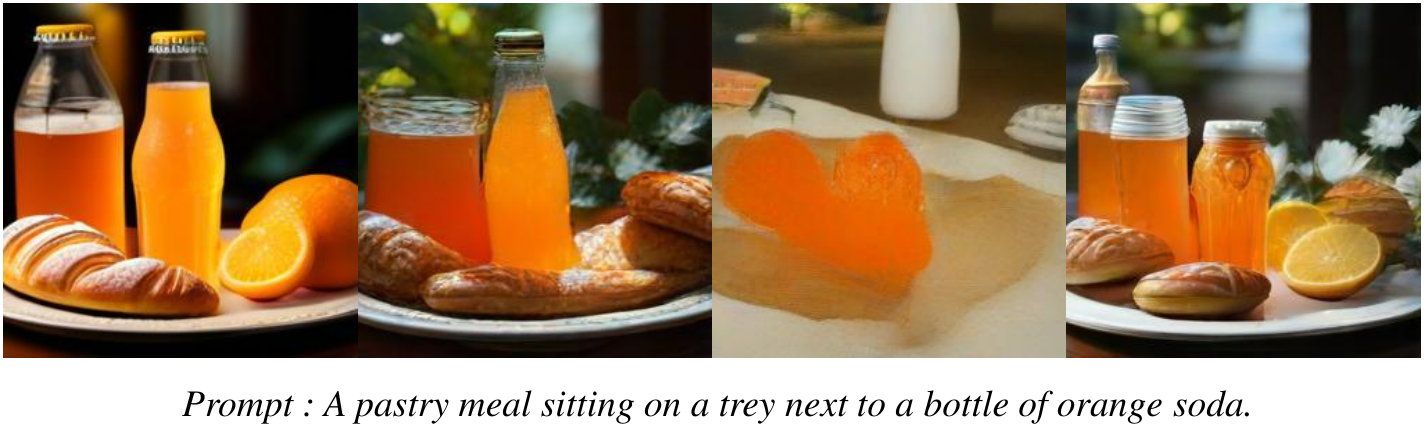}
    \vskip 0.5em 
  \includegraphics[width=\columnwidth,height=0.11\textheight]{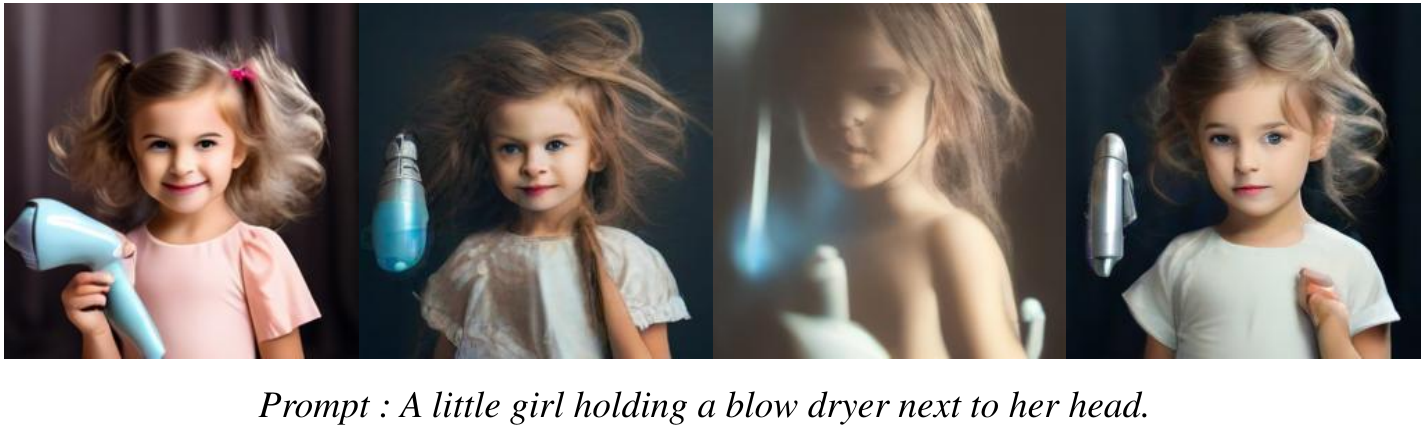}
    \vskip 0.5em 
  \includegraphics[width=\columnwidth,height=0.11\textheight]{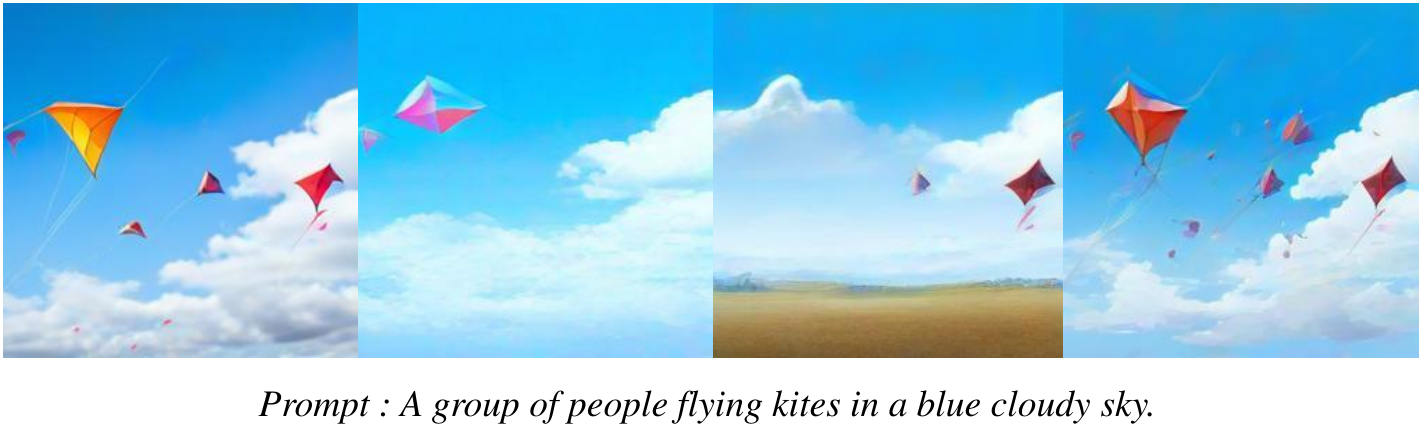}
      \vskip 0.5em 
  \includegraphics[width=\columnwidth,height=0.11\textheight]{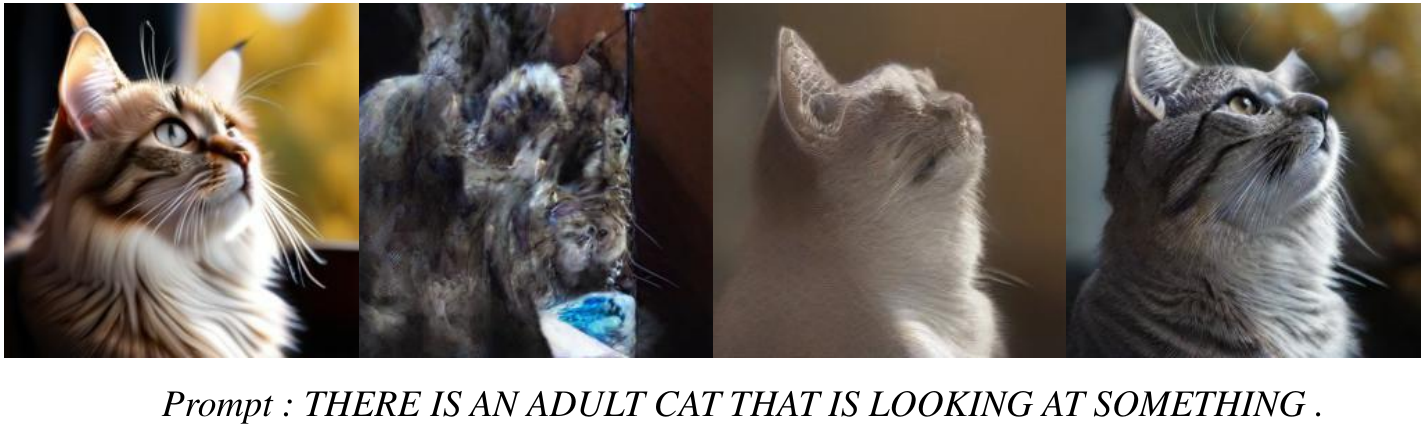}
    \vskip 0.5em
  \includegraphics[width=\columnwidth,height=0.11\textheight]{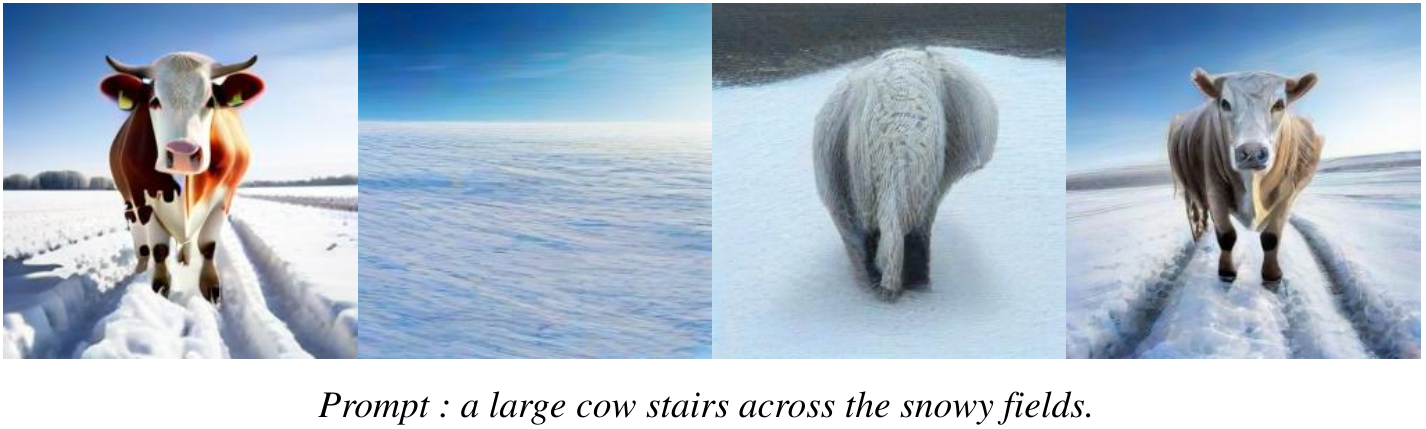}
    \vskip 0.5em
  \includegraphics[width=\columnwidth,height=0.11\textheight]{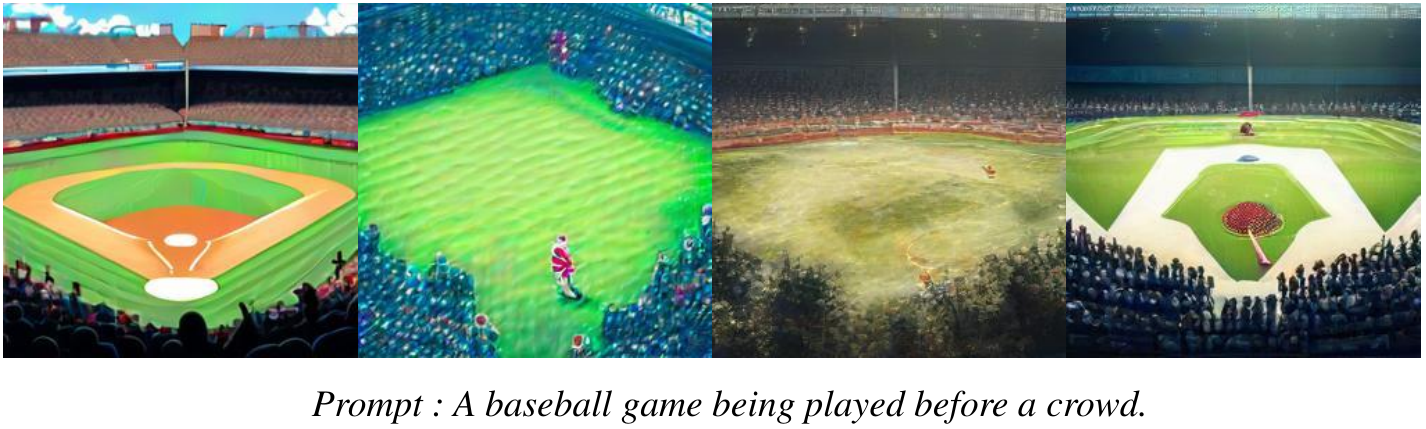}
    \vskip 0.5em
  \includegraphics[width=\columnwidth,height=0.11\textheight]{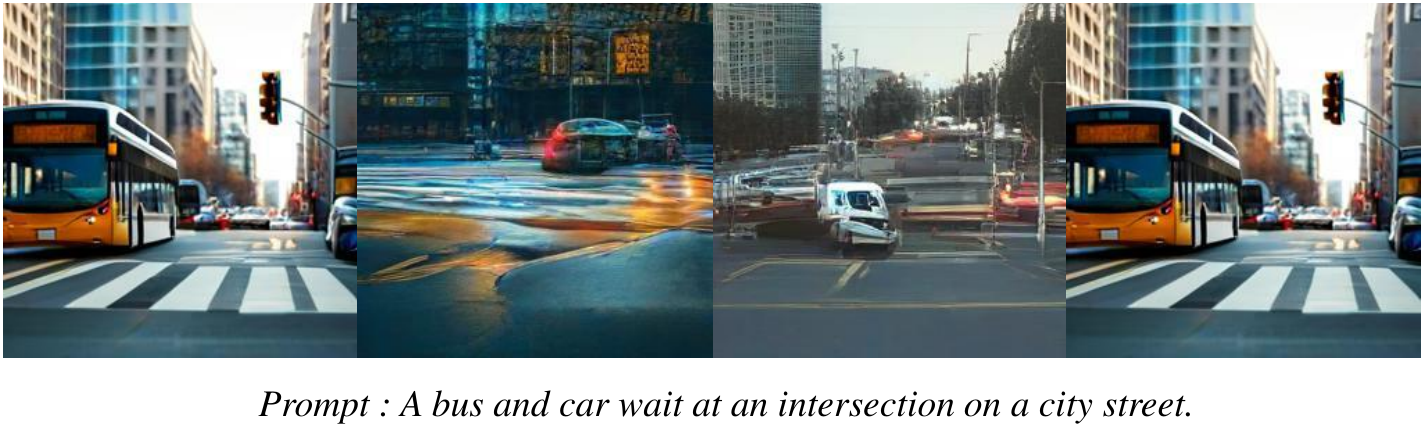}
  \caption{Samples generated by pruned PixArt model at 50\% sparsity and 256x256 resolution on Coco datasets.
  }
  \label{samplespixart1}
\end{figure}

\begin{figure}[t] 
  \centering
  \includegraphics[width=\columnwidth,height=0.125\textheight]{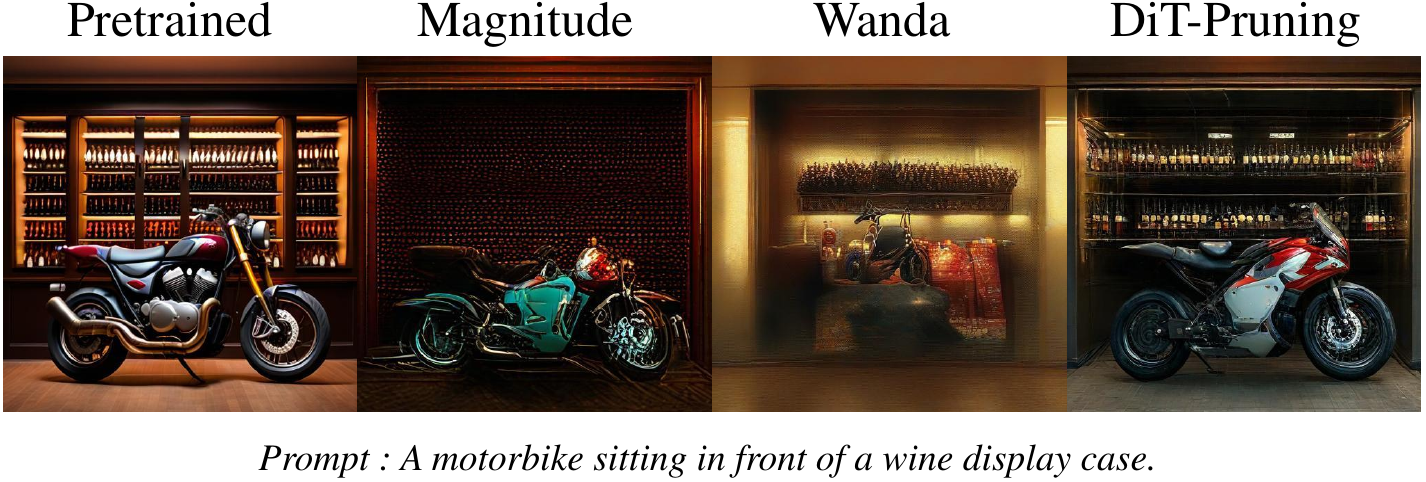}
  \vskip 0.5em 
  \includegraphics[width=\columnwidth,height=0.11\textheight]{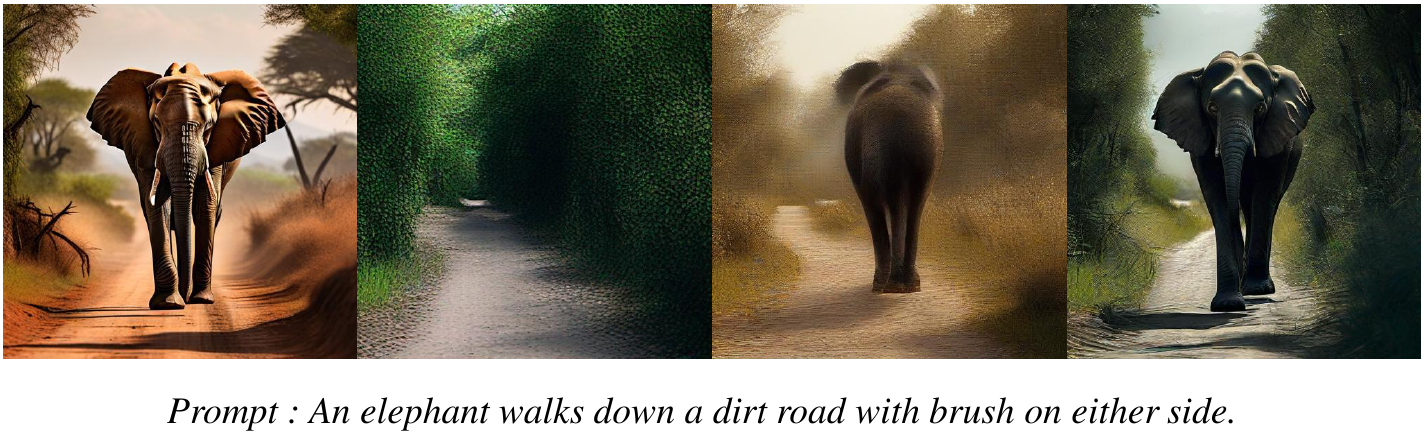}
    \vskip 0.5em 
  \includegraphics[width=\columnwidth,height=0.11\textheight]{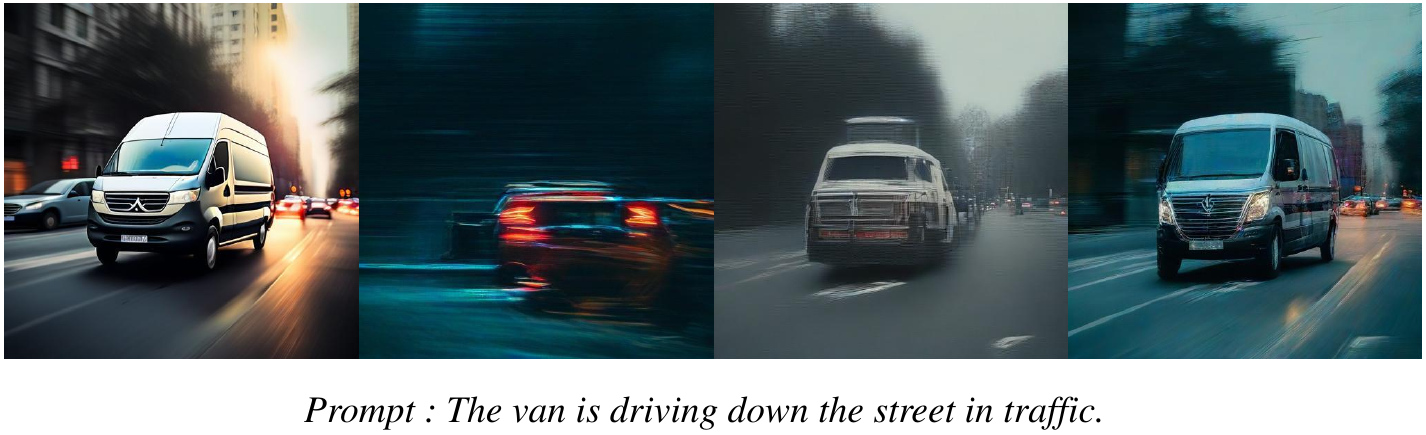}
    \vskip 0.5em 
  \includegraphics[width=\columnwidth,height=0.11\textheight]{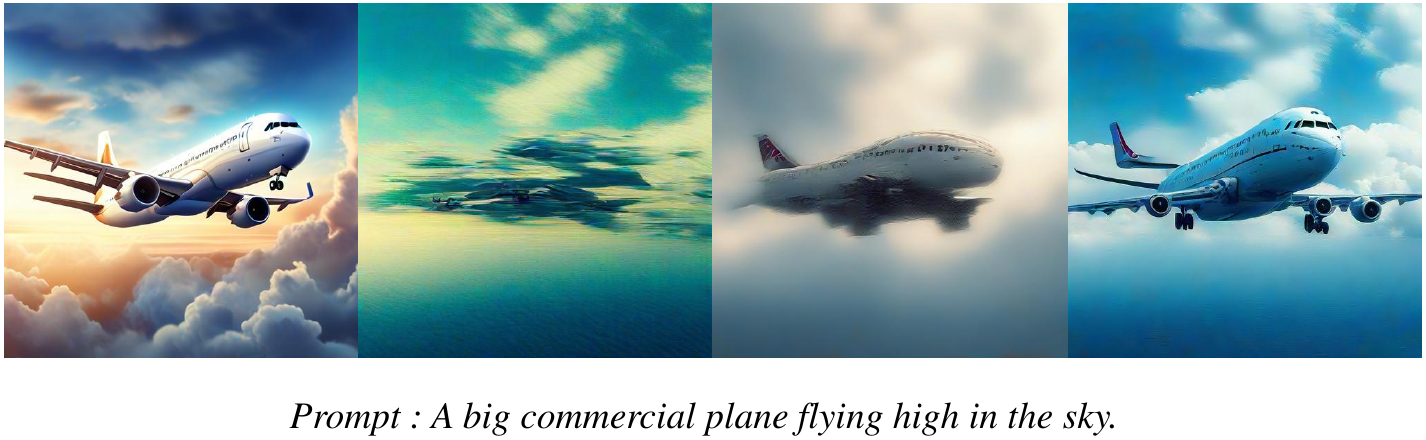}
      \vskip 0.5em 
  \includegraphics[width=\columnwidth,height=0.11\textheight]{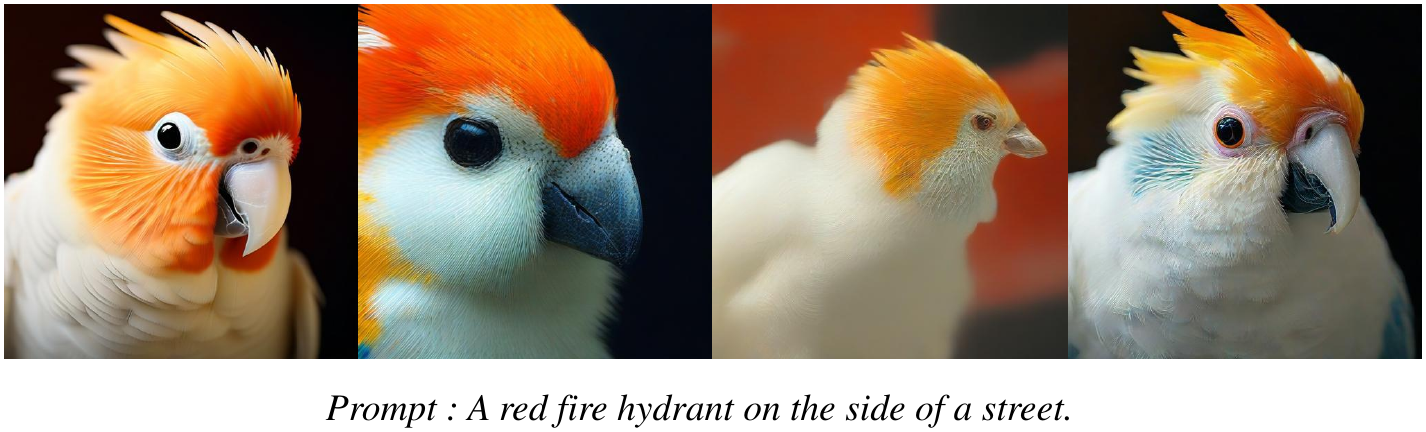}
    \vskip 0.5em
  \includegraphics[width=\columnwidth,height=0.11\textheight]{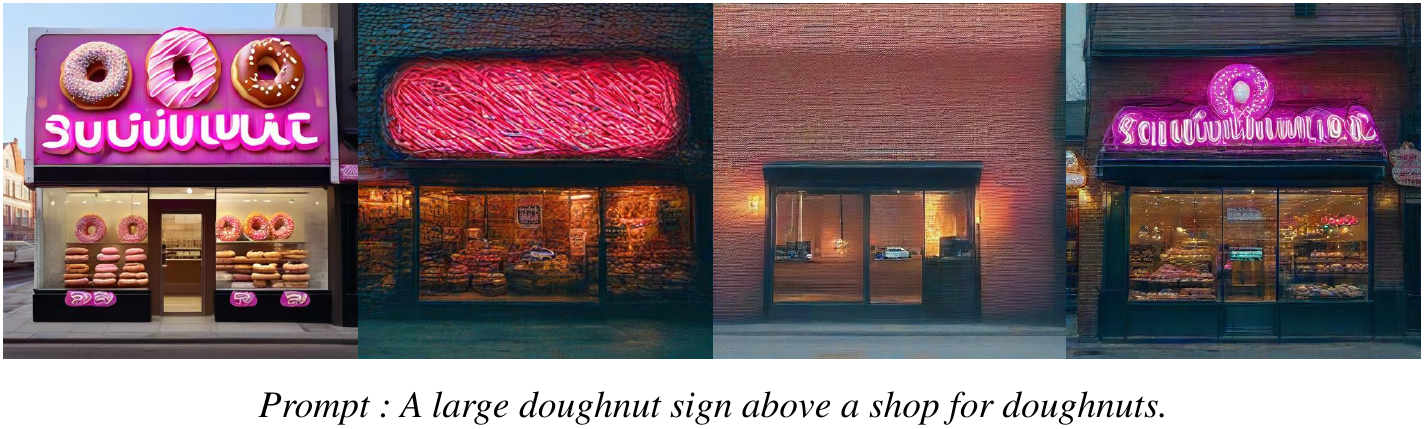}
    \vskip 0.5em
  \includegraphics[width=\columnwidth,height=0.11\textheight]{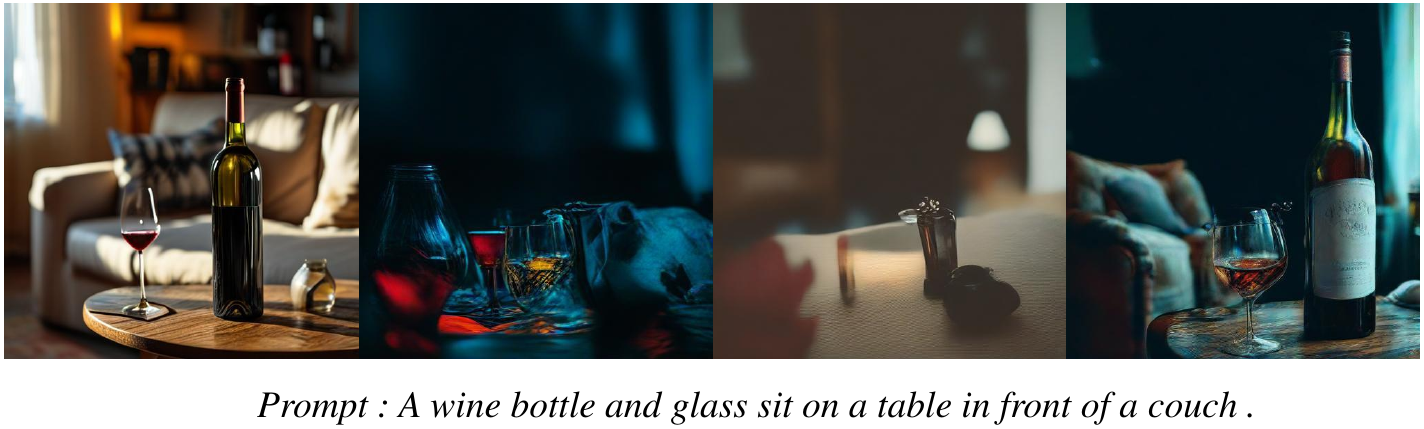}
    \vskip 0.5em
  \includegraphics[width=\columnwidth,height=0.11\textheight]{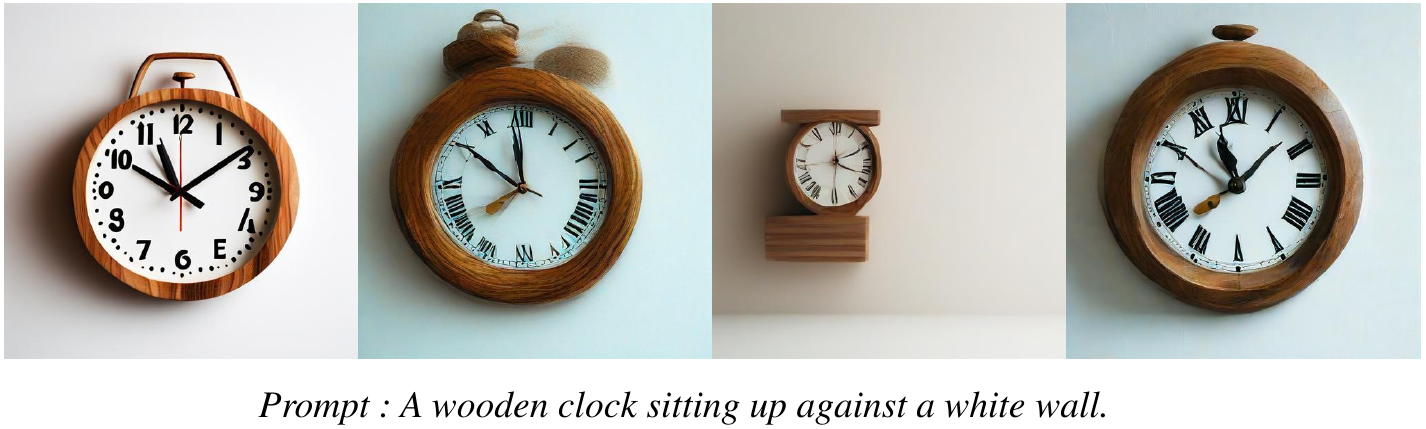}
  \caption{Samples generated by pruned PixArt model at 50\% sparsity and 512x512 resolution on Coco datasets.
  }
  \label{samplespixart2}
\end{figure}

\begin{figure}[t] 
  \centering
  \includegraphics[width=\columnwidth,height=0.125\textheight]{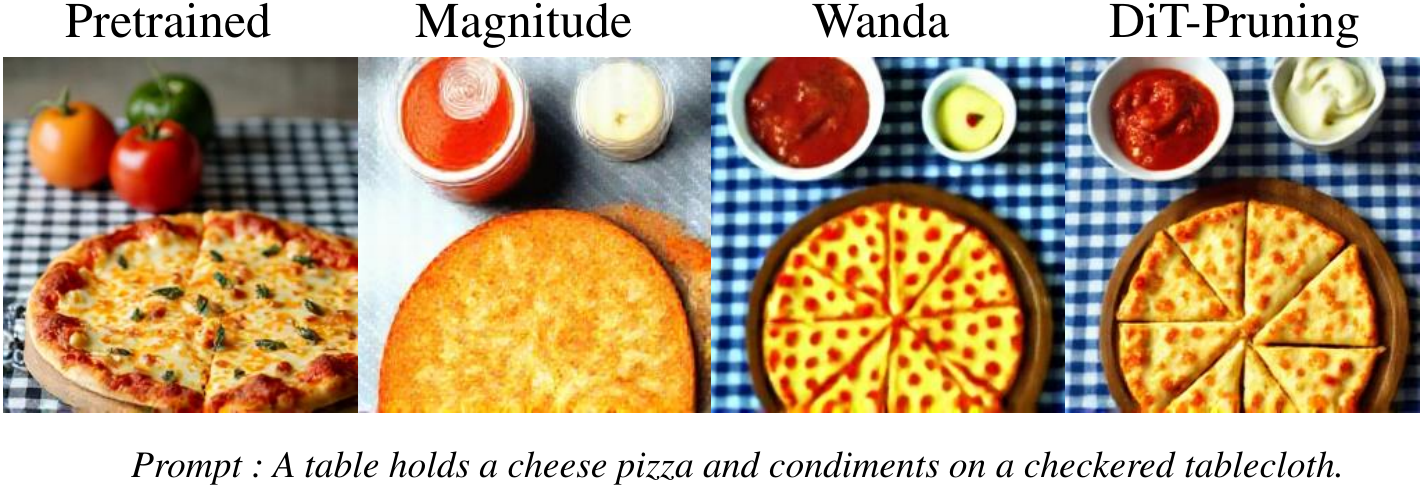}
  \vskip 0.5em 
  \includegraphics[width=\columnwidth,height=0.11\textheight]{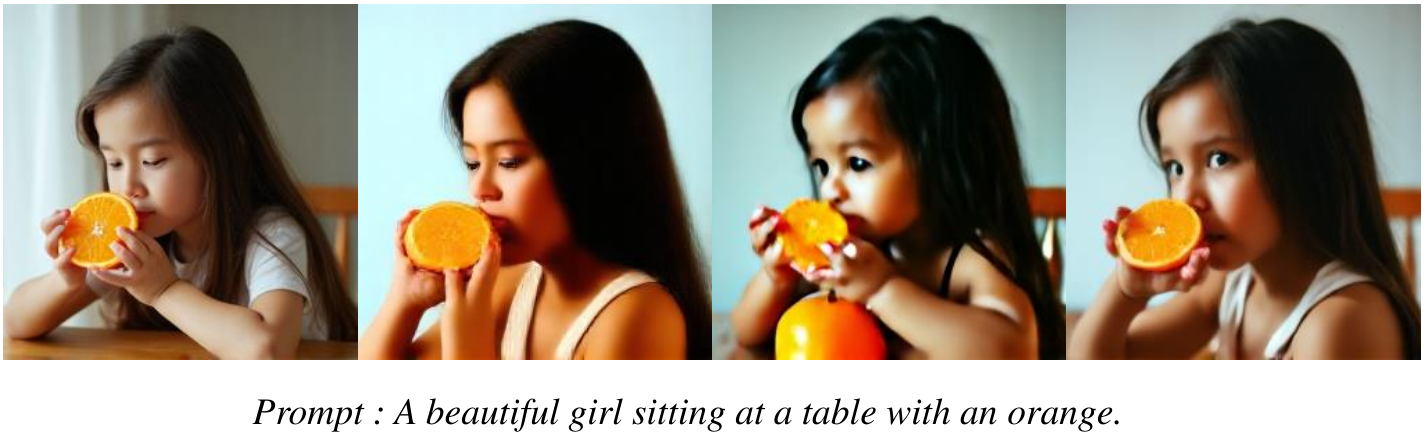}
    \vskip 0.5em 
  \includegraphics[width=\columnwidth,height=0.11\textheight]{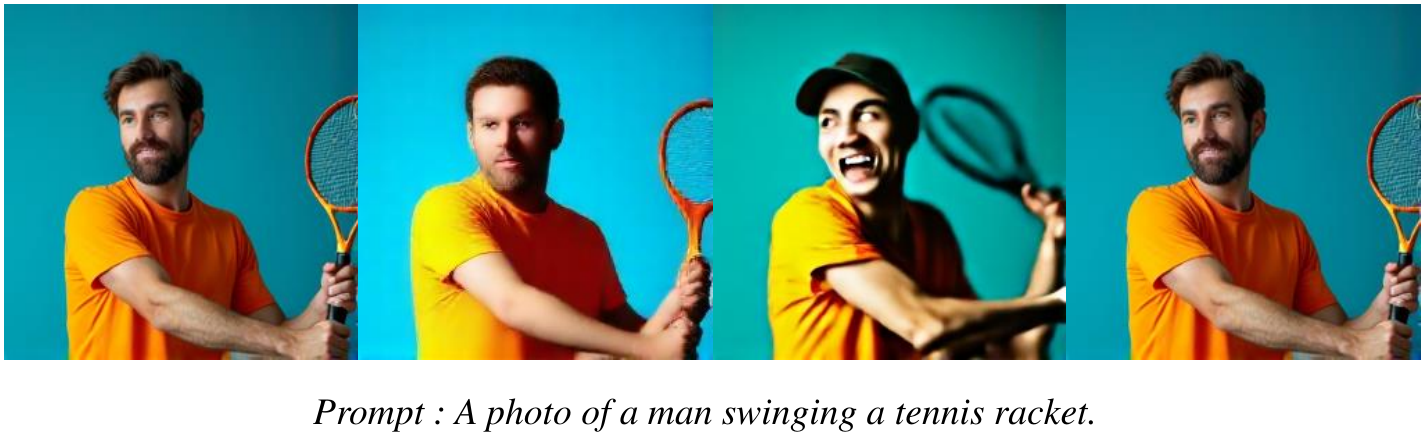}
    \vskip 0.5em 
  \includegraphics[width=\columnwidth,height=0.11\textheight]{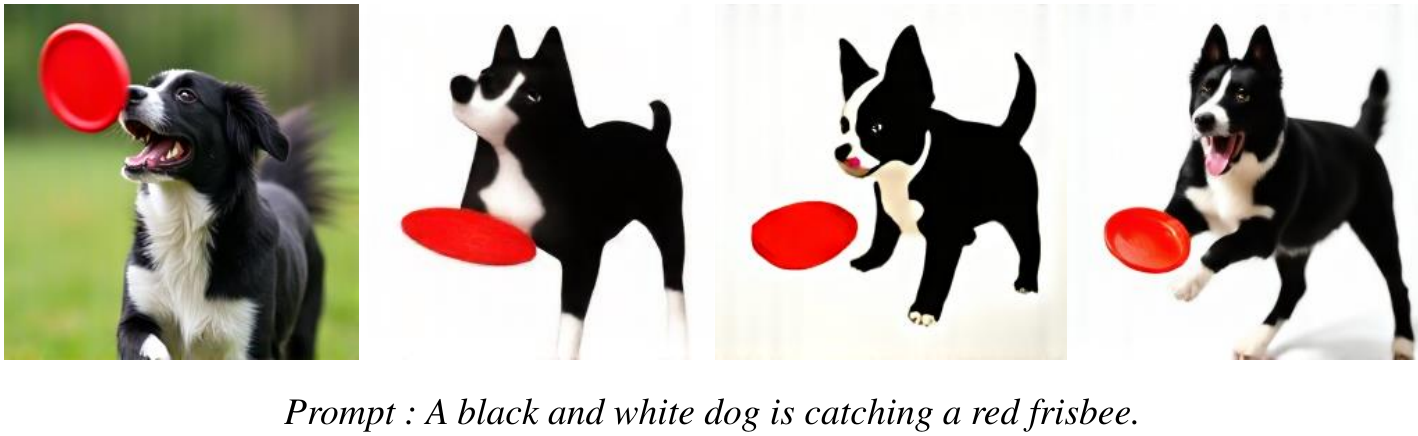}
      \vskip 0.5em 
  \includegraphics[width=\columnwidth,height=0.11\textheight]{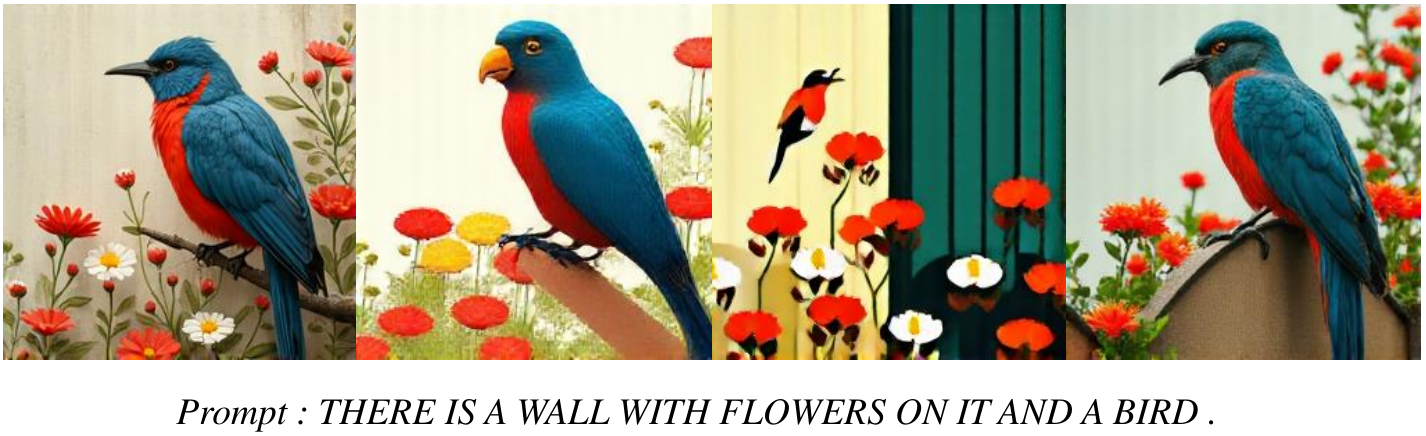}
    \vskip 0.5em
  \includegraphics[width=\columnwidth,height=0.11\textheight]{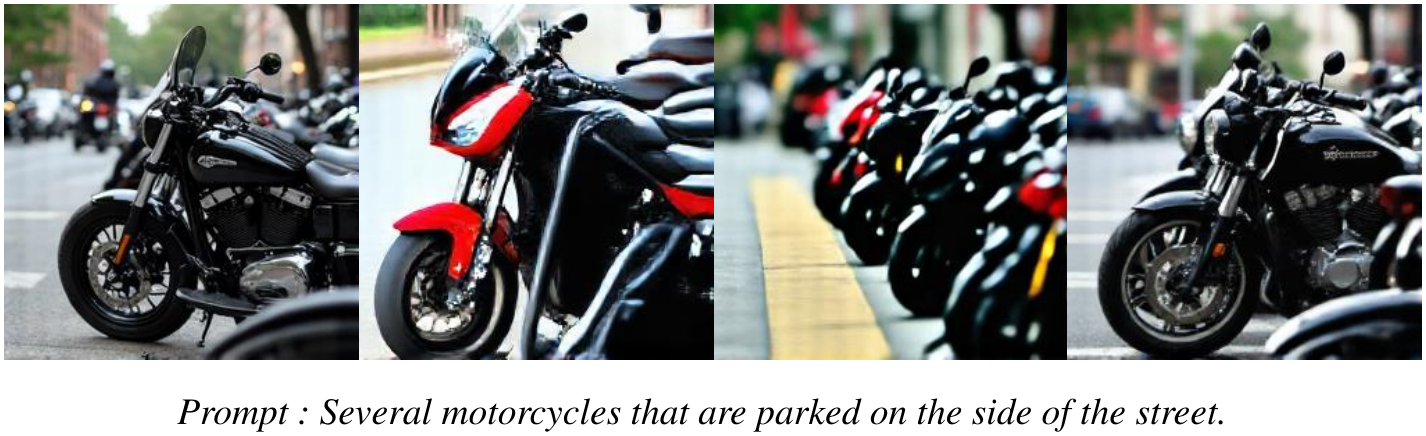}
    \vskip 0.5em
  \includegraphics[width=\columnwidth,height=0.11\textheight]{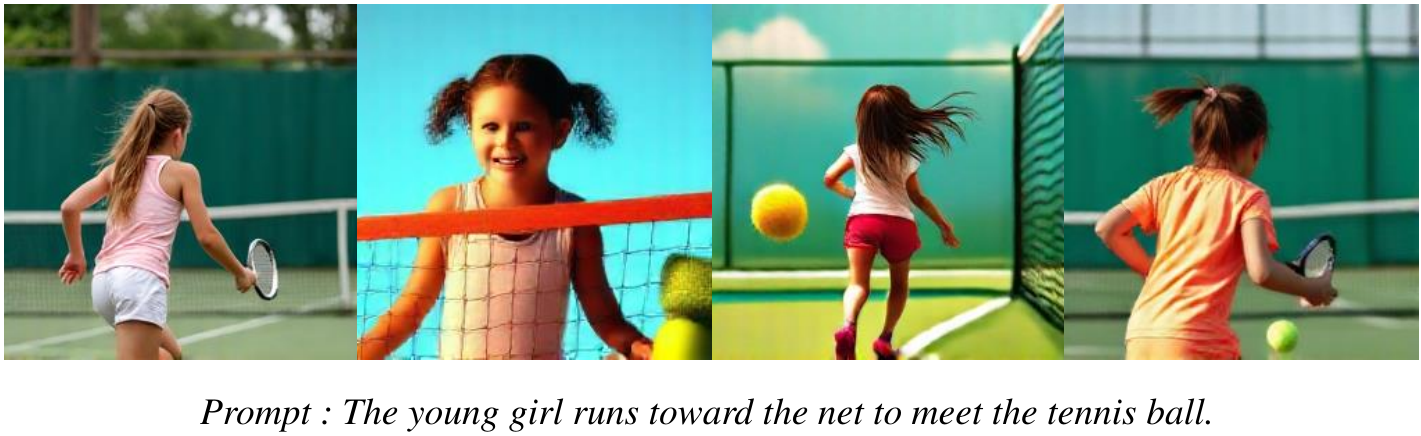}
    \vskip 0.5em
  \includegraphics[width=\columnwidth,height=0.11\textheight]{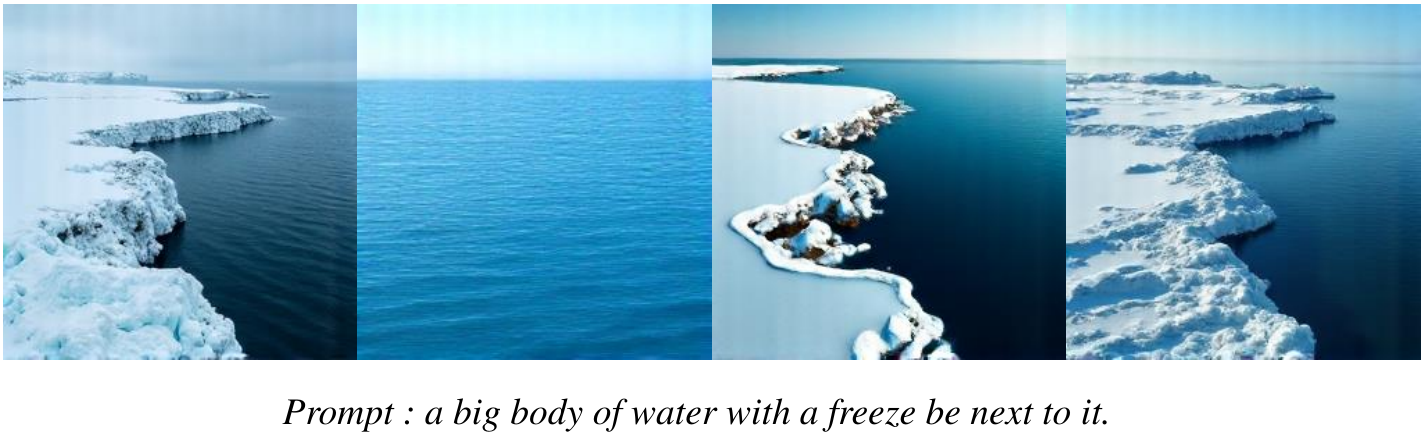}
  \caption{Samples generated by pruned Flux model at 50\% sparsity and 256x256 resolution on Coco datasets.
  }
  \label{samplesflux1}
\end{figure}

\begin{figure}[t] 
  \centering
  \includegraphics[width=\columnwidth,height=0.125\textheight]{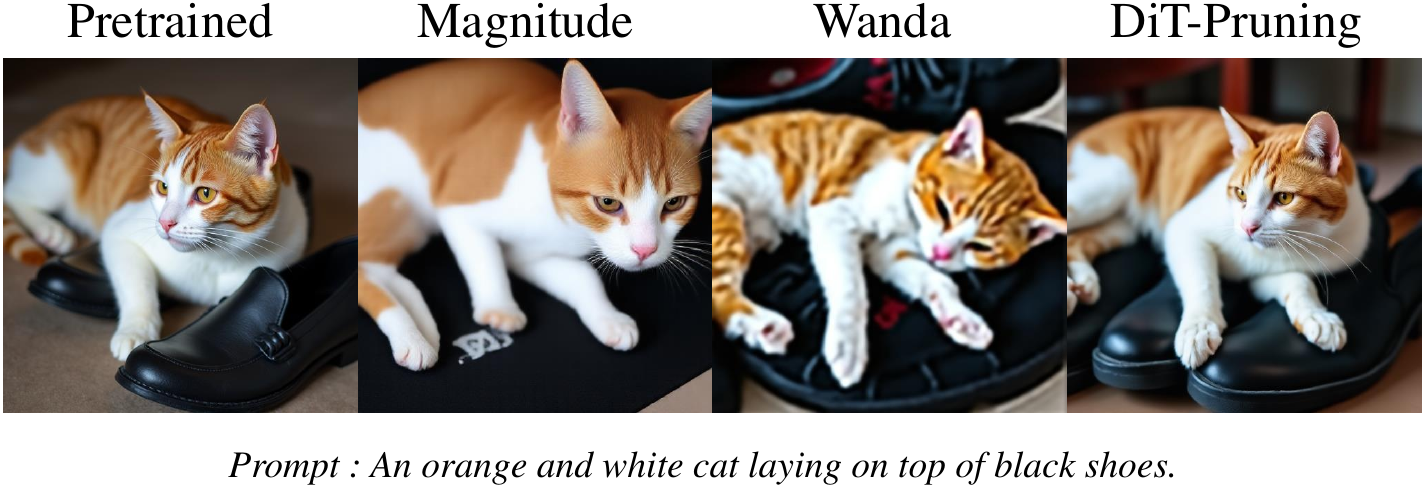}
  \vskip 0.5em 
  \includegraphics[width=\columnwidth,height=0.11\textheight]{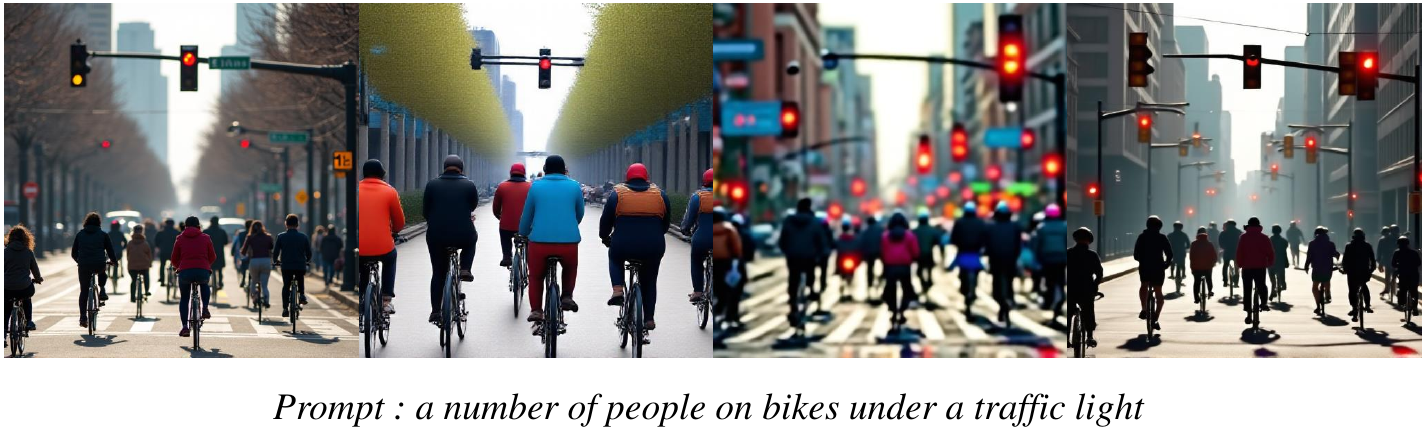}
    \vskip 0.5em 
  \includegraphics[width=\columnwidth,height=0.11\textheight]{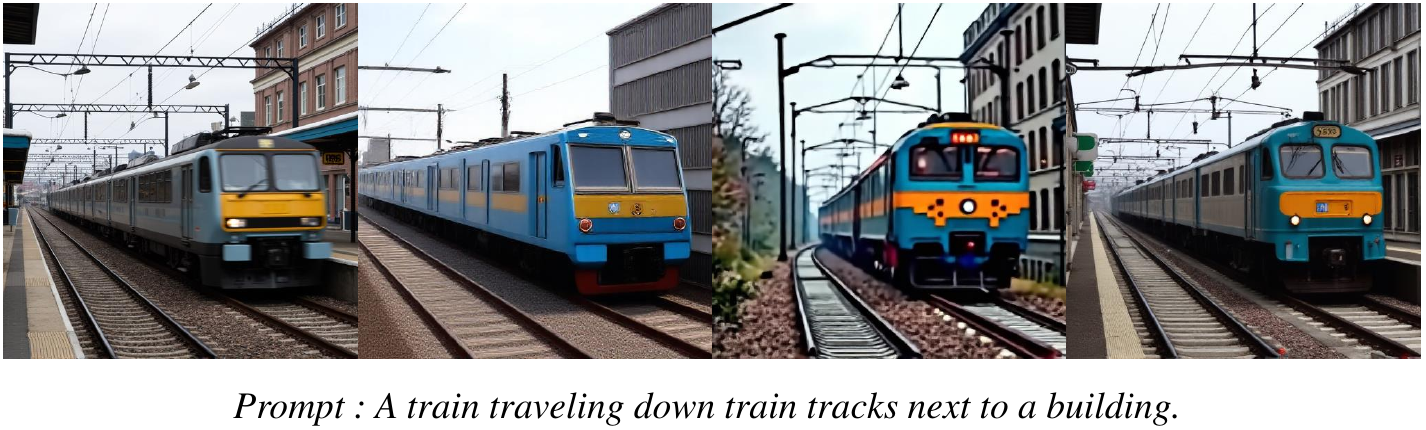}
    \vskip 0.5em 
  \includegraphics[width=\columnwidth,height=0.11\textheight]{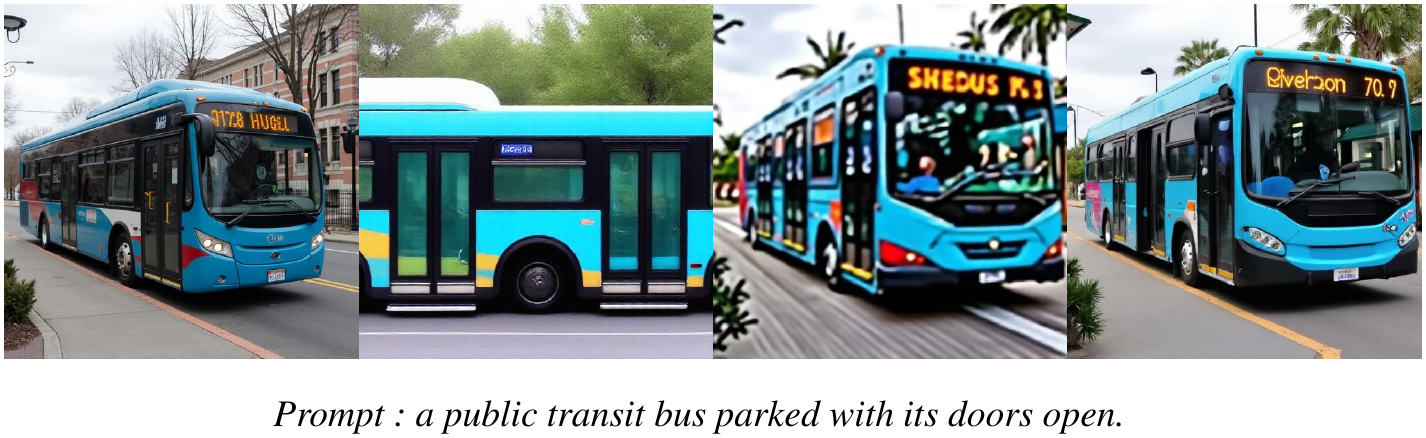}
      \vskip 0.5em 
  \includegraphics[width=\columnwidth,height=0.11\textheight]{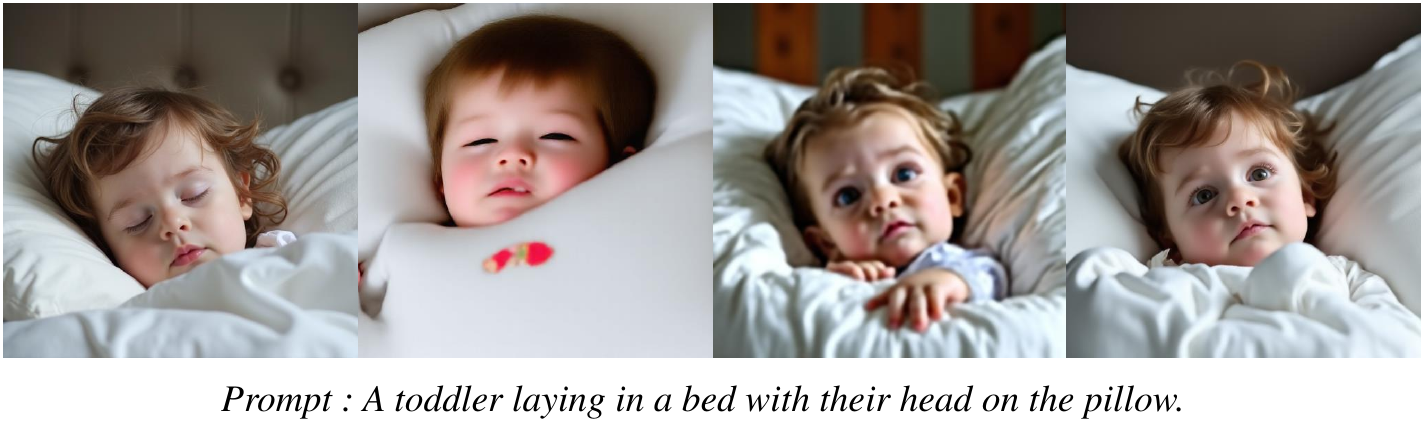}
    \vskip 0.5em
  \includegraphics[width=\columnwidth,height=0.11\textheight]{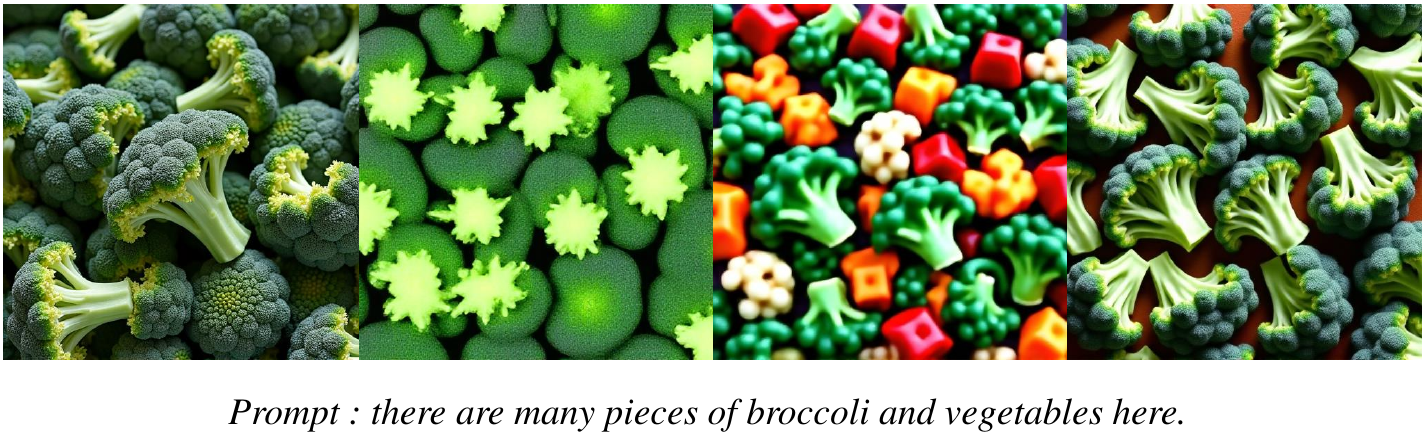}
    \vskip 0.5em
  \includegraphics[width=\columnwidth,height=0.11\textheight]{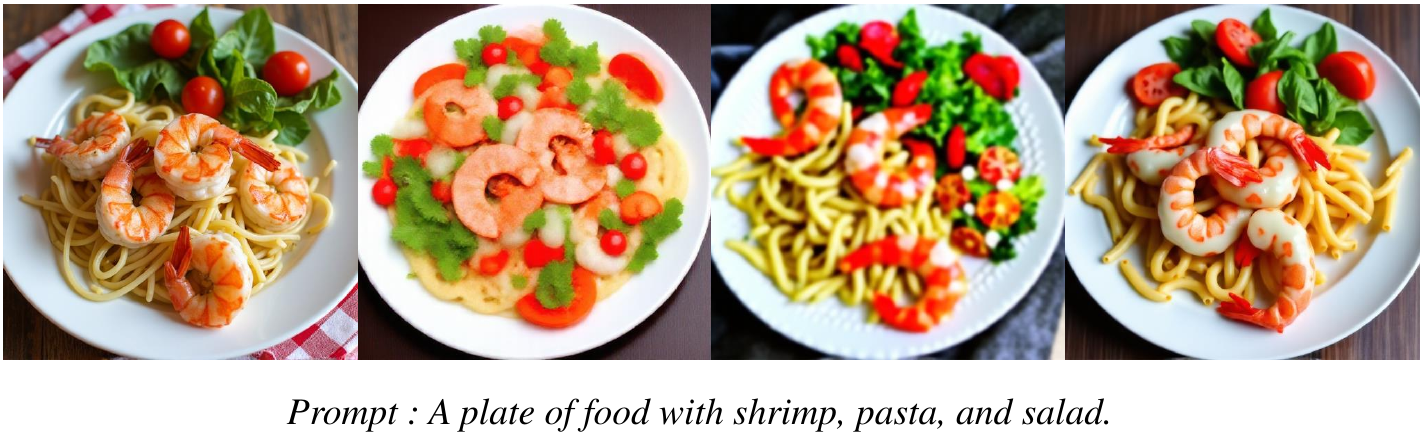}
    \vskip 0.5em
  \includegraphics[width=\columnwidth,height=0.11\textheight]{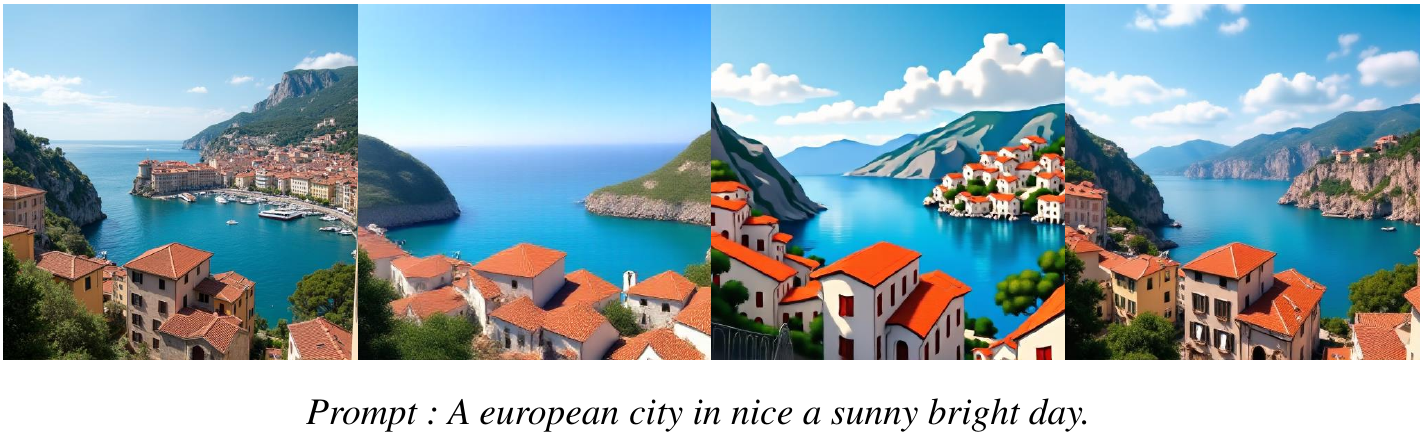}
  \caption{Samples generated by pruned Flux model at 50\% sparsity and 512x512 resolution on Coco datasets.
  }
  \label{samplesflux2}
\end{figure}

\begin{figure}[t] 
  \centering
  \includegraphics[width=\columnwidth,height=0.125\textheight]{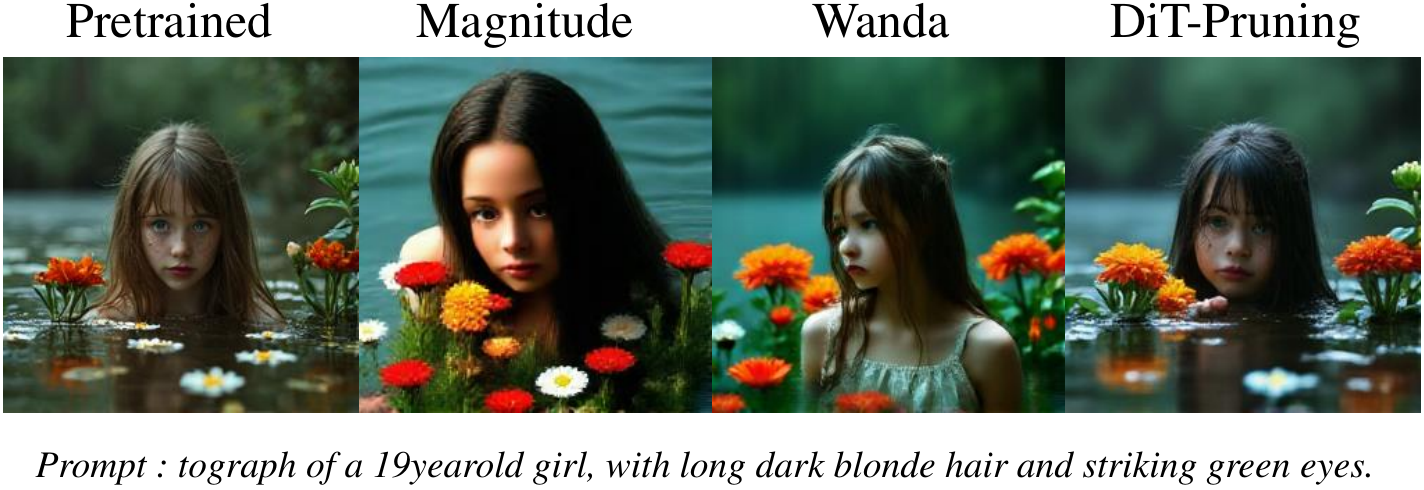}
  \vskip 0.5em 
  \includegraphics[width=\columnwidth,height=0.11\textheight]{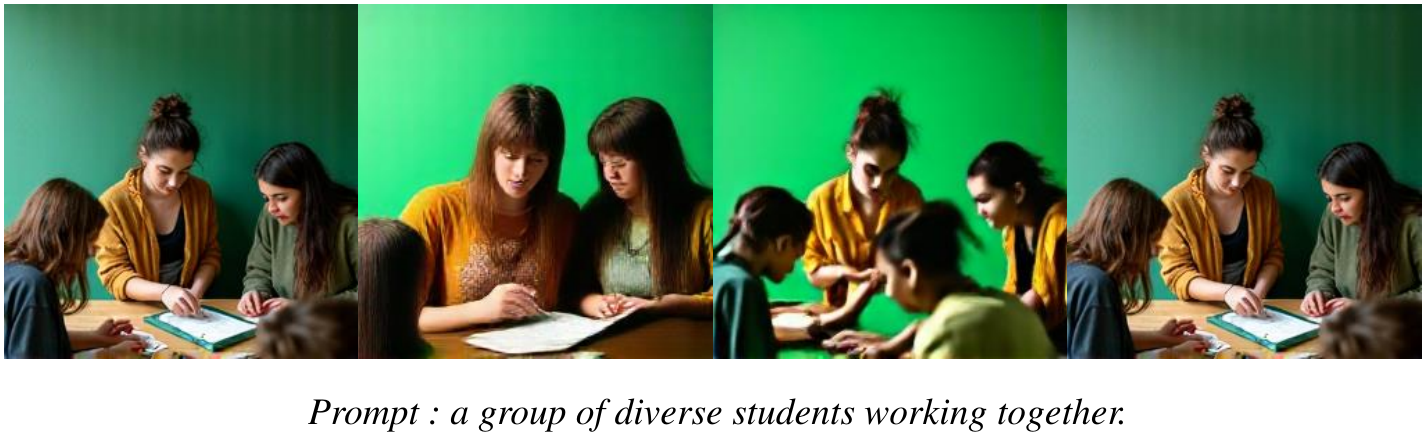}
    \vskip 0.5em 
  \includegraphics[width=\columnwidth,height=0.11\textheight]{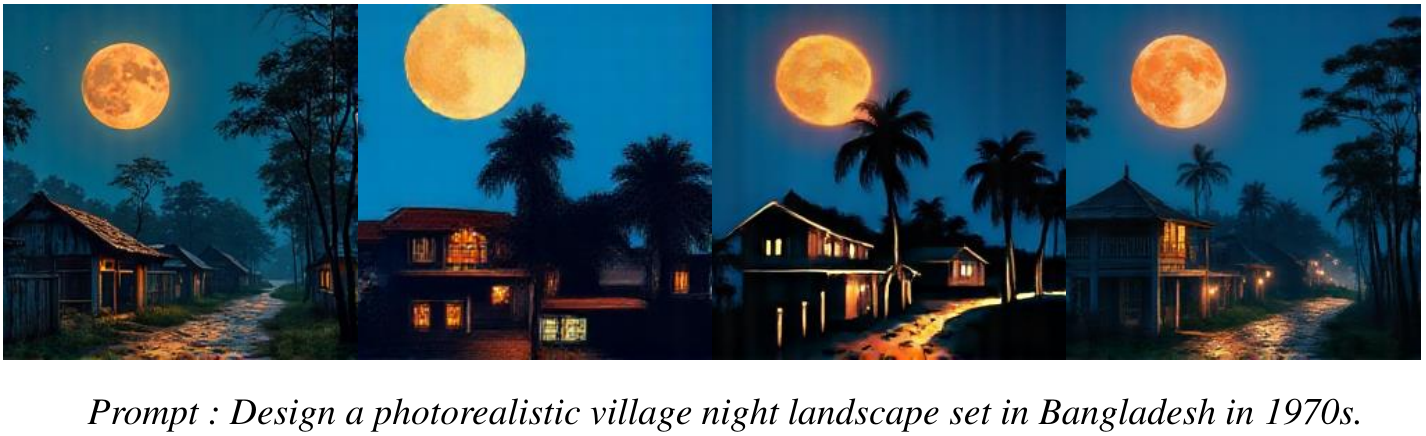}
    \vskip 0.5em 
  \includegraphics[width=\columnwidth,height=0.11\textheight]{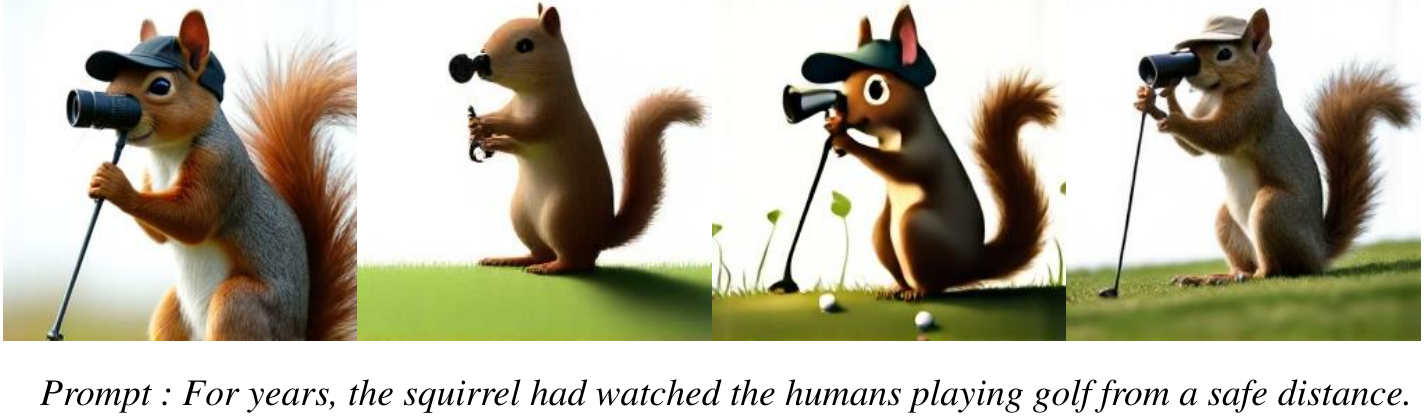}
      \vskip 0.5em 
  \includegraphics[width=\columnwidth,height=0.11\textheight]{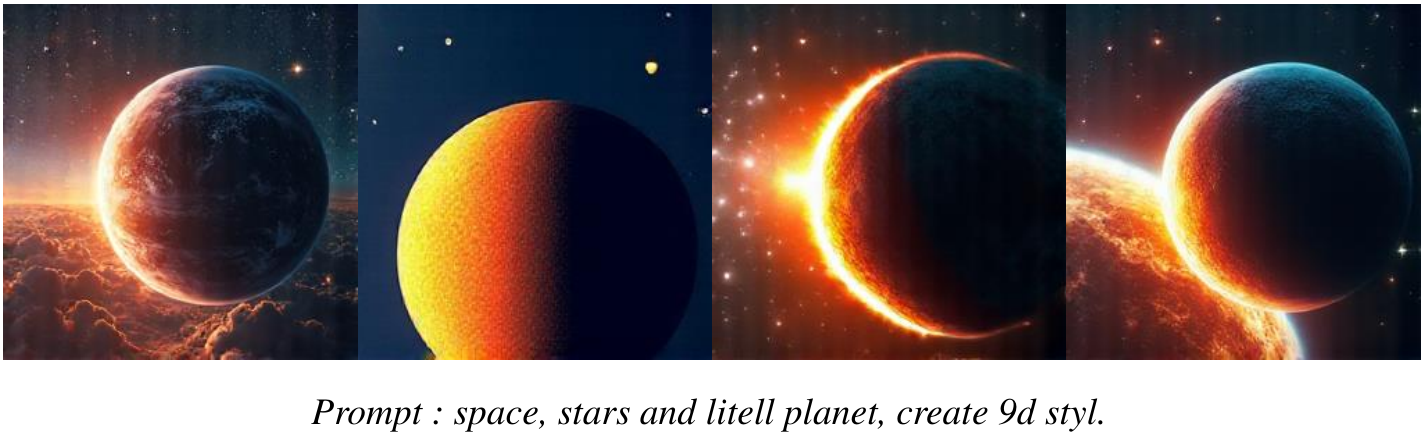}
    \vskip 0.5em
  \includegraphics[width=\columnwidth,height=0.11\textheight]{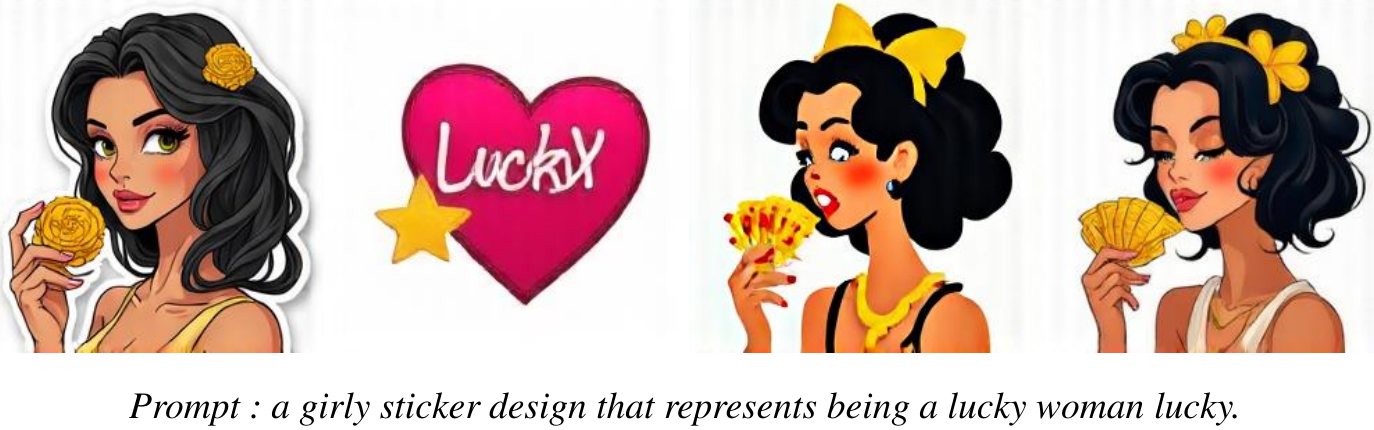}
    \vskip 0.5em
  \includegraphics[width=\columnwidth,height=0.11\textheight]{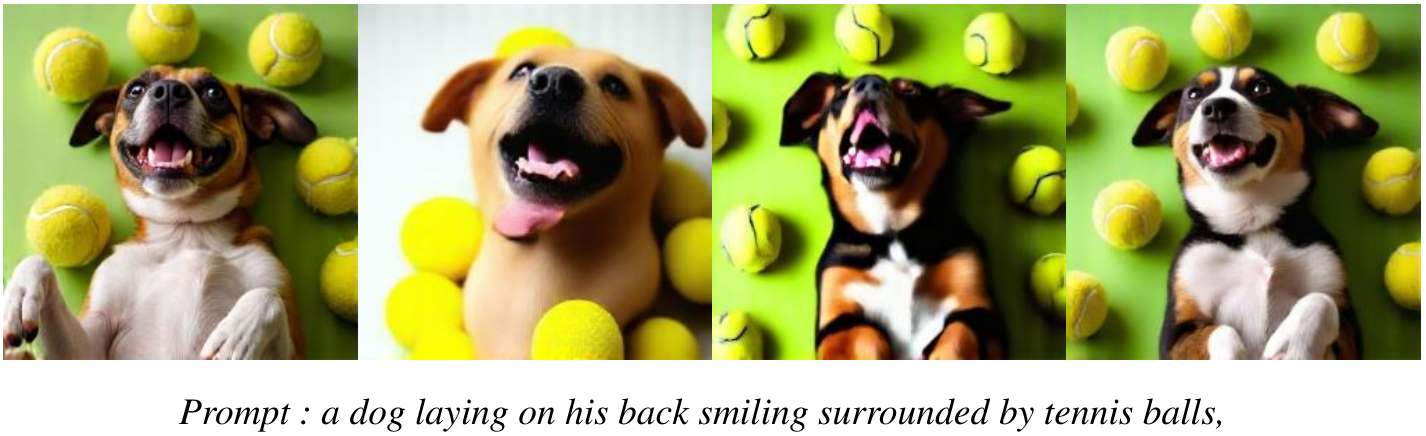}
    \vskip 0.5em
  \includegraphics[width=\columnwidth,height=0.11\textheight]{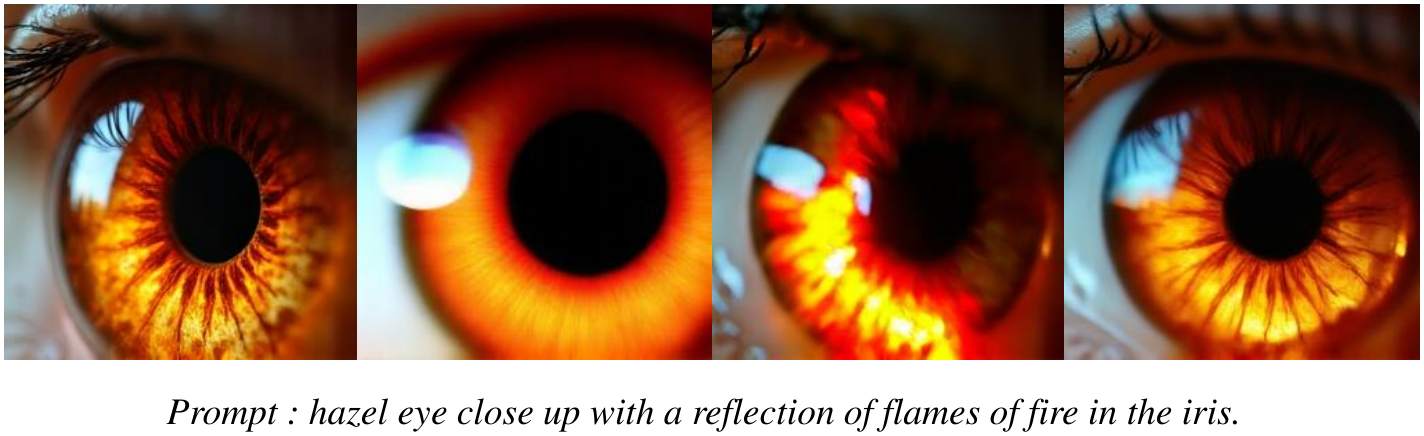}
  \caption{Samples generated by pruned Flux model at 50\% sparsity and 256x256 resolution on MJHQ datasets.
  }
  \label{samplesflux3}
\end{figure}

\begin{figure}[t] 
  \centering
  \includegraphics[width=\columnwidth,height=0.125\textheight]{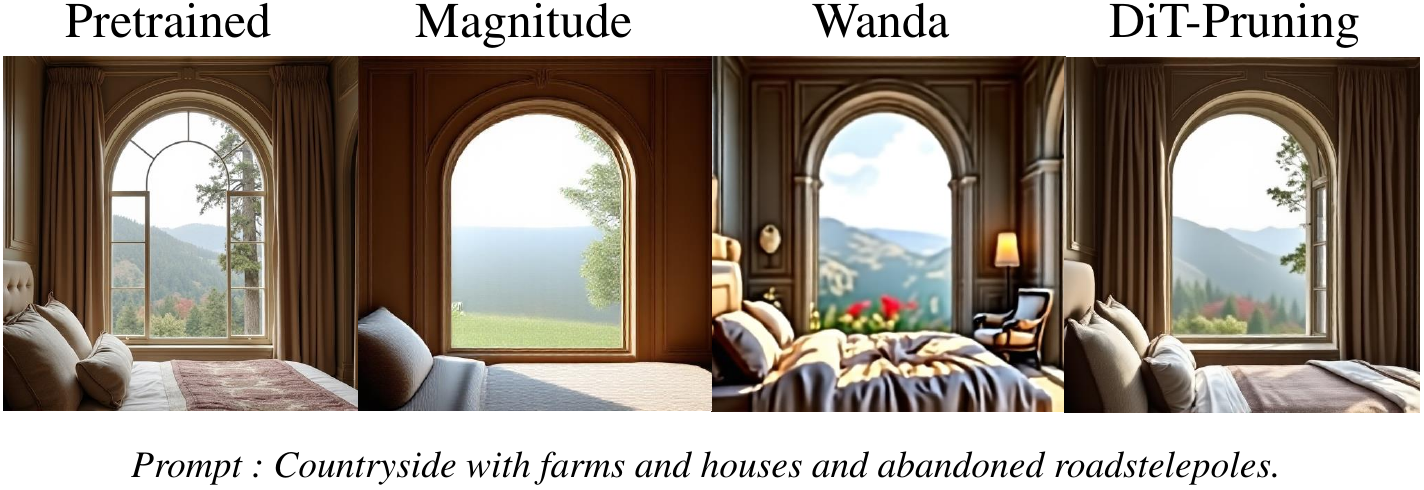}
  \vskip 0.5em 
  \includegraphics[width=\columnwidth,height=0.11\textheight]{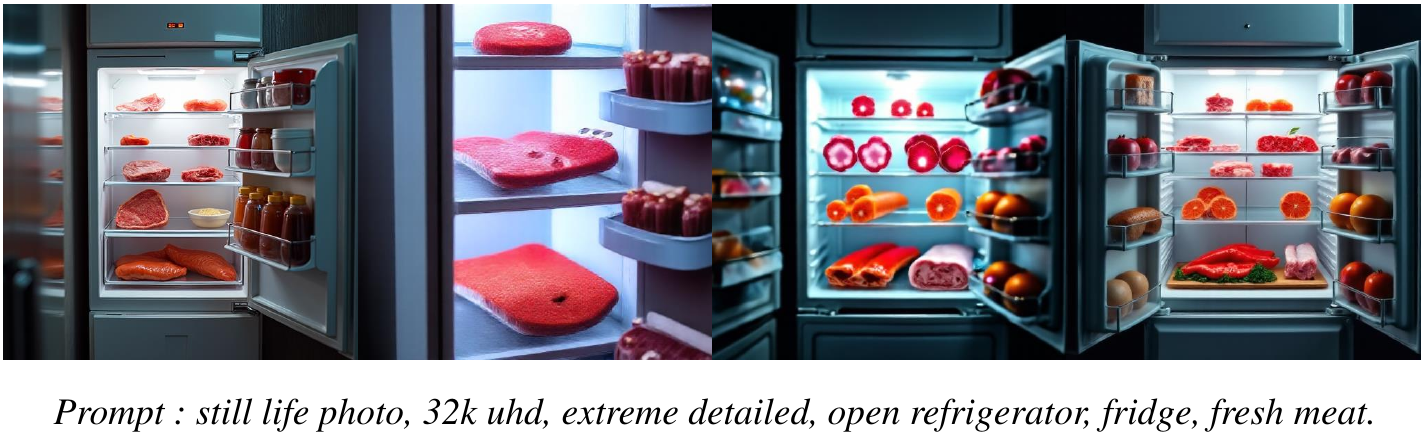}
    \vskip 0.5em 
  \includegraphics[width=\columnwidth,height=0.11\textheight]{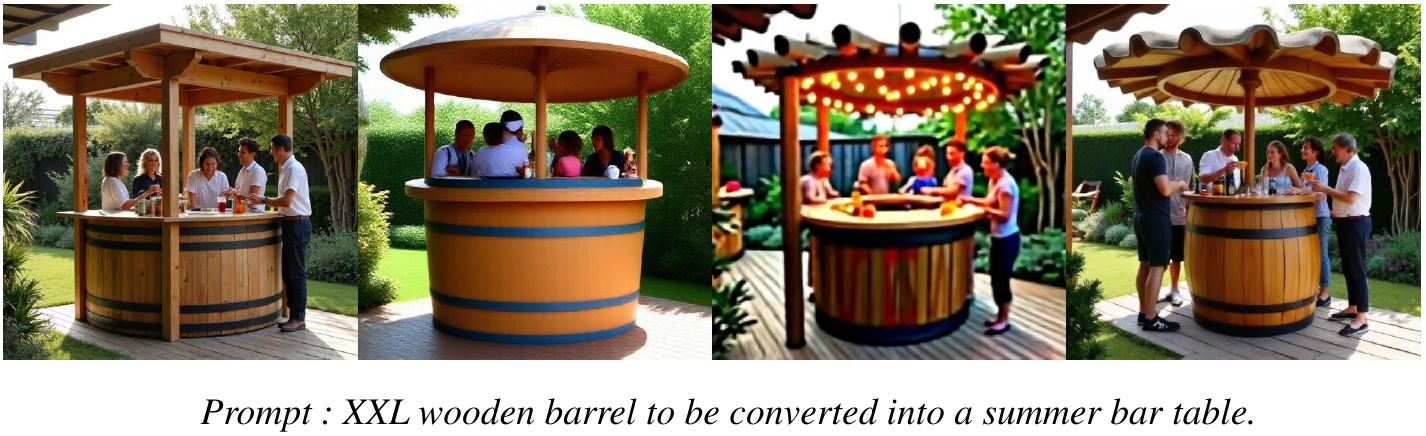}
    \vskip 0.5em 
  \includegraphics[width=\columnwidth,height=0.11\textheight]{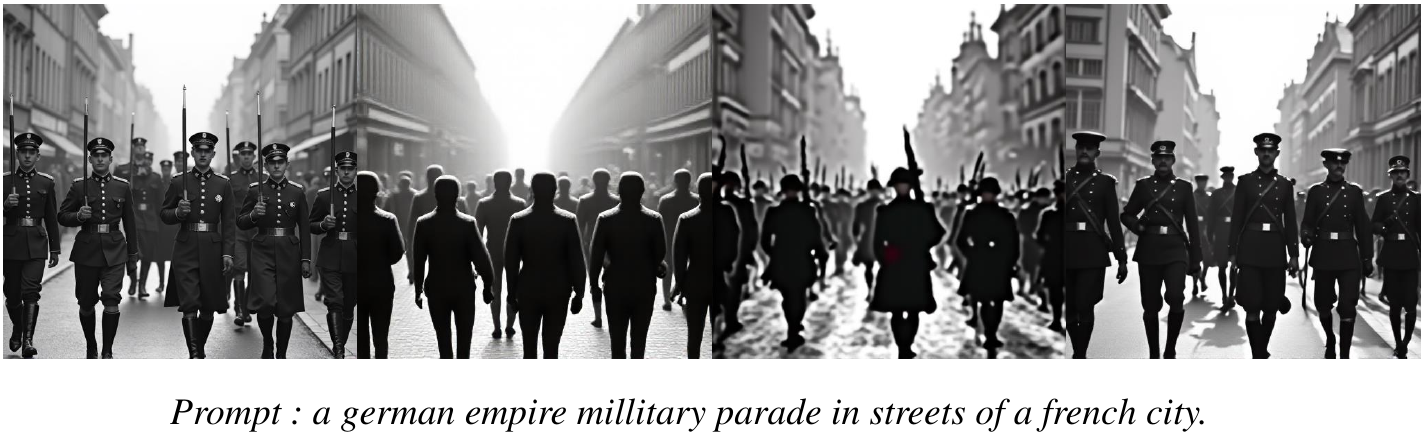}
      \vskip 0.5em 
  \includegraphics[width=\columnwidth,height=0.11\textheight]{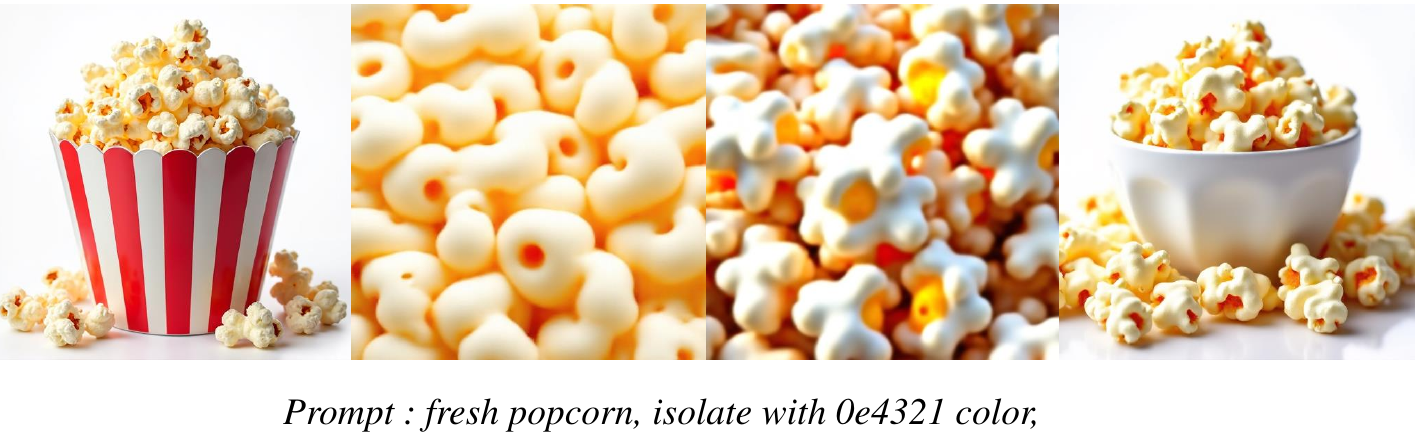}
    \vskip 0.5em
  \includegraphics[width=\columnwidth,height=0.11\textheight]{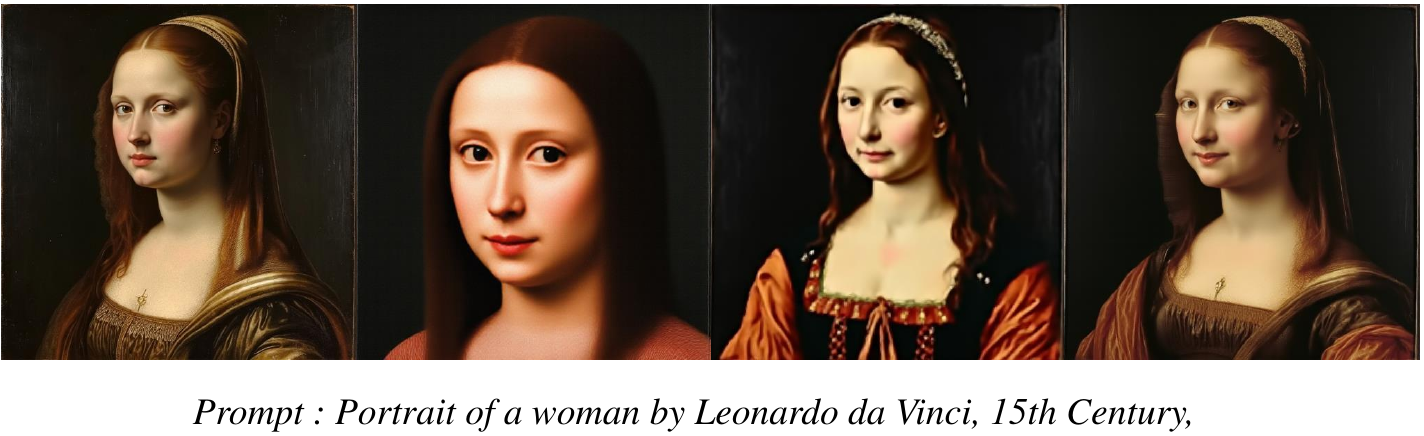}
    \vskip 0.5em
  \includegraphics[width=\columnwidth,height=0.11\textheight]{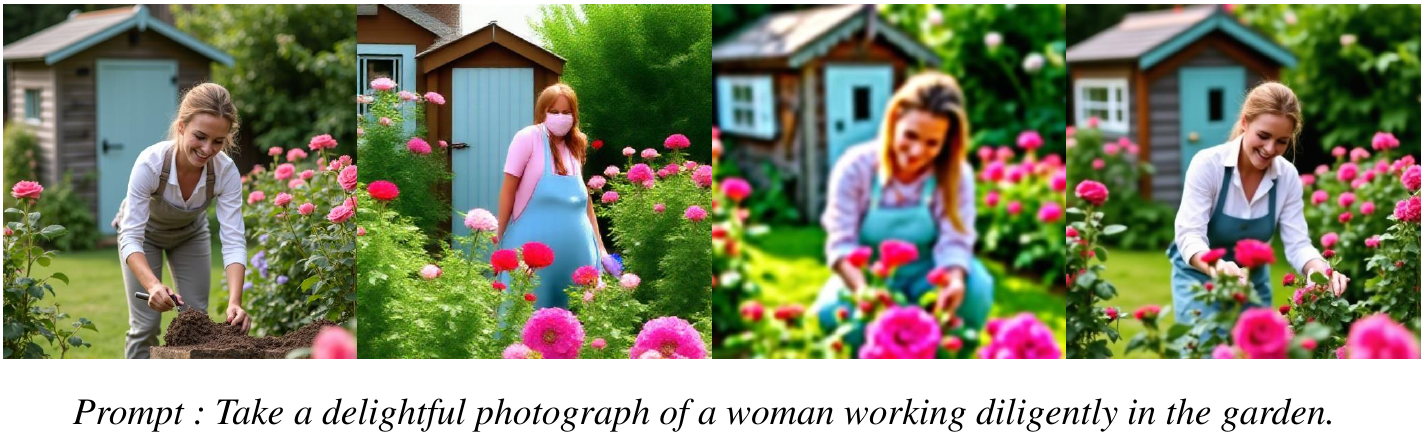}
    \vskip 0.5em
  \includegraphics[width=\columnwidth,height=0.11\textheight]{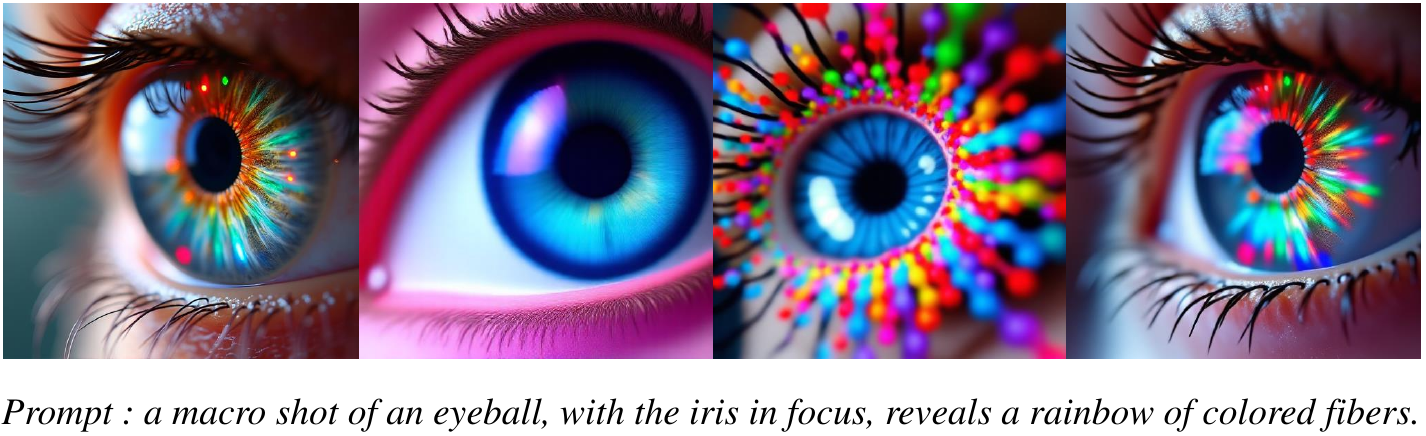}
  \caption{Samples generated by pruned Flux model at 50\% sparsity and 512x512 resolution on MJHQ datasets.
  }
  \label{samplesflux4}
\end{figure}

Figure~\ref{samplesss}, Figure~\ref{samplespixart1}-\ref{samplespixart2}, and Figure~\ref{samplesflux1}-\ref{samplesflux4} provide qualitative comparisons between DiT-Pruning, Wanda, and the dense models under 50\% sparsity level. Across all settings, our method generates images that remain visually closer to the dense model, preserving both structural details and semantic consistency. In contrast, Wanda exhibits noticeable degradation as sparsity increases, including distorted structures, blurred content, and weakened semantic alignment. These results demonstrate that DiT-Pruning effectively preserves generation quality under aggressive sparsification and generalizes consistently across different DiT-based architectures.

\section{Limitations and Broader Impacts}
Our method demonstrates effective pruning of DiTs, achieving favorable generation quality at different sparsity levels (e.g., 20\%, 40\%, 50\%). However, several limitations remain. Under higher sparsity ratios (e.g., 70\% and 80\%), generated images exhibit noticeable degradation, including artifacts, reduced fidelity, and inaccurate semantics. This indicates that our pruning strategy, while efficient, cannot fully preserve model performance under extreme sparsity and suggests the trade-off between compression ratio and generation quality. Moreover, DiTs appear inherently more sensitive to pruning than LLMs, highlighting the need for strategies that better exploit their denoising dynamics and timestep-dependent characteristics. Finally, our experiments focus on standard image generation benchmarks, and the robustness of our approach in more diverse visual domains remains unexplored.

\end{document}